\PassOptionsToPackage{table}{xcolor}
\documentclass{fairmeta}
\usepackage{times}

\usepackage{amsmath,amsfonts,bm}

\def\eqref#1{equation~\ref{#1}}

\def\1{\bm{1}}

\DeclareMathAlphabet{\mathsfit}{\encodingdefault}{\sfdefault}{m}{sl}
\SetMathAlphabet{\mathsfit}{bold}{\encodingdefault}{\sfdefault}{bx}{n}

\DeclareMathOperator*{\argmax}{arg\,max}
\DeclareMathOperator*{\argmin}{arg\,min}

\usepackage{amsmath,amssymb,mathtools}
\usepackage{graphicx}
\usepackage{booktabs,multirow,makecell}
\usepackage{tabularx,longtable}
\usepackage{colortbl}
\usepackage{placeins}
\usepackage{caption}
\usepackage{wrapfig}
\usepackage{needspace}
\usepackage{enumitem}
\setlist[itemize]{leftmargin=1.5em,itemsep=3pt,parsep=0pt,topsep=5pt}
\usepackage{microtype}
\usepackage{hyperref}
\hypersetup{hidelinks,
  pdftitle={SceneForesee: Learning Scene Evolution for Spatiotemporal Reasoning},
  pdfauthor={Chaoyue Li, Shoudong Han, Haoyang Wu, Boxue Yang, Sijia Chen, Yufan Chen, Ruping Zhang}}
\usepackage{url}

\title{SceneForesee: Learning Scene Evolution for Spatiotemporal Reasoning}

\author[1]{Chaoyue Li}
\author[1]{Shoudong Han}
\author[1]{Haoyang Wu}
\author[2]{Boxue Yang}
\authorbreak
\author[1]{Sijia Chen}
\author[1]{Yufan Chen}
\author[1]{Ruping Zhang}

\affiliation[1]{Huazhong University of Science and Technology}
\affiliation[2]{Shanghai Jiao Tong University}

\metadata[Project]{\url{https://github.com/mikubaka88/SceneForesee-EvoSee}}

\newcommand{\benchmark}{SceneForesee}
\newcommand{\task}{\textsc{SEU}}
\newcommand{\method}{EvoSee}

\newcommand{\tablefont}{\fontsize{9.2}{10.6}\selectfont}
\newcommand{\compacttablefont}{\fontsize{8.6}{10.0}\selectfont}
\newcommand{\gain}[1]{{\fontsize{7.5}{8.4}\selectfont\color{gainred}(#1)}}
\newcommand{\bestgain}[2]{\makecell[c]{\textbf{#1}\\[-1pt]\gain{#2}}}

\newcommand{\methodlabelsmall}{\makecell[l]{\bfseries\method{} (Ours)}}

\newcolumntype{Y}{>{\centering\arraybackslash}X}
\newcolumntype{Z}{>{\raggedright\arraybackslash}X}
\newcolumntype{C}[1]{>{\centering\arraybackslash}m{#1}}
\newcolumntype{L}[1]{>{\raggedright\arraybackslash}m{#1}}
\newlength{\halffloatwidth}
\makeatletter
\newcommand{\finishwrap}{\par
  \ifnum\c@WF@wrappedlines>1
    \vspace{\dimexpr\c@WF@wrappedlines\baselineskip-\baselineskip\relax}
  \fi
  \WFclear}
\makeatother

\definecolor{tablegroup}{HTML}{F1F2F4}
\definecolor{observedshade}{HTML}{DCEBF3}
\definecolor{futureshade}{HTML}{FBE5D6}
\definecolor{combinedshade}{HTML}{E8E4F3}
\definecolor{entityshade}{HTML}{D5EEE5}
\definecolor{eventshade}{HTML}{F7E3B2}
\definecolor{evidenceshade}{HTML}{F2D8E5}
\definecolor{trajectoryshade}{HTML}{A9D7CF}
\definecolor{understandingshade}{HTML}{C8C0DF}
\definecolor{oursrow}{HTML}{E9EEF9}
\definecolor{gainred}{HTML}{C23B3B}
\abstract{Large multimodal models (LMMs) need to understand how entities move and interact as a scene develops. This requires following the same entities over time and relating their observed motion to subsequent changes. We formulate Scene Evolution Understanding (\task{}), a task for predicting and interpreting this change. Given observed video, a model estimates entity histories and predicts one scene continuation through future trajectories and Events, accompanied by Visual Evidence from observed motion and a scene description. We construct \benchmark{}, a benchmark for evaluating entity continuity and scene evolution reasoning through paired observed and future scenes with shared entity identities. It contains 144.57K temporal pairs, 1.05M entity correspondences, and 272.49K Events paired with observed evidence. We propose a spatiotemporal reasoning framework, \method{}, that learns future geometry and Event semantics through shared candidate states and joint assignment. It selects one continuation per entity from observation and uses the selected changes for entity refinement, retention, and scene representation. On SceneForesee, EvoSee outperforms Qwen3.5-9B by 10.45 points in Combined Point-F1 and 17.24 points in Event Global F1. On the 4D reasoning benchmark Dyn-Bench, it improves question answering accuracy and grounding J\&F by 4.8 and 3.58 points over their respective base systems. These results support scene evolution as both a prediction target and a dynamic representation that brings predicted entity changes into spatiotemporal reasoning.}

\begin{document}
\maketitle

\section{Introduction}
\label{sec:intro}

Large multimodal models (LMMs) are moving from visual recognition toward reasoning about motion and interaction in videos \citep{qwen2025qwen3vl,huang2026dynbench}. An entity moving behind another retains its identity even as its position, visibility, and spatial relations change. This continuity makes the changes attributable to the same entity, while its observed motion informs what follows. \textbf{Scene evolution} concerns the development of persistent entities and their relations over time, placing continuity at the center of dynamic video reasoning.

Continuity and change appear across several lines of dynamic video research. Video language models interpret extended visual content \citep{bai2025qwen25vl,wang2025internvl35}, while motion grounding and evidence localization relate answers to dynamic observations \citep{deng2025motiongrounded,gao2026dyntrace,wei2026seeing}. Tracking and forecasting establish entity identities and predict their spatial continuation \citep{meinhardt2022trackformer,dendorfer2022quovadis,ashutosh2025fiction}. Event prediction and predictive representations address the changes that follow observation \citep{wang2025futurebench,assran2025vjepa2}. At the intersection of these capabilities, we study how spatial and semantic predictions can refer to the same observed entities and jointly inform scene reasoning.

\begin{figure*}[t]
  \centering
  \includegraphics[page=1,width=\textwidth]{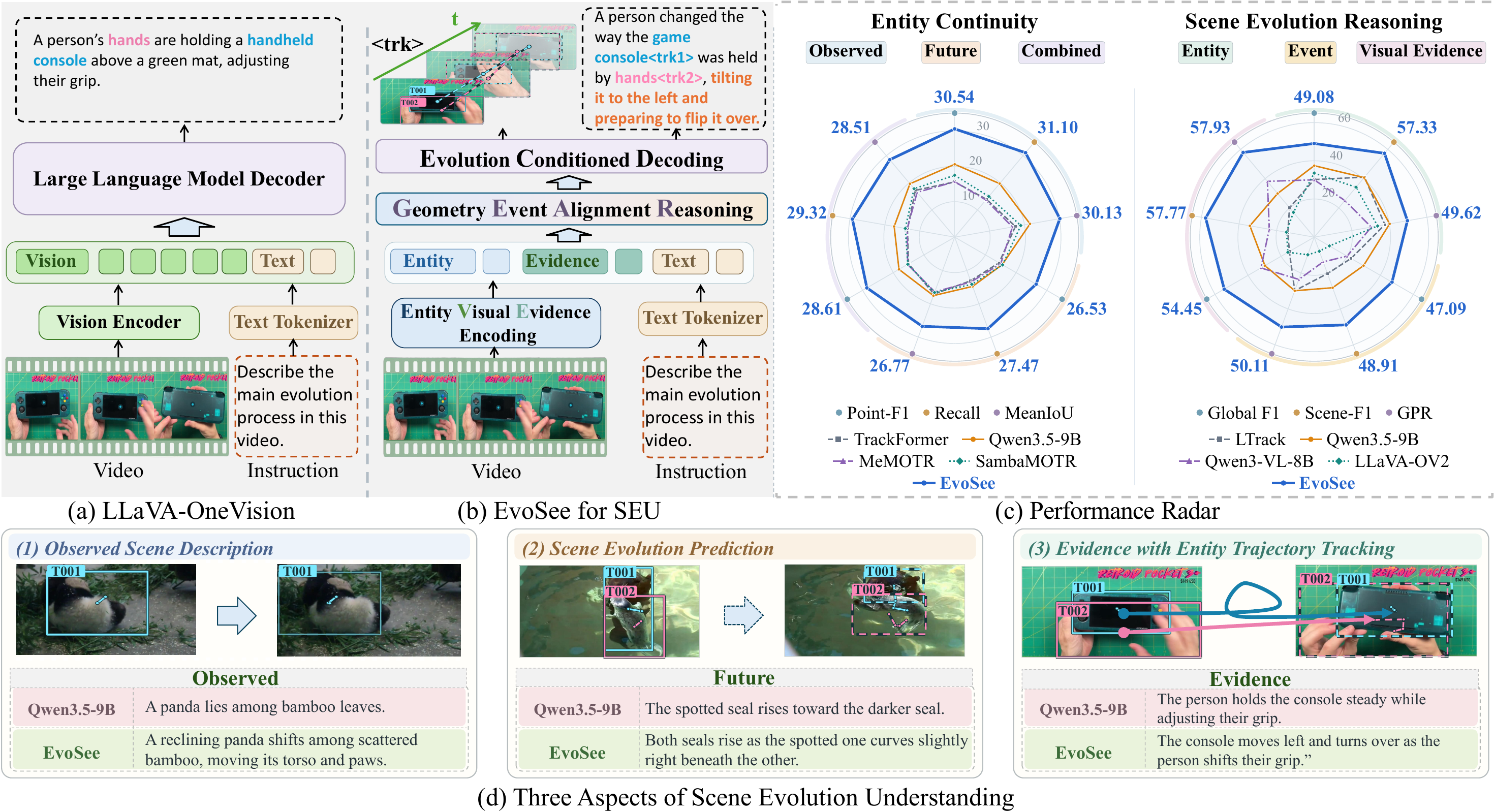}
  \caption{Scene Evolution Understanding. (a,b) Description and entity continuation from observation. (c) Eighteen continuity and reasoning metrics, plotted as percentage scores; labels give EvoSee values. (d) Observed movement, anticipated interaction, and evidence. Future frames show the evaluation reference. Tracking baselines denote task adaptations.}
  \label{fig:task}
\end{figure*}

Relating spatial and semantic change requires knowing which observations belong to the same entity. Persistent identity supplies this reference when observation ends and future prediction begins. In Figure~\ref{fig:task}(a,b), following the console and the person's hands makes their changing positions and interaction attributable to the same participants. A trajectory describes where an entity moves, while an Event summarizes its state or relations. These spatial and semantic descriptions concern the same continuation and can be learned and evaluated with a common entity reference. Visual Evidence records the observed motion associated with that continuation, such as the changing grip and console orientation in the example. Together, these descriptions relate predicted change to the entities and motion from which it is inferred, providing context for scene reasoning.

We formulate \textbf{Scene Evolution Understanding} (\task{}) as a task for predicting and interpreting how persistent entities develop. From observed video, a model estimates entity histories and infers one scene continuation through future trajectories and Events. Its output also identifies associated Visual Evidence in observation and describes the evolving scene. We construct \textbf{\benchmark{}}, a benchmark for evaluating scene evolution understanding through paired observed and future intervals with shared entity identities. Its \textbf{144.57K} temporal pairs contain \textbf{1.05M} entity correspondences and \textbf{272.49K} Events paired with observed evidence. This organization supports complementary evaluations of entity continuity and scene evolution reasoning on the same scenes.

To learn and use scene evolution, we propose a spatiotemporal reasoning framework, \textbf{\method{}}. EvoSee treats the observed entity as the reference for both its future motion and the Events in which it participates. For each entity, it learns candidate continuations whose geometric and Event predictions share future states and joint assignment during training. It selects one candidate using the observed entity state and evidence. The selected change then refines the entity state and contributes to entity retention and scene representation.

EvoSee outperforms the evaluated baselines in entity continuity and Event category recognition on SceneForesee. Against Qwen3.5-9B, it also improves complete Event prediction, where the category and all participants must match. Ablations relate these gains to joint assignment, continuation selection, and the use of predicted change. The value of this change extends to question answering and grounding on Dyn-Bench \citep{huang2026dynbench}, where the learned scene representation supports 4D reasoning.

\Needspace{8\baselineskip}
Our contributions are summarized as follows.
\begin{itemize}
  \item We formulate a new task, \textbf{Scene Evolution Understanding}, which aims to predict and interpret how persistent entities evolve over time. The task represents one scene continuation through future entity motion and Events, together with associated visual evidence from observation.
  \item We introduce \textbf{\benchmark{}}, a benchmark for evaluating entity continuity and scene evolution reasoning. Shared identities associate observed motion, future trajectories, and Events, supporting complementary evaluations of continuity and change on the same scenes.
  \Needspace{7\baselineskip}
  \item We propose \textbf{\method{}}, an entity-centric spatiotemporal reasoning framework that models future geometry and Event semantics as complementary aspects of scene evolution. Shared future states and joint assignment align geometric and Event predictions for each entity, and the selected changes update entity and scene representations.
  \item We demonstrate experimentally that modeling scene evolution improves future prediction and observed entity estimation. The learned representations also improve dynamic question answering and grounding on the \textbf{4D reasoning benchmark Dyn-Bench}.
\end{itemize}

\section{Related Work}
\label{sec:related}

\paragraph{Following entities through time.}
Tracking establishes correspondence between observations of the same entity. Object query trackers maintain this correspondence across frames \citep{meinhardt2022trackformer,gao2023memotr}, while other formulations predict identities \citep{gao2025motip} or associate tracked entities with descriptions and referring expressions \citep{li2024bensmot,wu2023rmot}. These entity histories support forecasting of subsequent locations and interactions beyond observation \citep{dendorfer2022quovadis,lu2026gptf,liang2019peeking,ashutosh2025fiction,nguyen2025hyperglm}. SEU builds on this identity correspondence to attribute future motion and Events to the same observed entities.

\paragraph{Predicting scene development.}
Beyond future locations, prediction concerns changes in entity states and relations. FutureBench and FeVOS study future events \citep{wang2025futurebench,lan2026fevos}, while scene graph forecasting represents changes through entities and their relations \citep{pistilli2026learning}. Anticipatory detection and predictive representations learn expected future states \citep{papais2025foresight,assran2025vjepa2}. SEU uses persistent identities to relate future trajectories, Events, and observed Visual Evidence within the same scene continuation.

\paragraph{Reasoning with motion and visual evidence.}
Video language models use dynamic visual content to produce descriptions and answers \citep{bai2025qwen25vl,wang2025internvl35,zhang2024llavavideo}. Motion grounding and evidence localization identify observations relevant to these outputs \citep{deng2025motiongrounded,galoaa2026motiono,wei2026seeing}. DynTrace represents observed motion through traces, and Dyn-Bench evaluates dynamic question answering and grounding \citep{gao2026dyntrace,huang2026dynbench}. EvoSee uses predicted entity continuations to enrich the scene representation for such reasoning.

\section{Task and Benchmark}
\label{sec:task}

\subsection{Scene Evolution Understanding}

Scene Evolution Understanding requires a model to estimate entity histories from observed video and predict one scene continuation through future trajectories and Events with explicit participants. The observed entities provide the reference for these predictions. The output also identifies Visual Evidence in observation and describes the scene evolution. Figure~\ref{fig:task}(d) illustrates observed motion, anticipated interaction, and visual evidence through example scenes. Future frames supply training and evaluation targets; Appendix~\ref{app:task} specifies sampling and output format.

\subsection{Benchmark Construction}
\label{sec:benchmark}

\begin{figure*}[t]
  \centering
  \includegraphics[page=2,width=\textwidth]{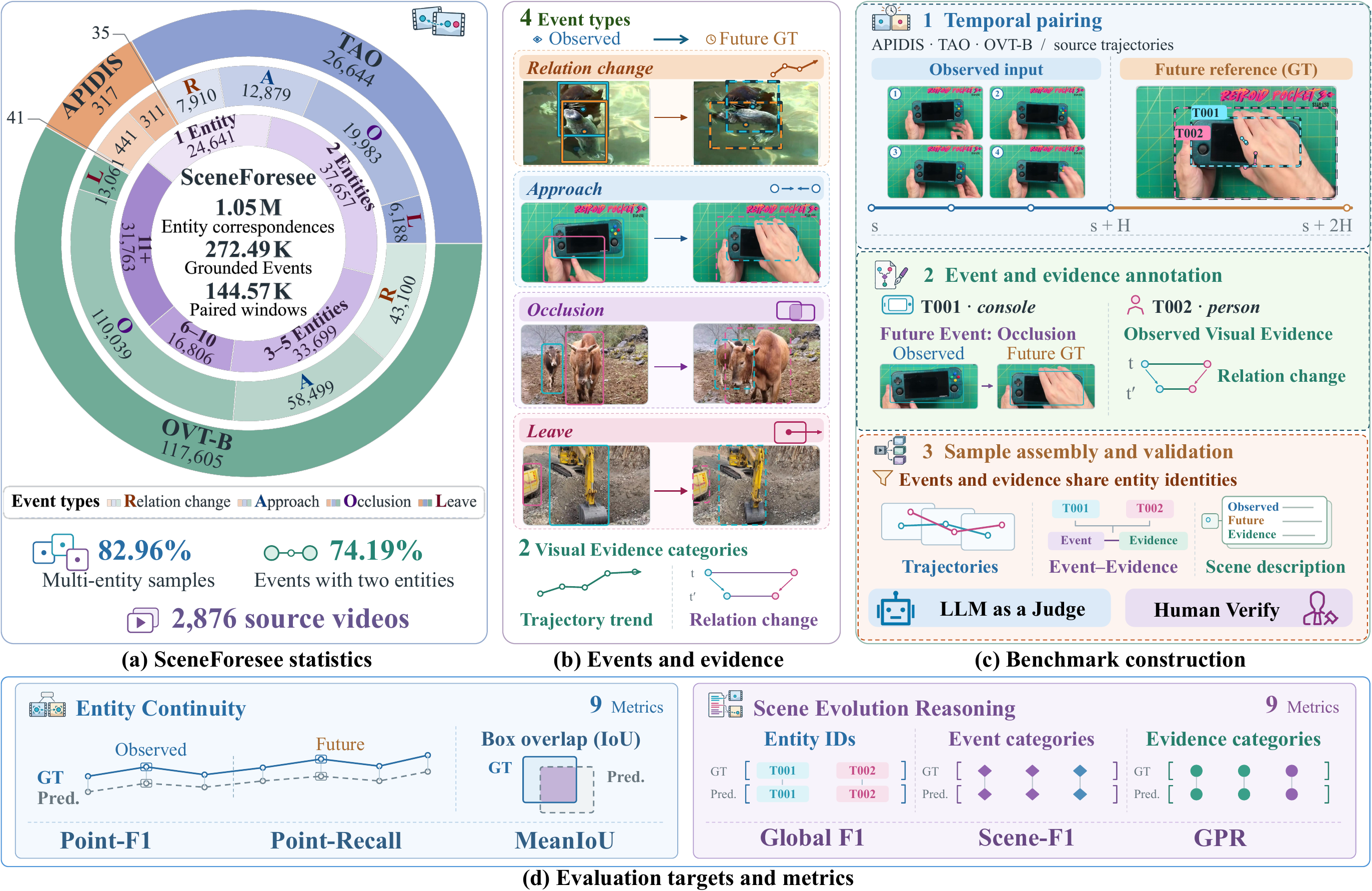}
  \caption{SceneForesee. (a) Corpus statistics. (b) Event and Visual Evidence categories. (c) Temporal pairing, annotation, and sample validation. (d) Entity Continuity and Scene Evolution Reasoning metrics. Ring labels give counts; sector angles are schematic.}
  \label{fig:benchmark}
\end{figure*}

SceneForesee pairs adjacent, equally long observed and future intervals from APIDIS, OVT-B, and TAO, covering sports interactions, open category entities, and varied natural motion \citep{devleeschouwer2008distributed,liang2024ovtb,dave2020tao}. Timestamped boxes preserve shared identities between observed and future intervals, with at least one continuing entity per sample. These identities anchor the annotations in Figure~\ref{fig:benchmark}(c).

The four Event categories in Figure~\ref{fig:benchmark}(b) summarize trajectory configurations, providing discrete targets for predicting evolution under shared identities. Approach uses decreasing separation. Occlusion uses box overlap throughout the image at the transition from observation to prediction. Leave uses proximity to the image edge at the future endpoint, while relation change covers remaining configurations. Each Event records participants and a future interval, paired with observed Visual Evidence of individual trajectory trends or pairwise relation changes. Figure~\ref{fig:benchmark}(c) pairs the console's anticipated occlusion with the entities' observed relation. A scene description expresses their motion and development.

Common annotation criteria apply to all sources, with LLM assessment and manual inspection of generated annotations. Data splits follow source videos and disjoint APIDIS episodes, keeping overlapping windows together. Figure~\ref{fig:benchmark}(a) summarizes 2,876 videos, with multiple entities in 82.96\% of samples and two participants in 74.19\% of Events. Appendix~\ref{app:construction} details construction and splits.

\subsection{Evaluation Metrics}

Figure~\ref{fig:benchmark}(d) presents complementary evaluations of entity continuity and scene evolution reasoning on the same temporal pairs.

\textbf{Entity Continuity.} Identity matching from observed trajectories remains fixed across all intervals. Point-F1@0.5 and Point-Recall@0.5 measure recovery of entity locations under this correspondence, while MeanIoU measures spatial overlap. Combined scores weight observed and future scores by their annotated point counts.

\textbf{Scene Evolution Reasoning.} Entity scores evaluate identities referenced in descriptions; Event and Visual Evidence scores evaluate category occurrences. Global F1 and the geometric mean of global precision and recall (GPR) pool matches across scenes; Scene-F1 weights scenes equally. Complete Event micro-F1 additionally requires the category and all participants to match. BERTScore F1 measures similarity between complete descriptions \citep{zhang2020bertscore}. Appendix~\ref{app:evaluator} defines matching and aggregation.

\section{Method}
\label{sec:method}

\begin{figure*}[ht]
  \centering
  \includegraphics[page=3,width=\textwidth]{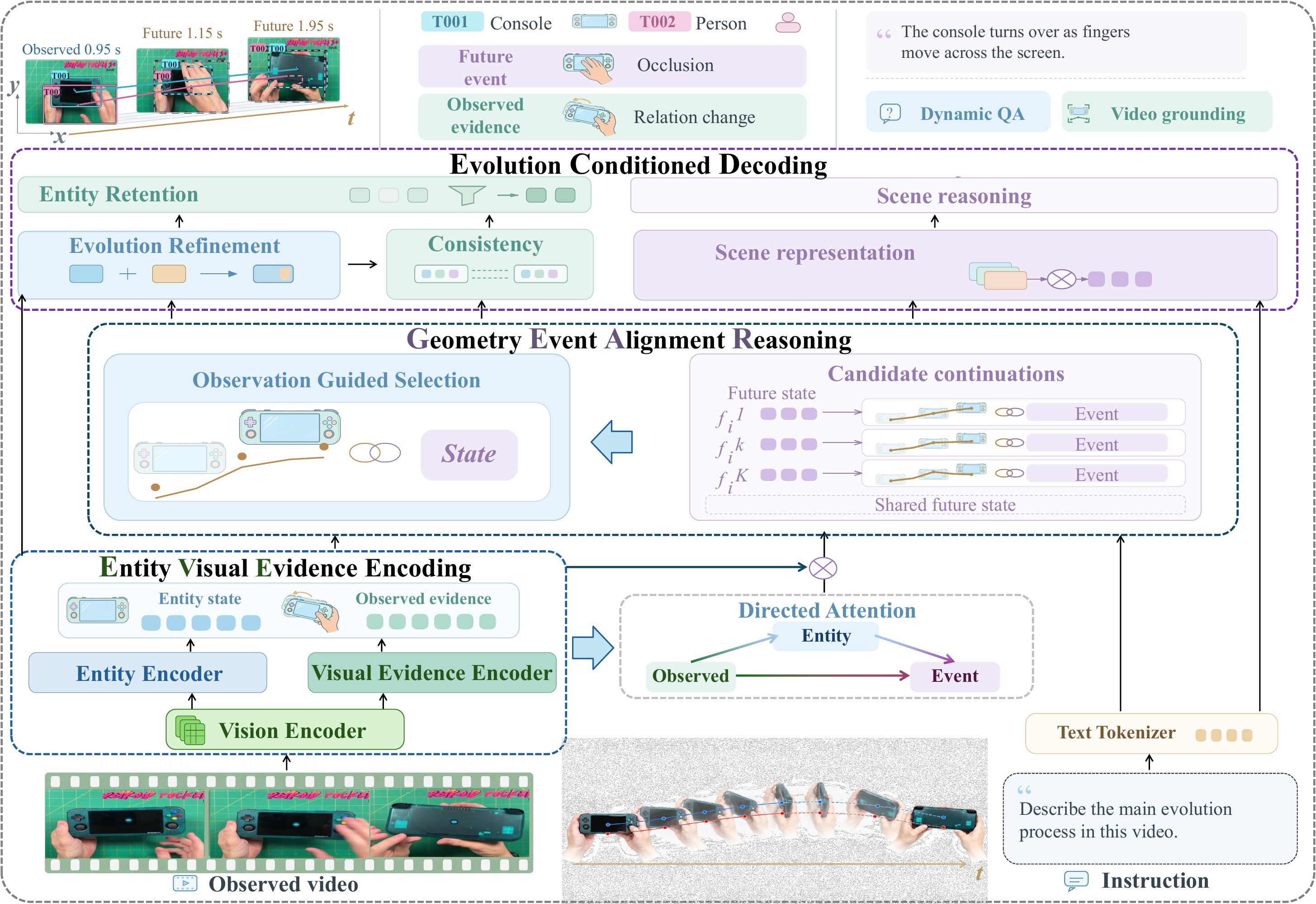}
  \caption{EvoSee. Entity state and evidence condition shared geometric and Event continuations. Selected changes inform refinement, retention, and scene representation. Entity in directed attention denotes geometric state.}
  \label{fig:method}
\end{figure*}

EvoSee learns entity continuations from observed video and uses the predicted changes for scene reasoning. Figure~\ref{fig:method} shows how observed entity states and evidence condition shared geometric and Event predictions. A selector uses the observed representations to choose one continuation per entity. The selected change refines the entity state and contributes to retention and scene representation.

\subsection{Entity Visual Evidence Encoding}
\label{sec:representation}

Entity Visual Evidence Encoding (EVE) uses a shared learned query $q_i$ to encode entity state and observed motion. The query attends to observed visual memory $X_{\mathrm{obs}}$ for state $a_i$ at the observation boundary and to frame features $X_t$ for temporal states $h_i^t$ \citep{meinhardt2022trackformer,vaswani2017attention}. Normalized relevance scores aggregate these temporal states into evidence $e_i$. Evidence intervals supervise temporal relevance, and evidence categories supervise the aggregated representation. Appendix~\ref{app:representation} gives the encoding equations.

\subsection{Geometry Event Alignment Reasoning}
\label{sec:joint-modeling}
\label{sec:hypotheses}

Geometry Event Alignment Reasoning (GEAR) infers future geometry and Event semantics from entity state and evidence. Their sum $u_i^a=a_i+e_i$ initializes geometric and semantic states $u_i^g,u_i^s$. Stacking $u_i^a,u_i^g,u_i^s$ in this order gives $U_i$. Each attention head $h$ projects normalized $U_i$ into queries $Q_i^h$, keys $K_i^h$, and values $V_i^h$ of width $d_h$, computing output $A_i^h$ under mask $\mathcal M$.
\begin{equation}
 A_i^h=\operatorname{softmax}\!\left(\frac{Q_i^h(K_i^h)^\top}{\sqrt{d_h}}+\mathcal M\right)V_i^h,
 \qquad
 \mathcal M=\begin{pmatrix}0&-\infty&-\infty\\0&0&-\infty\\0&0&0\end{pmatrix}.
 \label{eq:evolution-attention}
\end{equation}
The mask implements the dependencies in Figure~\ref{fig:method}, allowing each state to attend only to itself and preceding states. Concatenated head outputs are projected for a residual update of $U_i$.

Each updated state independently weights feedforward transformations, preserving these dependencies. The resulting geometric and semantic states condition a future sequence $f_i\in\mathbb R^{L\times d}$ with $L$ time steps and feature width $d$. Each of $K$ candidate continuations uses embedding $d_k\in\mathbb R^d$ and transition $\phi_k$ \citep{rupprecht2017learning}. The vector $\mathbf1_L$ broadcasts the embedding across time. As depicted in Figure~\ref{fig:method}, decoders $D_g$ and $D_s$ share candidate state $f_i^k$ to predict geometry $g_i^k$ and Event logits $s_i^k$, with task context embedding $c$ informing semantics.
\begin{equation}
 f_i^k=f_i+\phi_k(f_i+\mathbf1_Ld_k^\top),\qquad g_i^k=D_g(f_i^k),\qquad s_i^k=D_s(f_i^k;c).
 \label{eq:hypothesis}
\end{equation}
Joint assignment selects one candidate for training against annotated future geometry $g_i^\star$ and Event targets $s_i^\star$, using their combined prediction error.
\begin{equation}
 \widehat k_i\in\argmin_k\!\left\{\ell_{\mathrm{geo}}(g_i^k,g_i^\star)+\lambda_s\ell_{\mathrm{sem}}(s_i^k,s_i^\star)\right\}.
 \label{eq:hypothesislearning}
\end{equation}
Here $\ell_{\mathrm{geo}}$ measures box error, $\ell_{\mathrm{sem}}$ measures Event classification error, and $\lambda_s=0.20$. Each Event supplies its category target to every participant, including both identities for pairwise Events. All future supervision uses the assigned index, coupling geometric and Event learning within the same candidate.

The assigned index also supervises selection from observation. A selector $R$ maps entity state, evidence, and observed motion embedding $o_i$ into candidate probabilities $\boldsymbol\rho_i$. It learns $\widehat k_i$ through negative log likelihood. At inference, the most probable candidate $k_i^*$ determines future boxes, visibility, persistence, and Event scores, giving one continuation per retained entity. Appendix~\ref{app:joint-learning} details selection and the endpoint separation regularizer used during training.

\subsection{Evolution Conditioned Decoding}
\label{sec:future-update}

Evolution Conditioned Decoding (ECD) refines the observed entity state using predicted change. Projecting temporally averaged future boxes and the sigmoid of temporally averaged Event logits gives summaries $\mathbf g_i$ and $\mathbf s_i$. The summaries use probability weighted decoded predictions during training and the selected candidate at inference. Map $\Phi$ combines them with observed state and evidence into evolution representation $\mathbf m_i$. A sigmoid gate from $\Gamma$ weights the projected evolution representation $P_f(\mathbf m_i)$.
\begin{equation}
 \mathbf m_i=\Phi([a_i;\mathbf g_i;\mathbf s_i;e_i]),\qquad
 a_i^+=a_i+\sigma\!\left(\Gamma([a_i;\mathbf m_i])\right)\odot P_f(\mathbf m_i).
 \label{eq:evolution-representation}
\end{equation}
Here $\sigma$ is sigmoid and $\odot$ denotes elementwise multiplication. The residual keeps $a_i$ as a reference, while the updated state supports observed trajectory estimation and entity confidence.

\label{sec:consistency}
Entity retention combines confidence $\ell_i^{\mathrm{obj}}$ from updated state $a_i^+$ with agreement among observed and predicted features. Six cosine distances measure relations among evidence, entity state, geometry, semantics, evolution representation, and task context. Their mean $B_i$ enters retention score $\kappa_i=\ell_i^{\mathrm{obj}}-\lambda_{\mathrm{cons}}B_i$, with nonnegative weight $\lambda_{\mathrm{cons}}$. A margin ranking loss favors lower distances for queries matched to annotations than for unmatched queries. The highest retention scores define set $\mathcal I^*$. Appendix~\ref{app:consistency} specifies the relations and supervision.

\label{sec:scene-reasoning}
Retained entities supply trajectories and Event scores from their selected continuations. Complete Events specify a category and all participants using predicted entity IDs, which evaluation maps through the observation correspondence. For scene reasoning, EvoSee averages entity states $a_i$, evidence $e_i$, and evolution representations $\mathbf m_i$ over all queries. Their concatenation with task context $c$ is projected into scene representation $h_{\mathrm{scene}}$ (Appendix~\ref{app:consistency}). This representation supplies observed and predicted context for description, dynamic question answering, and grounding \citep{deng2025motiongrounded,gao2026dyntrace}.

\section{Experiments}
\label{sec:experiments}

SceneForesee comparisons share test samples, observation intervals, temporal extents, and metrics. Qwen3.5-9B and EvoSee predict trajectories; other language models use shared predictions from the TrackFormer-style task adaptation \citep{meinhardt2022trackformer}. Tables~\ref{tab:trajectory}--\ref{tab:semantic} report eighteen metrics as percentages; ablation and difficulty tables average each metric triplet. Ablations share training and evaluation conditions. Dyn-Bench evaluates the learned scene representation for 4D reasoning. Appendix~\ref{app:protocols} details both protocols.

\subsection{Comparison with Baselines}

\textbf{Entity Continuity.} Table~\ref{tab:trajectory} shows EvoSee leading all nine continuity metrics. Combined Point-F1 exceeds Qwen3.5-9B by 10.45 points, while Future Point-F1 exceeds SambaMOTR by 10.92 points. Recall and overlap gains indicate more complete recovery of entity locations and more accurate spatial continuation.
\begin{table*}[ht]
\centering
\begin{minipage}{\textwidth}
\captionsetup{width=\linewidth}
\centering\tablefont
\renewcommand{\arraystretch}{1.10}
\setlength{\tabcolsep}{1.3pt}
\captionsetup{width=\linewidth,position=top}
\caption{Entity Continuity. P-F1 and P-R denote Point-F1@0.5 and Point-Recall@0.5. Underlined best baselines define red gains; bold marks best results. Tracking adaptations are trained for SceneForesee.}
\label{tab:trajectory}
\begin{tabularx}{\linewidth}{L{0.18\linewidth}*{9}{Y}}
\toprule
\cellcolor{tablegroup}&\multicolumn{3}{>{\columncolor{observedshade}}c}{\textbf{Observed}}&\multicolumn{3}{>{\columncolor{futureshade}}c}{\textbf{Future}}&\multicolumn{3}{>{\columncolor{combinedshade}}c}{\textbf{Combined}}\\
\multicolumn{1}{>{\columncolor{tablegroup}}c}{\multirow{-2}{*}{\textbf{Method}}}&\cellcolor{observedshade}\textbf{P-F1}&\cellcolor{observedshade}\textbf{P-R}&\cellcolor{observedshade}\textbf{mIoU}&\cellcolor{futureshade}\textbf{P-F1}&\cellcolor{futureshade}\textbf{P-R}&\cellcolor{futureshade}\textbf{mIoU}&\cellcolor{combinedshade}\textbf{P-F1}&\cellcolor{combinedshade}\textbf{P-R}&\cellcolor{combinedshade}\textbf{mIoU}\\
\midrule
TrackFormer-style&15.59&13.94&17.38&14.46&13.11&17.08&15.05&13.53&17.24\\
MeMOTR-style&15.62&13.71&16.55&15.00&13.44&16.43&15.35&13.60&16.50\\
LTrack-style&14.59&13.19&16.13&12.78&11.60&15.58&13.72&12.42&15.87\\
SambaMOTR&17.46&14.98&18.94&\underline{15.61}&14.16&16.61&15.28&14.16&17.84\\
Qwen3.5-9B&\underline{20.55}&\underline{19.72}&\underline{21.58}&15.49&\underline{14.78}&\underline{17.55}&\underline{18.16}&\underline{17.35}&\underline{19.65}\\
\rowcolor{oursrow}\textbf{\method{} (Ours)}&\bestgain{30.54}{+9.99}&\bestgain{31.10}{+11.38}&\bestgain{30.13}{+8.55}&\bestgain{26.53}{+10.92}&\bestgain{27.47}{+12.69}&\bestgain{26.77}{+9.22}&\bestgain{28.61}{+10.45}&\bestgain{29.32}{+11.97}&\bestgain{28.51}{+8.86}\\
\bottomrule
\end{tabularx}
\end{minipage}
\end{table*}

\textbf{Scene Evolution Reasoning.} Table~\ref{tab:semantic} shows Global F1 gains of 8.62 points over Qwen2.5-VL-7B for Entity, 17.24 over Qwen3.5-9B for Event, and 22.22 over Qwen3-VL-8B for Visual Evidence. Scene-F1 gains also hold when scenes receive equal weight. Figure~\ref{fig:task}(c) places these improvements alongside entity continuity across all eighteen metrics.
\begin{table*}[ht]
\centering
\begin{minipage}{\textwidth}
\captionsetup{width=\linewidth}
\centering\tablefont
\renewcommand{\arraystretch}{1.04}
\setlength{\tabcolsep}{1.2pt}
\captionsetup{width=\linewidth,position=top}
\caption{Scene Evolution Reasoning. Event and Evidence match category occurrences. Underlined best baselines define red gains; bold marks best results. OV1.5 and OV2 are 8B LLaVA-OneVision models.}
\label{tab:semantic}
\begin{tabularx}{\linewidth}{L{0.18\linewidth}*{9}{Y}}
\toprule
\cellcolor{tablegroup}&\multicolumn{3}{>{\columncolor{entityshade}}c}{\textbf{Entity}}&\multicolumn{3}{>{\columncolor{eventshade}}c}{\textbf{Event}}&\multicolumn{3}{>{\columncolor{evidenceshade}}c}{\textbf{Visual Evidence}}\\
\multicolumn{1}{>{\columncolor{tablegroup}}c}{\multirow{-2}{*}{\textbf{Method}}}&\cellcolor{entityshade}\textbf{GF1}&\cellcolor{entityshade}\textbf{SF1}&\cellcolor{entityshade}\textbf{GPR}&\cellcolor{eventshade}\textbf{GF1}&\cellcolor{eventshade}\textbf{SF1}&\cellcolor{eventshade}\textbf{GPR}&\cellcolor{evidenceshade}\textbf{GF1}&\cellcolor{evidenceshade}\textbf{SF1}&\cellcolor{evidenceshade}\textbf{GPR}\\
\midrule
\rowcolor{tablegroup}\multicolumn{10}{c}{\itshape Tracking and entity continuity methods}\\
TrackFormer-style&29.59&40.42&37.18&19.56&16.97&28.33&12.17&10.59&17.63\\
MeMOTR-style&32.45&42.63&40.31&21.03&18.13&28.72&14.18&13.29&19.37\\
LTrack-style&30.18&41.06&37.60&22.30&20.32&\underline{29.87}&15.03&14.04&20.12\\
SambaMOTR&33.91&42.25&43.02&21.87&20.06&28.10&16.46&13.46&21.93\\
\rowcolor{tablegroup}\multicolumn{10}{c}{\itshape Multimodal and video language models}\\
Qwen3.5-9B&37.45&40.73&40.14&\underline{29.85}&\underline{28.20}&\underline{29.87}&29.82&34.23&29.84\\
Qwen3-VL-8B&30.04&25.03&30.39&19.90&13.78&23.56&\underline{32.23}&23.90&\underline{38.17}\\
InternVL3.5-8B&32.92&30.53&33.92&15.86&12.56&17.20&30.13&27.32&32.66\\
Qwen2.5-VL-7B&\underline{40.46}&\underline{51.21}&\underline{43.24}&11.67&12.05&11.93&27.48&\underline{34.29}&28.10\\
LLaVA-Video-7B&35.84&38.11&36.16&10.16&8.49&10.67&20.43&22.10&21.47\\
LLaVA-OV1.5&33.69&34.81&34.63&12.79&9.58&13.42&24.11&21.04&25.30\\
LLaVA-OV2&33.54&34.14&33.75&9.21&7.86&9.81&15.90&14.79&16.94\\
\rowcolor{tablegroup}\multicolumn{10}{c}{\itshape Proposed method}\\
\rowcolor{oursrow}\textbf{\method{} (Ours)}&\bestgain{49.08}{+8.62}&\bestgain{57.33}{+6.12}&\bestgain{49.62}{+6.38}&\bestgain{47.09}{+17.24}&\bestgain{48.91}{+20.71}&\bestgain{50.11}{+20.24}&\bestgain{54.45}{+22.22}&\bestgain{57.77}{+23.48}&\bestgain{57.93}{+19.76}\\
\bottomrule
\end{tabularx}
\end{minipage}
\end{table*}

\Needspace{9\baselineskip}
\begin{wraptable}{r}{\halffloatwidth}
\vspace{-\intextsep}
\centering\compacttablefont
\renewcommand{\arraystretch}{1.12}
\setlength{\tabcolsep}{2pt}
\captionsetup{width=\linewidth,position=top}
\caption{Description similarity and complete Events.}
\label{tab:description-binding}
\begin{tabularx}{\linewidth}{@{}L{0.40\linewidth}YY@{}}
\toprule
\textbf{Method}&\makecell{\textbf{BERTScore}\\\textbf{F1}}&\textbf{Micro-F1}\\
\midrule
Qwen3.5-9B&95.98&19.94\\
\rowcolor{oursrow}\textbf{\method{} (Ours)}&\textbf{96.26}&\textbf{40.76}\\
\bottomrule
\end{tabularx}
\captionsetup{width=\linewidth}
\end{wraptable}

Table~\ref{tab:description-binding} evaluates complete Events and scene descriptions. EvoSee raises micro-F1 from Qwen3.5-9B's 19.94 to 40.76 when category and all participants must match together. Although the two methods have similar BERTScore F1, EvoSee is more accurate when Event categories and participants are evaluated jointly. Appendix~\ref{app:language-results} gives present, future, and evidence description scores.
\finishwrap

\FloatBarrier
\subsection{Ablation Study and Discussion}

Table~\ref{tab:ablation} tests continuation learning and the use of predicted change. Appendix~\ref{app:extended-results} gives complete metrics and additional variants.
\begin{table*}[ht]
\centering
\begin{minipage}{\textwidth}
\captionsetup{width=\linewidth}
\centering\tablefont
\renewcommand{\arraystretch}{1.11}
\setlength{\tabcolsep}{3.2pt}
\captionsetup{width=\linewidth,position=top}
\caption{Ablations on SceneForesee. Columns report dimension means.}
\label{tab:ablation}
\begin{tabularx}{\linewidth}{C{0.025\linewidth}L{0.26\linewidth}*{6}{Y}}
\toprule
\cellcolor{tablegroup}&\cellcolor{tablegroup}&\multicolumn{3}{>{\columncolor{trajectoryshade}}c}{\textbf{Trajectory}}&\multicolumn{3}{>{\columncolor{understandingshade}}c}{\textbf{Reasoning}}\\
\multicolumn{1}{>{\columncolor{tablegroup}}c}{\multirow{-2}{*}{\textbf{ID}}}&\multicolumn{1}{>{\columncolor{tablegroup}}c}{\multirow{-2}{*}{\textbf{Variant}}}&\cellcolor{observedshade}\textbf{Observed}&\cellcolor{futureshade}\textbf{Future}&\cellcolor{combinedshade}\textbf{Combined}&\cellcolor{entityshade}\textbf{Entity}&\cellcolor{eventshade}\textbf{Event}&\cellcolor{evidenceshade}\textbf{Evidence}\\
\midrule
1&Geometry-only Assignment&27.82&21.89&24.03&44.83&41.27&43.15\\
2&w/o Evolution Refinement&29.22&18.27&22.11&48.59&42.71&50.12\\
3&Observation-only Refinement&29.10&22.60&25.93&49.36&44.52&52.41\\
4&w/o Continuation Selection&25.37&18.89&22.00&45.61&38.70&43.65\\
5&Single Hypothesis&27.86&21.88&24.16&46.98&40.67&46.47\\
6&Full Attention&28.49&24.52&25.97&49.90&44.13&50.07\\
7&w/o Visual Evidence&27.02&19.73&22.57&46.68&39.89&36.90\\
\midrule
\rowcolor{oursrow}\textbf{8}&\textbf{\method{} (Ours)}&\textbf{30.59}&\textbf{26.92}&\textbf{28.81}&\textbf{52.01}&\textbf{48.70}&\textbf{56.72}\\
\bottomrule
\end{tabularx}
\end{minipage}
\end{table*}

\textbf{Role of joint assignment.}
Rows 1 and 8 of Table~\ref{tab:ablation} compare geometric and joint assignment. With Event supervision retained, assignment from geometric error alone reduces Future and Event means by 5.03 and 7.43 points. Using both errors in Eq.~\ref{eq:hypothesislearning} improves geometric and Event predictions.

\textbf{How does predicted change inform refinement?}
Observation-only Refinement (row 3) restores the update removed in row 2 and improves future prediction. EvoSee (row 8) uses predicted change, adding 4.32, 4.18, and 4.31 points over row 3 in Future, Event, and Evidence, and 1.49 in Observed. The Observed gain supports refining entity estimates with predicted change.

\textbf{Choice of continuation.}
Against EvoSee (row 8), Single Hypothesis (row 5) reduces Future and Event means, while removing Continuation Selection (row 4) gives the lowest Combined mean. Both candidate diversity and selection from observation contribute to entity continuation.

\textbf{Direction of attention.}
Full Attention (row 6) reduces Future and Event means by 2.41 and 4.58 points relative to EvoSee (row 8). Directed attention in Eq.~\ref{eq:evolution-attention} conditions geometry on observation and Event semantics on observed and geometric states.

\textbf{Contribution of observed evidence.}
Removing Visual Evidence (row 7 vs. row 8) reduces the Evidence mean by 19.82 points, the largest drop in this dimension, and lowers Future and Event means. Observed motion cues thus support both evidence recognition and entity continuation.

\FloatBarrier
\subsection{Further Analysis}

\Needspace{10\baselineskip}
\begin{wraptable}[10]{r}{\halffloatwidth}
\vspace{-\intextsep}
\centering\compacttablefont
\renewcommand{\arraystretch}{1.16}
\setlength{\tabcolsep}{0.6pt}
\captionsetup{width=\linewidth,position=top}
\caption{Results by observed entity count.}
\label{tab:entity-count}
\begin{tabularx}{\linewidth}{@{}L{0.255\linewidth}C{0.110\linewidth}*{6}{Y}@{}}
\toprule
\multicolumn{2}{@{}>{\columncolor{tablegroup}}c}{}&\multicolumn{3}{>{\columncolor{trajectoryshade}}c}{\textbf{Trajectory}}&\multicolumn{3}{>{\columncolor{understandingshade}}c@{}}{\textbf{Reasoning}}\\
\multicolumn{1}{@{}>{\columncolor{tablegroup}}c}{\multirow{-2}{*}{\textbf{Method}}}&\cellcolor{tablegroup}\textbf{Count}&\cellcolor{observedshade}\textbf{Obs.}&\cellcolor{futureshade}\textbf{Fut.}&\cellcolor{combinedshade}\textbf{All}&\cellcolor{entityshade}\textbf{Ent.}&\cellcolor{eventshade}\textbf{Evt.}&\cellcolor{evidenceshade}\textbf{Evd.}\\
\midrule
\multirow{3}{*}{Qwen3.5-9B}&1--3&28.35&21.90&25.26&58.06&28.30&50.72\\
&4--6&22.99&17.30&20.31&53.46&28.28&24.95\\
&$\geq 7$&9.36&7.49&8.47&19.68&31.39&14.81\\
\midrule
\rowcolor{oursrow}&1--3&\textbf{44.06}&\textbf{38.96}&\textbf{41.57}&\textbf{71.50}&\textbf{52.96}&\textbf{65.12}\\
\rowcolor{oursrow}&4--6&\textbf{29.08}&\textbf{26.90}&\textbf{28.03}&\textbf{69.59}&\textbf{48.30}&\textbf{54.00}\\
\rowcolor{oursrow}\multirow{-3}{*}{\textbf{EvoSee (Ours)}}&$\geq 7$&\textbf{13.68}&\textbf{11.18}&\textbf{12.49}&\textbf{30.94}&\textbf{45.50}&\textbf{50.70}\\
\bottomrule
\end{tabularx}
\captionsetup{width=\linewidth}
\end{wraptable}

\textbf{Effect of entity count.}
Table~\ref{tab:entity-count} shows EvoSee gains in all six dimensions for every entity count group. With at least seven entities, its Entity and Event means exceed Qwen3.5-9B by 11.27 and 14.11 points, alongside improved future continuity. Evidence recognition also improves by 35.88 points. The gains persist as more identities and interactions must be distinguished. Its Event scores vary less than its trajectory scores across groups.
\finishwrap

\vspace{4pt}
\Needspace{10\baselineskip}
\begin{wraptable}{r}{\halffloatwidth}
\vspace{-\intextsep}
\centering\compacttablefont
\renewcommand{\arraystretch}{1.16}
\setlength{\tabcolsep}{0.6pt}
\captionsetup{width=\linewidth,position=top}
\caption{Results across temporal extents $H$.}
\label{tab:horizon}
\begin{tabularx}{\linewidth}{@{}L{0.255\linewidth}C{0.110\linewidth}*{6}{Y}@{}}
\toprule
\multicolumn{2}{@{}>{\columncolor{tablegroup}}c}{}&\multicolumn{3}{>{\columncolor{trajectoryshade}}c}{\textbf{Trajectory}}&\multicolumn{3}{>{\columncolor{understandingshade}}c@{}}{\textbf{Reasoning}}\\
\multicolumn{1}{@{}>{\columncolor{tablegroup}}c}{\multirow{-2}{*}{\textbf{Method}}}&\cellcolor{tablegroup}\textbf{$H$ (s)}&\cellcolor{observedshade}\textbf{Obs.}&\cellcolor{futureshade}\textbf{Fut.}&\cellcolor{combinedshade}\textbf{All}&\cellcolor{entityshade}\textbf{Ent.}&\cellcolor{eventshade}\textbf{Evt.}&\cellcolor{evidenceshade}\textbf{Evd.}\\
\midrule
\multirow{3}{*}{Qwen3.5-9B}&1--3&21.97&17.29&19.76&41.35&27.20&31.89\\
&4--6&15.78&11.75&13.82&32.50&37.21&29.41\\
&$\geq 7$&13.64&7.95&10.78&31.15&38.76&28.15\\
\midrule
\rowcolor{oursrow}&1--3&\textbf{30.46}&\textbf{27.47}&\textbf{29.04}&\textbf{53.45}&\textbf{49.02}&\textbf{57.26}\\
\rowcolor{oursrow}&4--6&\textbf{31.29}&\textbf{26.37}&\textbf{28.85}&\textbf{46.77}&\textbf{48.12}&\textbf{54.45}\\
\rowcolor{oursrow}\multirow{-3}{*}{\textbf{EvoSee (Ours)}}&$\geq 7$&\textbf{30.81}&\textbf{21.82}&\textbf{26.32}&\textbf{46.17}&\textbf{46.25}&\textbf{54.46}\\
\bottomrule
\end{tabularx}
\captionsetup{width=\linewidth}
\end{wraptable}

\textbf{Effect of temporal extent.}
Table~\ref{tab:horizon} compares equal observation and future durations, varying both the available motion history and the extent of prediction. EvoSee leads all six dimensions in every group. Its Future gain over Qwen3.5-9B is 10.18 points in the shortest group and 13.87 in the longest. Evidence gains exceed 25 points throughout, accompanying improved future localization across temporal scales. Combined scores also decline less than Qwen3.5-9B's.
\finishwrap

\Needspace{9\baselineskip}
\begin{wraptable}{r}{\halffloatwidth}
\vspace{-\intextsep}
\centering\compacttablefont
\renewcommand{\arraystretch}{1.16}
\setlength{\tabcolsep}{0.7pt}
\captionsetup{type=table,width=\linewidth,position=top}
\caption{EvoSee dimension means by source.}
\label{tab:source-summary}
\begin{tabularx}{\linewidth}{@{}L{0.27\linewidth}*{6}{Y}@{}}
\toprule
\cellcolor{tablegroup}&\multicolumn{3}{>{\columncolor{trajectoryshade}}c}{\textbf{Trajectory}}&\multicolumn{3}{>{\columncolor{understandingshade}}c@{}}{\textbf{Reasoning}}\\
\multicolumn{1}{@{}>{\columncolor{tablegroup}}c}{\multirow{-2}{*}{\textbf{Source}}}&\cellcolor{observedshade}\textbf{Obs.}&\cellcolor{futureshade}\textbf{Fut.}&\cellcolor{combinedshade}\textbf{All}&\cellcolor{entityshade}\textbf{Ent.}&\cellcolor{eventshade}\textbf{Evt.}&\cellcolor{evidenceshade}\textbf{Evd.}\\
\midrule
\rowcolor{oursrow}APIDIS&24.40&5.23&14.53&50.51&\textbf{65.63}&\textbf{81.25}\\
\rowcolor{oursrow}OVT-B&\textbf{32.27}&\textbf{29.45}&\textbf{30.88}&50.41&50.26&57.18\\
\rowcolor{oursrow}TAO&23.37&16.25&20.05&\textbf{59.92}&41.48&54.36\\
\bottomrule
\end{tabularx}
\captionsetup{width=\linewidth}
\end{wraptable}

\textbf{Variation across video sources.}
Table~\ref{tab:source-summary} shows distinct source profiles. OVT-B leads in trajectory means, TAO in Entity, and APIDIS in Event and Evidence. The source ranking thus depends on whether evaluation concerns spatial continuity, entity references, or semantic change. Appendix~\ref{app:source-results} gives scores and sample counts for each source.
\finishwrap

\begin{figure*}[!htbp]
  \centering
  \includegraphics[page=4,width=\textwidth]{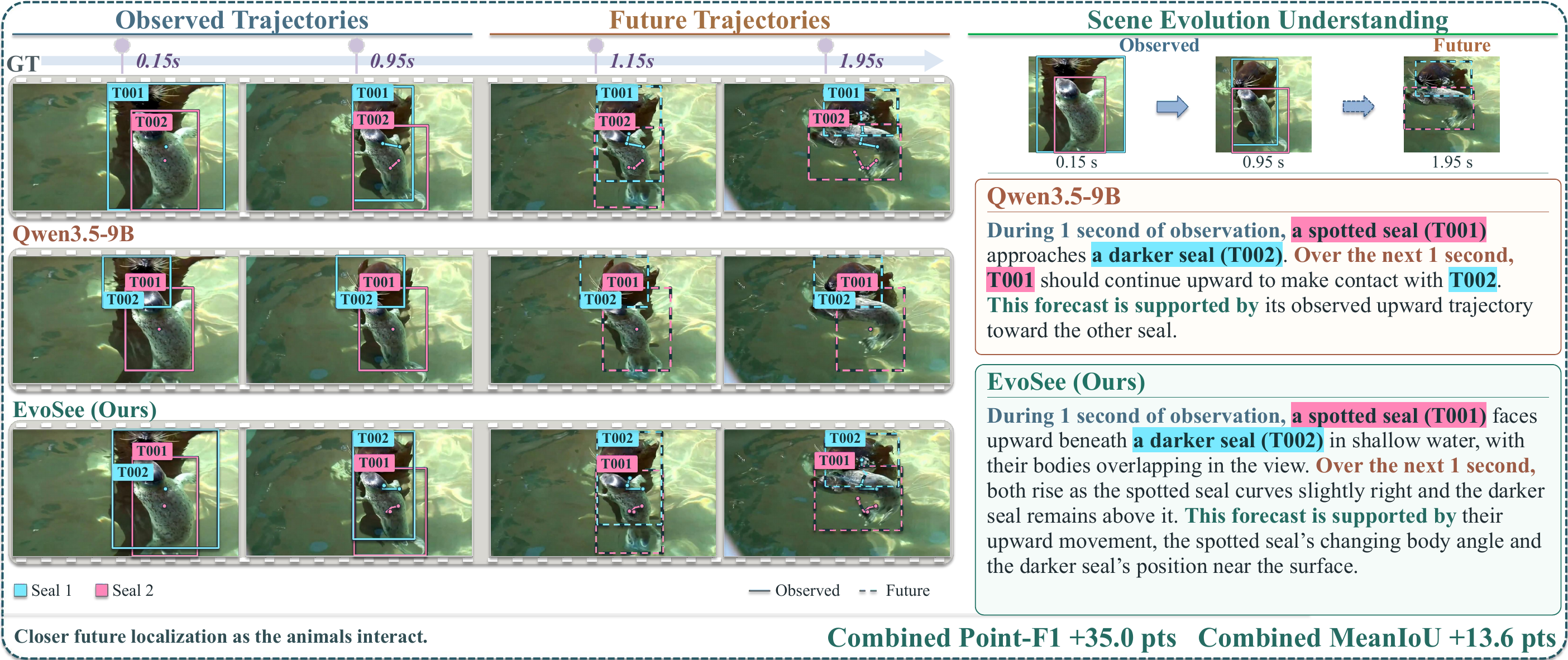}
  \captionsetup{width=\textwidth}
  \caption{Seal interaction at matched timestamps. Identity colors distinguish the seals; solid and dashed paths show observed and future centers. EvoSee retains both through changing overlap.}
  \label{fig:qualitative}
\end{figure*}

\Needspace{5\baselineskip}
\textbf{Qualitative Scene Evolution.}
Figure~\ref{fig:qualitative} shows EvoSee following both seals through changing overlap and capturing the spotted seal's curved continuation. Its description relates this movement to the other seal and their observed upward motion, connecting the predicted interaction with observed evidence. Appendix~\ref{app:qualitative} presents six further scenes.

\Needspace{5\baselineskip}
\subsection{Scene Evolution for 4D Reasoning}

EvoSee's scene representation augments Qwen3.5-9B for VQA and Sa2VA-InternVL2.5-8B for grounding on Dyn-Bench \citep{huang2026dynbench,yuan2026sa2va}. Observation-only Refinement retains each base system and substitutes observed summaries, testing the value of predicted change for these tasks.

\begin{table}[htbp]
\centering
\begin{minipage}[t]{\halffloatwidth}
\vspace{0pt}
\centering\compacttablefont
\renewcommand{\arraystretch}{1.16}
\setlength{\tabcolsep}{1.0pt}
\captionsetup{width=\linewidth,position=top}
\begin{minipage}[t][32pt][t]{\linewidth}
\caption{Dyn-Bench grounding J\&F. Red gains use the Sa2VA-InternVL2.5-8B base.}
\label{tab:dyn-ground}
\end{minipage}\par
\begin{tabularx}{\linewidth}{@{}L{0.43\linewidth}*{4}{Y}@{}}
\toprule
\multicolumn{1}{@{}>{\columncolor{tablegroup}[0pt][\tabcolsep]}c}{\textbf{Method}}&\cellcolor{trajectoryshade}\textbf{Overall}&\cellcolor{trajectoryshade}\textbf{Inter.}&\cellcolor{trajectoryshade}\textbf{Object}&\multicolumn{1}{>{\columncolor{trajectoryshade}[\tabcolsep][0pt]}c@{}}{\textbf{Cam.}}\\
\midrule
UniPixel-7B&65.23&66.00&71.10&58.60\\
Sa2VA-Qwen2.5-VL-7B&72.80&73.40&75.90&69.10\\
Sa2VA-InternVL2.5-8B&75.70&76.80&80.20&70.10\\
Observation-only Refinement&77.02&77.12&80.21&73.73\\
\rowcolor{oursrow}\methodlabelsmall&\bestgain{79.28}{+3.58}&\bestgain{78.03}{+1.23}&\bestgain{80.28}{+0.08}&\bestgain{79.54}{+9.44}\\
\bottomrule
\end{tabularx}
\captionsetup{width=\linewidth}

\end{minipage}\hfill
\begin{minipage}[t]{\halffloatwidth}
\vspace{0pt}
\centering\compacttablefont
\renewcommand{\arraystretch}{1.16}
\setlength{\tabcolsep}{1.0pt}
\captionsetup{width=\linewidth,position=top}
\begin{minipage}[t][32pt][t]{\linewidth}
\caption{Dyn-Bench VQA accuracy. Red gains use the Qwen3.5-9B base. Overall aggregates all questions.}
\label{tab:dyn-vqa}
\end{minipage}\par
\begin{tabularx}{\linewidth}{@{}L{0.43\linewidth}*{4}{Y}@{}}
\toprule
\multicolumn{1}{@{}>{\columncolor{tablegroup}[0pt][\tabcolsep]}c}{\textbf{Method}}&\cellcolor{understandingshade}\textbf{Overall}&\cellcolor{understandingshade}\textbf{Inter.}&\cellcolor{understandingshade}\textbf{Object}&\multicolumn{1}{>{\columncolor{understandingshade}[\tabcolsep][0pt]}c@{}}{\textbf{Cam.}}\\
\midrule
Qwen3-VL-8B&61.4&59.00&76.20&56.00\\
DynTrace&65.8&66.33&80.53&58.63\\
Qwen3.5-9B&63.5&59.20&79.35&58.87\\
Observation-only Refinement&66.7&67.40&80.62&60.83\\
\rowcolor{oursrow}\methodlabelsmall&\bestgain{68.3}{+4.8}&\bestgain{69.80}{+10.60}&\bestgain{80.85}{+1.50}&\bestgain{61.50}{+2.63}\\
\bottomrule
\end{tabularx}
\captionsetup{width=\linewidth}

\end{minipage}
\end{table}

\textbf{Dynamic grounding.}
Table~\ref{tab:dyn-ground} shows overall J\&F gains of 3.58 points over Sa2VA-InternVL2.5-8B and 2.26 over Observation-only Refinement. Camera motion shows the largest gain over this control, 5.81 points, supporting predicted change as context for grounding under changing viewpoints.

\textbf{Dynamic question answering.}
Table~\ref{tab:dyn-vqa} shows overall gains of 4.8 points over Qwen3.5-9B and 1.6 over Observation-only Refinement, with a 2.40 point interaction gain over this control. Across VQA and grounding, predicted change most benefits interactions and changing viewpoints, while object motion scores remain close.

\section{Conclusion}
\label{sec:conclusion}

SceneForesee evaluates how persistent entities move and how their relations develop through continuity and change in the same scenes. EvoSee learns entity continuations through shared future states and joint assignment, then uses predicted change to update entity and scene representations. Results on SceneForesee and Dyn-Bench support scene evolution as both a prediction target and a dynamic representation for spatiotemporal reasoning. SceneForesee currently represents motion in the image plane and evaluates four Event categories. Extending this formulation to 3D trajectories and a broader Event vocabulary is a direction for future work.

\section*{AI Use Statement}

Generative AI tools assisted with synthetic data generation and language polishing. The authors take responsibility for the final data and manuscript.

\FloatBarrier
\bibliography{cites}
\bibliographystyle{bibstyle}

\clearpage
\appendix
\renewcommand{\tablefont}{\fontsize{9.0}{10.4}\selectfont}
\pdfbookmark[0]{Appendix}{appendix-start}
\section*{Appendix}
\setcounter{table}{0}
\setcounter{figure}{0}
\renewcommand{\thetable}{A\arabic{table}}
\renewcommand{\thefigure}{A\arabic{figure}}
\renewcommand{\theHtable}{appendix.\arabic{table}}
\renewcommand{\theHfigure}{appendix.\arabic{figure}}
\numberwithin{equation}{section}
\section{Benchmark Construction and Evaluation}
\label{app:task}
\label{app:evaluator}

\subsection{Temporal Sampling and Annotations}

The observation boundary separates the video available for inference from the future continuation used for training and evaluation. For video $V$ and a window beginning at $s$, duration $H$ defines the boundary $t_b$ and the paired observed and future intervals.
\begin{equation}
 t_b=s+H,\qquad V^{\mathrm{obs}}=V[s,s+H],\qquad V^{\mathrm{fut}}=V(s+H,s+2H],\qquad H\in\{1,\ldots,10\}\ \mathrm{s}.
\end{equation}
Videos are normalized to 5 frames per second, and candidate windows advance by $H/4$. Each observed interval supplies four uniformly sampled keyframes. The model receives the observed video and keyframes, with $H$ specified by the task context. Annotations from this interval supervise entity states and Visual Evidence, while future annotations supervise trajectories and Events. Each sample calls for one scene evolution prediction comprising observed and future trajectories, Events, Visual Evidence, and a scene description.

\paragraph{Entity states.}
Each entity annotation includes an identity within the sample, timestamped boxes in the image plane, visibility, and view. Entities that continue across the observation boundary retain this identity in their future persistence and state annotations. Every sample contains at least one continuing entity. Counting shared identities over all temporal pairs gives 1.05M entity correspondences. Counting observed and future trajectory records separately gives 2.14M records. Both totals accumulate over samples.

\paragraph{Events and Visual Evidence.}
The four Event categories, relation change, approach, occlusion, and leave, summarize entity configurations using observed motion and states at the transition to prediction or the future endpoint. Each Event records its participating identities and the requested future interval. Its paired Visual Evidence records an individual trajectory trend or pairwise relation change, together with the participating identities, observed keyframes, and an observed interval. The paired evidence annotations record associated motion in observation and supervise evidence localization and category prediction from observed frames. All 272.49K Events have paired evidence, and each of the 144.57K samples has a scene description.

\subsection{Annotation Construction and Data Splits}
\label{app:construction}

SceneForesee constructs annotations around identities shared across each temporal pair. These identities associate timestamped boxes and visibility with continuing entities, whose trajectories determine the Event annotations. The same construction criteria apply to APIDIS, OVT-B, and TAO. Together, these sources cover sports interactions, open category entities, and varied natural motion \citep{devleeschouwer2008distributed,liang2024ovtb,dave2020tao}.

Event assignment describes relative motion, spatial overlap, and position in the field of view under these shared identities. The categories provide discrete targets for predicting the configurations represented by continuous trajectories. Approach requires the distance between entity centers to decrease by more than 10\% during observation and remain at most 1.10 times the observation endpoint distance at the future endpoint.

Occlusion uses box overlap at the temporal transition from observation to prediction, with pairwise IoU above 0.05 at the last observed or first future state. This criterion applies throughout the image, including its interior. Leave uses position relative to the image edge at the future endpoint, with the box center within the outer 10\% of either image axis. Relation change records the remaining configurations when none of these rules applies and is assigned to the first continuing identity in source order. The geometric criteria define Event supervision for the same entities and temporal pairs used in trajectory prediction.

Pairwise Events are enumerated in source identity order, with approach preceding occlusion for each pair, followed by individual leave Events. Retaining the first three records and their paired evidence defines the subset of configurations covered by Event evaluation. A sample is retained when it contains an entity annotated across the observation boundary, an Event assigned to the requested continuation, and its evidence in observation.

Visual Evidence associates each Event with observed motion. Its category follows the Event's participant structure, assigning relation change to pairwise Events and trajectory trend to individual Events. The evidence description uses the first observed motion segment containing a participating identity. If no segment shares an Event participant, it uses the first available observed segment as scene context. If no segment is available, the observed interval supplies the description and time span. The resulting evidence record carries the Event participants and observed keyframes, with its temporal extent confined to observation.

Construction included LLM assessment and manual inspection of the generated annotations. The train, validation, and test sets contain 116,681, 14,259, and 13,626 samples, respectively. Partitions use source videos, with disjoint chronological episodes defining the APIDIS partitions, so overlapping windows remain together. Each source contributes the windows whose continuing entities and Events meet the common construction criteria. Table~\ref{tab:scale} reports the resulting scale, counting units, and Event distribution.

\begin{table*}[t]
\centering
\begin{minipage}{\textwidth}
\captionsetup{width=\linewidth}
\centering\tablefont
\renewcommand{\arraystretch}{1.08}
\captionsetup{width=\linewidth,position=top}
\caption{Scale and Event distribution of \benchmark{}. Entity correspondences count identities shared across paired windows within each sample. Trajectory records count observed and future windows separately. Each Event has one associated Visual Evidence annotation. Val denotes validation.}
\label{tab:scale}
\begin{tabularx}{\linewidth}{@{}ZY@{}}
\toprule
\multicolumn{1}{@{}c}{\textbf{Statistic}}&\textbf{Count}\\
\midrule
Train / Val / Test&116,681 / 14,259 / 13,626\\
Total samples&144,566\\
Source videos&2,876\\
Entity correspondences&1,046,165\\
Observed trajectory records&1,098,141\\
Future trajectory records&1,046,165\\
Entity trajectory records&2,144,306\\
Event and evidence pairs&272,487\\
Scene descriptions&144,566\\
Event categories&4\\
Visual Evidence categories&2\\
\bottomrule
\end{tabularx}
\par\medskip
\begin{tabularx}{\linewidth}{@{}Z*{6}{Y}@{}}
\toprule
\multicolumn{1}{@{}c}{\textbf{Source}}&\textbf{Samples}&\makecell{\textbf{Relation}\\\textbf{change}}&\textbf{Approach}&\textbf{Occlusion}&\textbf{Leave}&\textbf{Events}\\
\midrule
APIDIS&317&41&441&311&35&828\\
TAO&26,644&7,910&12,879&19,983&6,188&46,960\\
OVT-B&117,605&43,100&58,499&110,039&13,061&224,699\\
\midrule
All sources&144,566&51,051&71,819&130,333&19,284&272,487\\
\bottomrule
\end{tabularx}
\end{minipage}
\end{table*}

Table~\ref{tab:scale} shows the distribution of scene evolution in the retained annotations. Occlusion and approach are the most frequent Events, capturing spatial overlap and relative motion, while relation change and leave cover residual configurations and proximity to the image edge. Separate counts for observed trajectories and continuing identities distinguish entities present in observation from those followed into the future.

The retained scenes also vary in entity count and Event composition. Observed windows contain a mean of 7.60 entities and a median of 3, ranging from 1 to 89. Multiple Events occur in 47.72\% of samples and multiple Event categories in 27.34\%. Two entities participate in 74.19\% of Events. These distributions support analysis of continuity and change across scene densities.

\FloatBarrier

\subsection{Entity Continuity}

Let $P$ and $G$ denote all predicted and annotated identities in one sample. Trajectory overlap in the observed interval determines matched subsets $P_{\mathrm m}\subseteq P$ and $G_{\mathrm m}\subseteq G$, with a bijection $\pi:P_{\mathrm m}\rightarrow G_{\mathrm m}$. This partial matching remains fixed for future evaluation. Unmatched predictions remain in the prediction counts, and unmatched annotations remain in the reference counts. For interval $W$, let $\mathcal{A}_W$ contain point pairs aligned by $\pi$, frame, and view, and define
\begin{equation}
 m_W=\sum_{(g,p)\in\mathcal{A}_W}
 \mathbf{1}[\operatorname{IoU}(g,p)\geq0.5].
\end{equation}
Let $N_W^{\mathrm{gt}}$ and $N_W^{\mathrm{pred}}$ count annotated and predicted points in interval $W$. Point-Recall and Point-F1 normalize the matched count by these totals.
\begin{equation}
 \mathrm{PointRecall}_W=\frac{m_W}{N_W^{\mathrm{gt}}},\qquad \mathrm{PointF1}_W=\frac{2m_W}{N_W^{\mathrm{gt}}+N_W^{\mathrm{pred}}}.
\end{equation}
MeanIoU averages spatial overlap over the same annotated points.
\begin{equation}
 \mathrm{MeanIoU}_W=\frac{1}{N_W^{\mathrm{gt}}}\sum_{(g,p)\in\mathcal A_W}\operatorname{IoU}(g,p).
\end{equation}
Unmatched annotated points contribute zero to MeanIoU. The observed interval measures entity state estimation, and the future interval measures trajectory prediction under the same $\pi$. Combined scores are formed within each sample. For any of the three interval scores $M$, the paired result is
\begin{equation}
 M_{\mathrm{comb}}
 =\frac{N_{\mathrm{obs}}^{\mathrm{gt}}M_{\mathrm{obs}}
       +N_{\mathrm{fut}}^{\mathrm{gt}}M_{\mathrm{fut}}}
       {N_{\mathrm{obs}}^{\mathrm{gt}}+N_{\mathrm{fut}}^{\mathrm{gt}}}.
\end{equation}
The weights are the annotated point counts of that sample's observed and future intervals. The reported score is the arithmetic mean over test samples $\mathcal S$.
\begin{equation}
 \overline M_W=\frac{1}{|\mathcal S|}\sum_{s\in\mathcal S}M_{s,W},\qquad W\in\{\mathrm{obs},\mathrm{fut},\mathrm{comb}\}.
\end{equation}
Thus annotated point counts weight intervals within a sample, while samples receive equal weight in the final result. A score with a zero denominator is assigned zero.

\subsection{Scene Evolution Reasoning}

Scene Evolution Reasoning evaluates entity references and the categories of Events and Visual Evidence. Entity scoring extracts the identities referenced in predicted and reference descriptions, maps predictions through $\pi$, and compares the resulting sets. Event and Visual Evidence scoring compares category multisets within each scene, retaining repeated occurrences. For dimension $d$ in scene $s$, let $\mathcal P_s^d$ and $\mathcal G_s^d$ denote predicted and annotated collections, and $t_s^d$ their intersection count, with multiplicity for Event and Visual Evidence. Global precision and recall pool these counts across scenes.
\begin{equation}
 P_g^d=\frac{\sum_s t_s^d}{\sum_s|\mathcal P_s^d|},
 \qquad
 R_g^d=\frac{\sum_s t_s^d}{\sum_s|\mathcal G_s^d|}.
\end{equation}
Global F1 summarizes matching counts across the test set.
\begin{equation}
 \operatorname{GlobalF1}^d=\frac{2\sum_s t_s^d}{\sum_s|\mathcal P_s^d|+\sum_s|\mathcal G_s^d|}.
\end{equation}
Scene-F1 assigns equal weight to each scene.
\begin{equation}
 \operatorname{SceneF1}^d=\frac{1}{|\mathcal S|}\sum_{s\in\mathcal S}\frac{2t_s^d}{|\mathcal P_s^d|+|\mathcal G_s^d|}.
\end{equation}
GPR is the geometric mean of global precision and recall.
\begin{equation}
 \operatorname{GPR}^d=\sqrt{P_g^dR_g^d}.
\end{equation}
A zero denominator gives zero. Entity, Event, and Visual Evidence collections are matched independently, measuring identity references and category multiplicities. Reported scores multiply the ratios above by 100.

Visual Evidence scores compare individual trajectory trend and pairwise relation change categories from the observed context defined in Appendix~\ref{app:construction}. They measure category recognition, without testing temporal localization or the support a segment provides for a particular Event. Complete Event matching separately tests the predicted category and its participants.

\paragraph{Exact matching of complete Events.}
Complete Event micro-F1 follows the structured matching principle used in relation extraction \citep{huguetcabot2021rebel}. A correct Event must match its category and complete participant record. The score is
\begin{equation}
 F_{1,\mathrm{micro}}=100\times\frac{2\mathrm{TP}}{2\mathrm{TP}+\mathrm{FP}+\mathrm{FN}}.
 \label{eq:event-micro-f1}
\end{equation}
TP counts matched Events, FP unmatched predictions, and FN unmatched references, pooled across test scenes.

Complete Event records contain a category and all participating identities. These identities are part of the model output before evaluation and refer to its predicted entities. Participant roles or ordering follow the annotation schema. Evaluation maps the predicted identities through the observation correspondence $\pi$ used for trajectory evaluation. An Event with any unmapped or incorrect participant cannot match a reference and remains in the prediction count. Within each scene, maximum one-to-one exact matching retains multiplicities. Each predicted and annotated record can be used at most once. Unmatched predictions and references contribute FP and FN, respectively. Counts are pooled over all test scenes before computing micro-F1; a zero denominator gives zero. The matching key comprises the Event category and all participant identities, with no additional temporal localization condition. Thus the score tests a category and its participants jointly, while Event category scores compare category occurrences alone.

\subsection{Dimension Means and Group Aggregation}

The main ablation and difficulty tables summarize six dimensions. Observed, Future, and Combined each average Point-F1, Point-Recall, and MeanIoU for the corresponding interval. Entity, Event, and Visual Evidence each average Global F1, Scene-F1, and GPR. These arithmetic means give equal weight to the three metrics within a dimension. Table~\ref{tab:ablation} averages the corresponding scores reported in Section~\ref{app:extended-results} and rounds the resulting means to two decimal places. Differences between dimension means are calculated before this final rounding. Event micro-F1 with exact matching is reported separately and does not enter these means.

For a sample group, trajectory scores and Scene-F1 are averaged across its samples. Global F1 and GPR are calculated from the group's pooled true positive, prediction, and reference counts. Combining groups therefore requires pooling these counts before computing the global semantic metrics. Overall results use the complete test population, and source results apply the same definitions to each source's retained samples. The dimension mean is taken after the three constituent scores have been computed for the group.

\subsection{Scene Description Similarity}

BERTScore F1 with RoBERTa-large evaluates similarity between the predicted and reference scene descriptions \citep{zhang2020bertscore}. The complete description covers observed motion, predicted development, and evidence. We also score the present, future, and evidence parts separately to examine which content contributes to the overall result. Appendix~\ref{app:language-results} reports these component scores.

\FloatBarrier

\section{Method Details}
\label{app:method}

EvoSee encodes observed entity states and Visual Evidence, then uses these representations to learn candidate continuations. Each candidate shares future states between geometry and Event semantics, and joint assignment determines which candidate receives future supervision. A selector learns this assignment from observation. The predicted change then enters entity refinement, consistency ranking, and scene representation.

\subsection{Entity Visual Evidence Encoding}
\label{app:representation}

Visual features from the observed video and its keyframes form memory $X_{\mathrm{obs}}$. A learned query $q_i$ attends to this memory to obtain entity state $a_i$ at the observation boundary. The same query attends to frame features $X_t$ at each valid observed time $t\in\mathcal O_i$, producing temporal entity state $h_i^t$. These two uses of cross attention are
\begin{equation}
 a_i=\operatorname{CrossAttn}(q_i,X_{\mathrm{obs}},X_{\mathrm{obs}}),\qquad h_i^t=\operatorname{CrossAttn}(q_i,X_t,X_t).
 \label{eq:observed-entity}
\end{equation}
A learned map $\psi_e$ assigns each temporal state a relevance logit $r_i^t$. Normalizing its sigmoid response over observed times gives weights $\alpha_i^t$ for aggregating the states into evidence $e_i$.
\begin{equation}
 \alpha_i^t=\frac{\sigma(r_i^t)}{\sum_{\ell\in\mathcal O_i}\sigma(r_i^\ell)},\qquad e_i=\sum_{t\in\mathcal O_i}\alpha_i^t h_i^t.
 \label{eq:visual-evidence}
\end{equation}
Here $r_i^t=\psi_e(h_i^t)$, and $\sigma$ is the sigmoid function.
Evidence intervals supervise temporal relevance, while evidence categories supervise the aggregated features. These two objectives train the representation to identify the observed motion associated with each entity.

\subsection{Geometry Event Alignment Reasoning}

GEAR uses the entity and evidence representations to initialize its geometric and semantic states. Their sum $u_i^a=a_i+e_i$ is projected into $u_i^g,u_i^s\in\mathbb R^d$. Stacking these states as observation, geometry, and semantics gives $U_i\in\mathbb R^{3\times d}$. For attention head $h$ of width $d_h$, layer normalization followed by learned projections gives queries, keys, and values.
\begin{equation}
 Q_i^h=\operatorname{LN}(U_i)W_Q^h,\qquad K_i^h=\operatorname{LN}(U_i)W_K^h,\qquad V_i^h=\operatorname{LN}(U_i)W_V^h.
\end{equation}
The mask in Eq.~\ref{eq:evolution-attention} permits a receiving state to attend to itself and the states preceding it in this order. Thus $\mathcal M_{rs}=0$ for $s\leq r$ and $\mathcal M_{rs}=-\infty$ for $s>r$. The resulting head outputs $A_i^h$ return to the state space through concatenation, learned projection $W_O$, and a residual update.
\begin{equation}
 \widetilde U_i=U_i+\operatorname{Concat}_h\!\left[\operatorname{softmax}\!\left(\frac{Q_i^h(K_i^h)^\top}{\sqrt{d_h}}+\mathcal M\right)V_i^h\right]W_O.
 \label{eq:evolution-attention-residual}
\end{equation}
The observed state receives information only from observation. The geometric state additionally uses this observed context, and the semantic state uses both observation and geometry.

The updated states use separate transformation weights to preserve these dependencies. With normalized states $Z_i=\operatorname{LN}(\widetilde U_i)$, learned matrix $W_R$ and bias $b_R$ assign weights over $E$ feedforward transformations; $\mathbf 1_3$ broadcasts the bias across the three states.
\begin{equation}
 \Pi_i=\operatorname{softmax}\!\left(Z_iW_R+\mathbf 1_3b_R^\top\right).
\end{equation}
Softmax acts over the $E$ transformations independently for each state. Writing $\Pi_{i,:,m}$ for the weights of transformation $\operatorname{FFN}_m$, the update is
\begin{equation}
 U_i'=\widetilde U_i+\sum_{m=1}^{E}\operatorname{diag}(\Pi_{i,:,m})\,\operatorname{FFN}_m(Z_i).
\end{equation}
Each transformation acts separately on the three states, preserving the dependencies defined by the attention mask. The refined geometric and semantic states jointly condition the future sequence from which candidate continuations are learned.

\label{app:joint-learning}

For $L$ future steps, this sequence is $f_i\in\mathbb R^{L\times d}$. Each of $K$ candidate continuations transforms it with embedding $d_k\in\mathbb R^d$ and transition $\phi_k$, as in Eq.~\ref{eq:hypothesis}. The vector $\mathbf 1_L$ broadcasts the embedding across future time steps. Both decoders use the resulting state $f_i^k$. The Event decoder also receives embedding $c$, which represents task context in the entity feature space. One candidate index therefore determines future boxes, visibility, persistence, view, and Event semantics.

Joint assignment compares each candidate with future annotations using box error and Event classification error. Let $\mathcal F_i$ contain valid annotated future steps, $\mathbf b_{i,t}^k\in\mathbb R^4$ be a predicted box, and $\mathbf b_{i,t}^\star$ its reference. The geometric error is
\begin{equation}
 \ell_{\mathrm{geo}}^{i,k}
 =\frac{1}{4|\mathcal F_i|}
   \sum_{t\in\mathcal F_i}
   \left\|\mathbf b_{i,t}^k-\mathbf b_{i,t}^\star\right\|_1.
\end{equation}
The Event classification error uses logits averaged over the $L$ future steps, $\bar s_{i,j}^k=L^{-1}\sum_{t=1}^L s_{i,t,j}^k$. For each reference Event of category $j$, every recorded participant $i$ receives $y_{i,j}=1$, so both identities in a pairwise Event carry its category target. These labels over $J$ output categories define the semantic error.
\begin{equation}
 \ell_{\mathrm{sem}}^{i,k}
 =-\frac{1}{J}\sum_{j=1}^J
 \left[y_{i,j}\log\sigma(\bar s_{i,j}^k)
 +(1-y_{i,j})\log\big(1-\sigma(\bar s_{i,j}^k)\big)\right].
\end{equation}
The assigned candidate minimizes $\ell_{\mathrm{geo}}^{i,k}+0.20\ell_{\mathrm{sem}}^{i,k}$, and its index $\widehat k_i$ is shared across all future supervision terms. This index also supplies the target for selecting a continuation from observation.

The selector $R$ combines entity state $a_i$, evidence $e_i$, and embedding $o_i$ of the entity's observed trajectory history. Its $K$ output logits define candidate probabilities $\boldsymbol\rho_i$ and the most probable candidate $k_i^*$.
\begin{equation}
 \boldsymbol\rho_i=\operatorname{softmax}\!\left(R(a_i+e_i+o_i)\right),\qquad k_i^*\in\argmax_k\rho_i^k.
 \label{eq:commit}
\end{equation}
The selector learns the assigned candidate by minimizing
\begin{equation}
 \mathcal L_{\mathrm{prior}}=-\frac{1}{|\mathcal I|}
 \sum_{i\in\mathcal I}\log\rho_i^{\widehat k_i},
\end{equation}
where $\mathcal I$ contains matched entities with valid future annotations. Future annotations determine the training target, while the selection distribution is inferred from observation.

Trajectory diversity is encouraged through the separation of candidate endpoints. For final predicted center $\mathbf p_i^k$, the regularizer is
\begin{equation}
 \mathcal L_{\mathrm{diversity}}
 =\frac{1}{|\mathcal I|K(K-1)}
 \sum_{i\in\mathcal I}\sum_{k\ne\ell}
 \left[m-\|\mathbf p_i^k-\mathbf p_i^\ell\|_1\right]_+,
 \qquad m=0.05.
\end{equation}
Here $[x]_+=\max(0,x)$, and coordinates are normalized to the image extent. This expression applies for $K>1$ and nonempty $\mathcal I$. With $K=1$, the diversity term is zero. The regularizer acts on endpoint separation while candidate assignment uses the complete valid trajectory and its Event prediction.

The candidate probabilities also define a weighted future state.
\begin{equation}
 \widetilde f_i=\sum_{k=1}^K\rho_i^kf_i^k.
\end{equation}
During training, ECD uses weighted predictions formed from the boxes and Event logits decoded for each candidate, as defined below. Inference uses $k_i^*\in\argmax_k\rho_i^k$ and $\widetilde f_i=f_i^{k_i^*}$. The same index determines future boxes, visibility, persistence, and Event predictions, preserving one geometric and semantic continuation for each entity. The selected Event scores express the categories associated with that entity. Complete Event records additionally specify all participants and are evaluated by joint category and identity matching in Appendix~\ref{app:evaluator}. With $K=1$, its sole future state directly defines the continuation.

\subsection{Evolution Conditioned Decoding}
\label{app:consistency}

ECD summarizes geometric and Event predictions as features for entity refinement. At inference, it uses the selected continuation, whose box at future time $t$ is $\mathbf b_{i,t}^{k_i^*}\in\mathbb R^4$ and whose Event logits are $s_{i,t}^{k_i^*}$. Temporal averaging and projection give the geometric and semantic summaries.
\begin{equation}
 \mathbf g_i=P_g\!\left(\frac{1}{L}\sum_{t=1}^L\mathbf b_{i,t}^{k_i^*}\right),\qquad \mathbf s_i=P_s\!\left(\sigma\!\left(\frac{1}{L}\sum_{t=1}^Ls_{i,t}^{k_i^*}\right)\right).
\end{equation}
Here $P_g$ and $P_s$ project into $\mathbb R^d$. During training, the corresponding boxes and logits are weighted by selection probabilities.
\begin{equation}
 \mathbf b_{i,t}=\sum_k\rho_i^k\mathbf b_{i,t}^k,\qquad s_{i,t}=\sum_k\rho_i^ks_{i,t}^k.
\end{equation}
These weighted predictions undergo the same temporal averaging and projection as the selected predictions at inference. Equation~\ref{eq:evolution-representation} combines the resulting summaries with entity state and observed evidence into evolution representation $\mathbf m_i$.

The map $\Gamma$ in Eq.~\ref{eq:evolution-representation} uses entity state and evolution representation to compute a vector gate.
\begin{equation}
 \boldsymbol\gamma_i=\sigma\!\left(W_\gamma[a_i;\mathbf m_i]+b_\gamma\right),\qquad \boldsymbol\gamma_i\in(0,1)^d.
\end{equation}
Normalization and projection give $P_f(\mathbf m_i)=W_f\operatorname{LN}(\mathbf m_i)+b_f$. The update is
\begin{equation}
 a_i^+=a_i+\boldsymbol\gamma_i\odot P_f(\mathbf m_i).
 \label{eq:futureupdate}
\end{equation}
The gate weights the contribution of predicted change to each feature dimension of the entity state. The updated state supports observed trajectory estimation and produces entity confidence $\ell_i^{\mathrm{obj}}$.

Entity retention uses feature agreement alongside this confidence. Six ordered relation pairs specify the features compared by Consistency.
\begin{equation}
 \mathcal R_i=\big((e_i,a_i),(a_i,\mathbf g_i),(\mathbf g_i,\mathbf s_i),(\mathbf s_i,e_i),(\mathbf g_i,\mathbf m_i),(\mathbf s_i,c)\big).
\end{equation}
The pairs connect evidence to entity state, entity state to geometry, geometry to semantics, and semantics to evidence. They also compare geometry with the evolution representation and semantics with task context. Prediction losses supervise the predicted content, while relation supervision distinguishes queries matched to annotated entities from unmatched queries.

For relation $r$, let $(u_{i,r},v_{i,r})$ denote the corresponding pair in $\mathcal R_i$. Its cosine distance is
\begin{equation}
 b_{i,r}=1-\frac{u_{i,r}^{\top}v_{i,r}}{\max(\|u_{i,r}\|_2,\epsilon)\max(\|v_{i,r}\|_2,\epsilon)},\qquad \epsilon=10^{-6}.
\end{equation}
The stabilizer $\epsilon$ keeps the distance defined for zero norm features. Averaging the six distances gives $B_i$.
\begin{equation}
 B_i=\frac{1}{6}\sum_{r=1}^{6}b_{i,r}.
\end{equation}
A smaller $B_i$ indicates closer feature agreement. Its geometric and semantic inputs come from weighted predictions during training and the selected candidate at inference. The same mean distance contributes to relation supervision and entity retention.

\textbf{Relation supervision.} Annotation matching divides entity queries into matched set $\mathcal I_{\mathrm m}$ and unmatched set $\mathcal I_{\mathrm u}$. Let $\bar B_{\mathrm m}$ and $\bar B_{\mathrm u}$ be their respective mean distances. The ranking loss is
\begin{equation}
 \mathcal L_{\mathrm{rank}}=\big[m+\bar B_{\mathrm m}-\bar B_{\mathrm u}\big]_+,\qquad m=0.25.
\end{equation}
Training applies this loss to the aggregate distance and to each individual relation. It favors lower distances for entities associated with annotated trajectories. The term is evaluated when both sets are nonempty and is zero otherwise. Geometric, semantic, and evidence supervision jointly train the representations on which the relation distances depend.

\paragraph{Supervision and continuation learning.}
Observed annotations supervise entity estimation, while the evidence intervals and categories supervise temporal relevance and the evidence representation. Future box and Event losses determine the shared candidate assignment and supervise its geometric and semantic outputs. The assigned index trains the selector through $\mathcal L_{\mathrm{prior}}$, and $\mathcal L_{\mathrm{diversity}}$ encourages separation among candidate endpoints. Relation ranking encourages lower feature distances for matched queries than for unmatched queries. At inference, observed video and task context determine candidate selection, entity retention, and scene representation.

Entity retention combines entity confidence with the mean relation distance.
\begin{equation}
 \kappa_i=\ell_i^{\mathrm{obj}}-\lambda_{\mathrm{cons}}B_i.
 \label{eq:entityselection}
\end{equation}
The entities with the highest scores form the retained set $\mathcal I^*$. Smaller relation distances raise an entity's rank at fixed confidence, linking the final entity set to both observed support and predicted continuation.

The evolution representations also contribute to scene reasoning. Together with entity states and evidence, they are averaged over all $N_q$ queries.
\begin{equation}
 \bar a=\frac{1}{N_q}\sum_i a_i,\qquad \bar e=\frac{1}{N_q}\sum_i e_i,\qquad \bar m=\frac{1}{N_q}\sum_i\mathbf m_i.
\end{equation}
Concatenating these averages with task context $c$ and projecting them gives the scene representation.
\begin{equation}
 h_{\mathrm{scene}}=P_{\mathrm{scene}}\!\left([\bar a;\bar e;\bar m;c]\right).
 \label{eq:scene-context}
\end{equation}
The scene representation combines observed states with predicted geometry and Event semantics across all queries. Alongside this representation, the retained set $\mathcal I^*$ supplies structured trajectories and Events with one continuation per entity, while $a_i^+$ supplies observed trajectory estimates and entity confidence. SceneForesee evaluates these spatial and semantic predictions directly. Dyn-Bench evaluates the scene representation in question answering and grounding, testing the value of learned scene evolution for 4D reasoning.

\FloatBarrier

\section{Experimental Protocols and Additional Results}
\label{app:extended-results}

The additional results examine entity continuity and recognition of change across model variants, entity counts, temporal extents, and video sources. Each of the six evaluation dimensions contains three metrics. P-F1 and P-R denote Point-F1@0.5 and Point-Recall@0.5. GF1, SF1, and GPR denote Global F1, Scene-F1, and the geometric mean of global precision and recall. Tables report scores as percentages and place group size $N$ before the metrics. Bold indicates the leading value within each comparison.

The main ablation and difficulty tables summarize these dimensions using Obs., Fut., All, Ent., Evt., and Evd. for Observed, Future, Combined, Entity, Event, and Visual Evidence. Each dimension mean averages its three constituent metrics. Complete Event micro-F1 and description BERTScore F1 are reported separately.

\subsection{Comparison Protocols}
\label{app:protocols}

\paragraph{SceneForesee comparisons.}
SceneForesee comparisons use the same test samples, observed video intervals, temporal extents, and metrics, with one scene prediction per method and sample. The TrackFormer-style, MeMOTR-style, LTrack-style, and SambaMOTR adaptations \citep{meinhardt2022trackformer,gao2023memotr,yu2023generalizing,segu2025samba} implement the cited tracking principles and are trained for this task rather than using the released checkpoints. Their outputs include trajectories and categories for both evaluations. Qwen3.5-9B \citep{qwen2026qwen35} and EvoSee predict trajectories and scene descriptions from observation. The remaining multimodal and video language models use shared trajectories from the TrackFormer-style adaptation as geometric context, including LLaVA-Video \citep{zhang2024llavavideo}, LLaVA-OneVision-1.5 \citep{an2025llavaonevision15}, and LLaVA-OneVision-2 \citep{an2026llavaonevision2}. For each method, matching its observed trajectories to annotations establishes the correspondence used for future continuity and Entity scoring.

\paragraph{Ablation variants.}
Nine ablations use the full model's training and evaluation conditions. Single Hypothesis sets $K=1$ to test the role of candidate continuations. Removing Consistency, Continuation Selection, Evolution Refinement, Entity Retention, or Visual Evidence tests feature agreement, selection from observation, refinement using predicted change, retention of continuing entities, or observed evidence, respectively. Every variant produces one scene prediction.

Observation-only Refinement retains the projection and gated residual update in ECD. It replaces the predicted geometric and Event inputs to the evolution representation with observed entity, motion, and evidence summaries of the same input dimensions. The resulting representation enters both the entity update and $h_{\mathrm{scene}}$, replacing predicted change in both paths shown in Figure~\ref{fig:method}. The candidate continuations and their prediction supervision remain in the model. This comparison tests the information used for refinement while retaining the update structure; w/o Evolution Refinement removes the update itself.

Geometry-only Assignment chooses $\widehat k_i$ using geometric error while retaining Event classification supervision with its original loss weight. Both geometric and Event predictions use this assigned index for supervision, and the same index trains the selector. Full Attention instead removes the directional constraint by setting every entry of $\mathcal M$ to zero, allowing observation, geometry, and semantics to attend to one another. Both variants retain EvoSee's candidate continuations and learning conditions.

\paragraph{Dyn-Bench transfer.}
On Dyn-Bench, scene representation $h_{\mathrm{scene}}$ carries predicted scene evolution into 4D reasoning. It augments Qwen3.5-9B for question answering and the Sa2VA-InternVL2.5-8B mask decoder for grounding. Each gain is measured against the corresponding base system. Additional comparisons include Qwen3-VL-8B and DynTrace for question answering, and UniPixel and Sa2VA for grounding \citep{huang2026dynbench,gao2026dyntrace,liu2025unipixel,yuan2026sa2va}. DynTrace uses Qwen3-VL-8B, and the published Qwen3-VL and grounding reference results follow Dyn-Bench.

Inter., Object, and Cam. denote interaction, object motion, and camera motion. VQA Overall is accuracy across all questions, while each dimension averages the accuracies of its three constituent tasks. Grounding Overall averages the three dimension J\&F scores. Gains use the corresponding overall scores of the matched base systems. Observation-only Refinement retains these respective base systems and uses the observed summaries defined above. Comparing this control with EvoSee tests the value of predicted change in the scene representation for VQA and grounding. These tasks evaluate 4D reasoning downstream of SceneForesee, whose trajectories describe the image plane over time.

\subsection{Complete Ablation Results}
Table~\ref{tab:extended-ablation-0} examines how continuation learning and refinement affect entity continuity. EvoSee leads every trajectory metric. Geometry-only Assignment retains Event supervision but reduces all future metrics, supporting the contribution of semantic error to joint assignment. Full Attention also lowers every future metric relative to directed attention.

Removing Evolution Refinement produces the largest loss in each future metric and in Combined Point-F1; Future Point-F1 falls by 8.44 points. Restoring the update with observed information recovers 4.62 points, and using predicted change adds a further 3.82 points. This additional gain extends to all three future metrics. EvoSee also improves the Observed mean to 30.59, exceeding Observation-only Refinement at 29.10 and the variant without refinement at 29.22 by 1.49 and 1.37 points. These gains support using the predicted continuation to refine entity estimates within the observed interval. Continuation Selection affects a different aspect of the prediction, with its removal producing the largest declines in Combined Point-Recall and MeanIoU. Single Hypothesis also lowers future accuracy. These comparisons distinguish the contributions of predicted change, selection from observation, and candidate continuations.

\Needspace{17\baselineskip}
\begingroup
\fontsize{9.0}{10.4}\selectfont
\setlength{\tabcolsep}{0.7pt}
\renewcommand{\arraystretch}{1.13}
\setlength{\LTcapwidth}{\textwidth}
\begin{longtable}{@{}L{.26\textwidth}*{9}{C{.077\textwidth}}@{}}
\caption{Complete ablation results for Entity Continuity.}\label{tab:extended-ablation-0}\\
\toprule
\multicolumn{1}{@{}>{\columncolor{tablegroup}}c}{}&\multicolumn{3}{>{\columncolor{observedshade}}c}{\textbf{Observed}}&\multicolumn{3}{>{\columncolor{futureshade}}c}{\textbf{Future}}&\multicolumn{3}{>{\columncolor{combinedshade}}c@{}}{\textbf{Combined}}\\
\cellcolor{tablegroup}\textbf{Variant}&\cellcolor{observedshade}\textbf{P-F1}&\cellcolor{observedshade}\textbf{P-R}&\cellcolor{observedshade}\textbf{mIoU}&\cellcolor{futureshade}\textbf{P-F1}&\cellcolor{futureshade}\textbf{P-R}&\cellcolor{futureshade}\textbf{mIoU}&\cellcolor{combinedshade}\textbf{P-F1}&\cellcolor{combinedshade}\textbf{P-R}&\cellcolor{combinedshade}\textbf{mIoU}\\
\midrule
\endfirsthead
\multicolumn{10}{@{}c@{}}{\tablename~\thetable{} (continued)}\\
\toprule
\multicolumn{1}{@{}>{\columncolor{tablegroup}}c}{}&\multicolumn{3}{>{\columncolor{observedshade}}c}{\textbf{Observed}}&\multicolumn{3}{>{\columncolor{futureshade}}c}{\textbf{Future}}&\multicolumn{3}{>{\columncolor{combinedshade}}c@{}}{\textbf{Combined}}\\
\cellcolor{tablegroup}\textbf{Variant}&\cellcolor{observedshade}\textbf{P-F1}&\cellcolor{observedshade}\textbf{P-R}&\cellcolor{observedshade}\textbf{mIoU}&\cellcolor{futureshade}\textbf{P-F1}&\cellcolor{futureshade}\textbf{P-R}&\cellcolor{futureshade}\textbf{mIoU}&\cellcolor{combinedshade}\textbf{P-F1}&\cellcolor{combinedshade}\textbf{P-R}&\cellcolor{combinedshade}\textbf{mIoU}\\
\midrule
\endhead
\midrule
\multicolumn{10}{r@{}}{Continued on next page}\\
\endfoot
\bottomrule
\endlastfoot
Qwen3.5-9B&20.55&19.72&21.58&15.49&14.78&17.55&18.16&17.35&19.65\\
\midrule
Geometry-only Assignment&26.93&28.48&28.05&22.39&21.24&22.05&24.04&23.09&24.96\\
w/o Evolution Refinement&29.77&28.67&29.23&18.09&17.03&19.68&21.44&21.63&23.25\\
Observation-only Refinement&29.31&28.96&29.02&22.71&22.36&22.74&25.74&25.88&26.17\\
w/o Continuation Selection&25.49&25.46&25.17&18.74&17.24&20.69&22.63&21.22&22.16\\
Single Hypothesis&28.65&26.56&28.38&21.39&21.68&22.58&24.55&22.85&25.08\\
Full Attention&28.00&29.51&27.95&24.32&24.03&25.20&25.07&26.36&26.49\\
w/o Visual Evidence&26.10&27.97&26.98&18.88&19.78&20.54&22.85&21.46&23.39\\
w/o Consistency&27.43&24.73&27.36&22.01&22.14&21.80&24.01&22.26&24.23\\
w/o Entity Retention&28.58&28.13&27.77&22.23&22.34&23.66&22.96&24.82&24.23\\
\midrule
\rowcolor{oursrow}\textbf{\method{} (Ours)}&\textbf{30.54}&\textbf{31.10}&\textbf{30.13}&\textbf{26.53}&\textbf{27.47}&\textbf{26.77}&\textbf{28.61}&\textbf{29.32}&\textbf{28.51}\\
\end{longtable}
\endgroup

Table~\ref{tab:extended-ablation-9} shows that the same ablations also affect Event prediction. Geometry-only Assignment lowers all Event scores, including Global F1 by 6.16 points. The concurrent trajectory and Event declines support using both errors in joint assignment. Full Attention also reduces all three Event metrics.

Removing Entity Retention produces the largest decline in all Entity metrics. Removing Consistency most affects Event Global F1 and GPR, whereas removing Continuation Selection most affects Event Scene-F1. Removing Visual Evidence produces the largest losses in evidence category recognition, including 19.42 points in Global F1, and also lowers Event scores. Observation-only Refinement improves Event and Visual Evidence scores over removing the update. Using predicted change further improves their Global F1 scores by 4.23 and 4.14 points, respectively. EvoSee leads all nine semantic metrics.

\Needspace{17\baselineskip}
\begingroup
\fontsize{9.0}{10.4}\selectfont
\setlength{\tabcolsep}{0.7pt}
\renewcommand{\arraystretch}{1.13}
\setlength{\LTcapwidth}{\textwidth}
\begin{longtable}{@{}L{.26\textwidth}*{9}{C{.077\textwidth}}@{}}
\caption{Complete ablation results for Scene Evolution Reasoning.}\label{tab:extended-ablation-9}\\
\toprule
\multicolumn{1}{@{}>{\columncolor{tablegroup}}c}{}&\multicolumn{3}{>{\columncolor{entityshade}}c}{\textbf{Entity}}&\multicolumn{3}{>{\columncolor{eventshade}}c}{\textbf{Event}}&\multicolumn{3}{>{\columncolor{evidenceshade}}c@{}}{\textbf{Visual Evidence}}\\
\cellcolor{tablegroup}\textbf{Variant}&\cellcolor{entityshade}\textbf{GF1}&\cellcolor{entityshade}\textbf{SF1}&\cellcolor{entityshade}\textbf{GPR}&\cellcolor{eventshade}\textbf{GF1}&\cellcolor{eventshade}\textbf{SF1}&\cellcolor{eventshade}\textbf{GPR}&\cellcolor{evidenceshade}\textbf{GF1}&\cellcolor{evidenceshade}\textbf{SF1}&\cellcolor{evidenceshade}\textbf{GPR}\\
\midrule
\endfirsthead
\multicolumn{10}{@{}c@{}}{\tablename~\thetable{} (continued)}\\
\toprule
\multicolumn{1}{@{}>{\columncolor{tablegroup}}c}{}&\multicolumn{3}{>{\columncolor{entityshade}}c}{\textbf{Entity}}&\multicolumn{3}{>{\columncolor{eventshade}}c}{\textbf{Event}}&\multicolumn{3}{>{\columncolor{evidenceshade}}c@{}}{\textbf{Visual Evidence}}\\
\cellcolor{tablegroup}\textbf{Variant}&\cellcolor{entityshade}\textbf{GF1}&\cellcolor{entityshade}\textbf{SF1}&\cellcolor{entityshade}\textbf{GPR}&\cellcolor{eventshade}\textbf{GF1}&\cellcolor{eventshade}\textbf{SF1}&\cellcolor{eventshade}\textbf{GPR}&\cellcolor{evidenceshade}\textbf{GF1}&\cellcolor{evidenceshade}\textbf{SF1}&\cellcolor{evidenceshade}\textbf{GPR}\\
\midrule
\endhead
\midrule
\multicolumn{10}{r@{}}{Continued on next page}\\
\endfoot
\bottomrule
\endlastfoot
Qwen3.5-9B&37.45&40.73&40.14&29.85&28.20&29.87&29.82&34.23&29.84\\
\midrule
Geometry-only Assignment&41.38&48.42&44.69&40.93&40.26&42.63&41.84&43.84&43.78\\
w/o Evolution Refinement&45.50&53.61&46.67&40.63&42.10&45.40&48.43&50.31&51.62\\
Observation-only Refinement&46.82&54.11&47.15&42.86&45.03&45.67&50.31&53.18&53.74\\
w/o Continuation Selection&43.46&49.00&44.36&38.54&36.96&40.59&41.31&45.43&44.20\\
Single Hypothesis&43.88&49.88&47.18&40.15&38.79&43.06&45.17&47.41&46.83\\
Full Attention&46.33&54.97&48.39&42.92&45.59&43.87&49.16&49.81&51.24\\
w/o Visual Evidence&43.31&50.33&46.39&38.74&38.41&42.53&35.03&38.65&37.01\\
w/o Consistency&44.45&50.60&45.46&36.01&37.87&37.83&46.32&50.22&48.32\\
w/o Entity Retention&39.93&44.15&41.74&40.33&43.32&41.49&46.23&50.43&49.35\\
\midrule
\rowcolor{oursrow}\textbf{\method{} (Ours)}&\textbf{49.08}&\textbf{57.33}&\textbf{49.62}&\textbf{47.09}&\textbf{48.91}&\textbf{50.11}&\textbf{54.45}&\textbf{57.77}&\textbf{57.93}\\
\end{longtable}
\endgroup

\subsection{Entity Count}

Table~\ref{tab:extended-entity-0} shows that EvoSee improves every trajectory metric in each entity count group. Its trajectory scores decrease across the three groups, while Future MeanIoU retains a 5.90 point advantage over Qwen3.5-9B in the highest group. Gains in both point recovery and overlap show that this continuity advantage holds across scene densities.

\begingroup
\fontsize{9.0}{10.4}\selectfont
\setlength{\tabcolsep}{0.7pt}
\renewcommand{\arraystretch}{1.13}
\setlength{\LTcapwidth}{\textwidth}
\begin{longtable}{@{}L{.075\textwidth}L{.185\textwidth}C{.065\textwidth}*{9}{C{.070\textwidth}}@{}}
\caption{Entity Continuity by observed entity count.}\label{tab:extended-entity-0}\\
\toprule
\multicolumn{3}{@{}>{\columncolor{tablegroup}}c}{}&\multicolumn{3}{>{\columncolor{observedshade}}c}{\textbf{Observed}}&\multicolumn{3}{>{\columncolor{futureshade}}c}{\textbf{Future}}&\multicolumn{3}{>{\columncolor{combinedshade}}c@{}}{\textbf{Combined}}\\
\cellcolor{tablegroup}\textbf{Entities}&\cellcolor{tablegroup}\textbf{Method}&\cellcolor{tablegroup}\textbf{$N$}&\cellcolor{observedshade}\textbf{P-F1}&\cellcolor{observedshade}\textbf{P-R}&\cellcolor{observedshade}\textbf{mIoU}&\cellcolor{futureshade}\textbf{P-F1}&\cellcolor{futureshade}\textbf{P-R}&\cellcolor{futureshade}\textbf{mIoU}&\cellcolor{combinedshade}\textbf{P-F1}&\cellcolor{combinedshade}\textbf{P-R}&\cellcolor{combinedshade}\textbf{mIoU}\\
\midrule
\endfirsthead
\multicolumn{12}{@{}c@{}}{\tablename~\thetable{} (continued)}\\
\toprule
\multicolumn{3}{@{}>{\columncolor{tablegroup}}c}{}&\multicolumn{3}{>{\columncolor{observedshade}}c}{\textbf{Observed}}&\multicolumn{3}{>{\columncolor{futureshade}}c}{\textbf{Future}}&\multicolumn{3}{>{\columncolor{combinedshade}}c@{}}{\textbf{Combined}}\\
\cellcolor{tablegroup}\textbf{Entities}&\cellcolor{tablegroup}\textbf{Method}&\cellcolor{tablegroup}\textbf{$N$}&\cellcolor{observedshade}\textbf{P-F1}&\cellcolor{observedshade}\textbf{P-R}&\cellcolor{observedshade}\textbf{mIoU}&\cellcolor{futureshade}\textbf{P-F1}&\cellcolor{futureshade}\textbf{P-R}&\cellcolor{futureshade}\textbf{mIoU}&\cellcolor{combinedshade}\textbf{P-F1}&\cellcolor{combinedshade}\textbf{P-R}&\cellcolor{combinedshade}\textbf{mIoU}\\
\midrule
\endhead
\midrule
\multicolumn{12}{r@{}}{Continued on next page}\\
\endfoot
\bottomrule
\endlastfoot
1--3&Qwen3.5-9B&6,400&26.41&27.65&31.00&19.81&20.58&25.31&23.28&24.25&28.26\\*
\rowcolor{oursrow}1--3&\textbf{EvoSee (Ours)}&6,400&\textbf{43.94}&\textbf{46.65}&\textbf{41.58}&\textbf{38.32}&\textbf{41.18}&\textbf{37.38}&\textbf{41.21}&\textbf{43.95}&\textbf{39.55}\\
\noalign{\vskip 2pt}
4--6&Qwen3.5-9B&2,335&25.08&21.56&22.32&18.35&15.75&17.81&21.92&18.81&20.19\\*
\rowcolor{oursrow}4--6&\textbf{EvoSee (Ours)}&2,335&\textbf{29.69}&\textbf{28.89}&\textbf{28.65}&\textbf{27.30}&\textbf{27.23}&\textbf{26.16}&\textbf{28.55}&\textbf{28.09}&\textbf{27.45}\\
\noalign{\vskip 2pt}
$\geq7$&Qwen3.5-9B&4,891&10.71&8.47&8.91&8.48&6.73&7.27&9.65&7.64&8.12\\*
\rowcolor{oursrow}$\geq7$&\textbf{EvoSee (Ours)}&4,891&\textbf{13.41}&\textbf{11.79}&\textbf{15.84}&\textbf{10.74}&\textbf{9.63}&\textbf{13.17}&\textbf{12.14}&\textbf{10.77}&\textbf{14.57}\\
\end{longtable}
\endgroup

Table~\ref{tab:extended-entity-9} shows gains in all semantic metrics for every entity count group. EvoSee's Entity scores decrease more than its Event and Visual Evidence scores as entity count rises. Its Visual Evidence Global F1 margin over Qwen3.5-9B grows to 33.09 points in the highest group, accompanied by Scene-F1 gains. Evidence category recognition thus retains an advantage even where entity references become less accurate.

\begingroup
\fontsize{9.0}{10.4}\selectfont
\setlength{\tabcolsep}{0.7pt}
\renewcommand{\arraystretch}{1.13}
\setlength{\LTcapwidth}{\textwidth}
\begin{longtable}{@{}L{.075\textwidth}L{.185\textwidth}C{.065\textwidth}*{9}{C{.070\textwidth}}@{}}
\caption{Scene Evolution Reasoning by observed entity count.}\label{tab:extended-entity-9}\\
\toprule
\multicolumn{3}{@{}>{\columncolor{tablegroup}}c}{}&\multicolumn{3}{>{\columncolor{entityshade}}c}{\textbf{Entity}}&\multicolumn{3}{>{\columncolor{eventshade}}c}{\textbf{Event}}&\multicolumn{3}{>{\columncolor{evidenceshade}}c@{}}{\textbf{Visual Evidence}}\\
\cellcolor{tablegroup}\textbf{Entities}&\cellcolor{tablegroup}\textbf{Method}&\cellcolor{tablegroup}\textbf{$N$}&\cellcolor{entityshade}\textbf{GF1}&\cellcolor{entityshade}\textbf{SF1}&\cellcolor{entityshade}\textbf{GPR}&\cellcolor{eventshade}\textbf{GF1}&\cellcolor{eventshade}\textbf{SF1}&\cellcolor{eventshade}\textbf{GPR}&\cellcolor{evidenceshade}\textbf{GF1}&\cellcolor{evidenceshade}\textbf{SF1}&\cellcolor{evidenceshade}\textbf{GPR}\\
\midrule
\endfirsthead
\multicolumn{12}{@{}c@{}}{\tablename~\thetable{} (continued)}\\
\toprule
\multicolumn{3}{@{}>{\columncolor{tablegroup}}c}{}&\multicolumn{3}{>{\columncolor{entityshade}}c}{\textbf{Entity}}&\multicolumn{3}{>{\columncolor{eventshade}}c}{\textbf{Event}}&\multicolumn{3}{>{\columncolor{evidenceshade}}c@{}}{\textbf{Visual Evidence}}\\
\cellcolor{tablegroup}\textbf{Entities}&\cellcolor{tablegroup}\textbf{Method}&\cellcolor{tablegroup}\textbf{$N$}&\cellcolor{entityshade}\textbf{GF1}&\cellcolor{entityshade}\textbf{SF1}&\cellcolor{entityshade}\textbf{GPR}&\cellcolor{eventshade}\textbf{GF1}&\cellcolor{eventshade}\textbf{SF1}&\cellcolor{eventshade}\textbf{GPR}&\cellcolor{evidenceshade}\textbf{GF1}&\cellcolor{evidenceshade}\textbf{SF1}&\cellcolor{evidenceshade}\textbf{GPR}\\
\midrule
\endhead
\midrule
\multicolumn{12}{r@{}}{Continued on next page}\\
\endfoot
\bottomrule
\endlastfoot
1--3&Qwen3.5-9B&6,400&59.77&54.54&59.86&28.76&26.50&29.65&48.74&53.16&50.25\\*
\rowcolor{oursrow}1--3&\textbf{EvoSee (Ours)}&6,400&\textbf{70.82}&\textbf{72.68}&\textbf{71.01}&\textbf{52.44}&\textbf{53.72}&\textbf{52.73}&\textbf{64.28}&\textbf{66.44}&\textbf{64.64}\\
\noalign{\vskip 2pt}
4--6&Qwen3.5-9B&2,335&53.63&50.29&56.46&27.90&28.68&28.26&24.61&25.32&24.93\\*
\rowcolor{oursrow}4--6&\textbf{EvoSee (Ours)}&2,335&\textbf{69.33}&\textbf{69.68}&\textbf{69.75}&\textbf{46.66}&\textbf{47.57}&\textbf{50.66}&\textbf{51.81}&\textbf{53.93}&\textbf{56.26}\\
\noalign{\vskip 2pt}
$\geq7$&Qwen3.5-9B&4,891&18.57&18.11&22.35&31.57&30.19&32.42&15.16&13.71&15.57\\*
\rowcolor{oursrow}$\geq7$&\textbf{EvoSee (Ours)}&4,891&\textbf{30.10}&\textbf{31.37}&\textbf{31.36}&\textbf{43.33}&\textbf{43.25}&\textbf{49.93}&\textbf{48.25}&\textbf{48.25}&\textbf{55.59}\\
\end{longtable}
\endgroup

\paragraph{Individual entity counts.}

Individual counts resolve variation within the broad density groups. Counts above 30 are pooled into $>30$, with trajectory scores and Scene-F1 averaged over samples and Global F1 and GPR recomputed from pooled matching counts.

Table~\ref{tab:extended-entity_exact-0} shows that EvoSee leads every trajectory metric for one to four entities and for reported counts from sixteen onward, including $>30$. Qwen3.5-9B has higher Point-F1 in several intermediate groups, including five to eleven entities. The advantage therefore varies within the broad entity count groups.

\begingroup
\fontsize{9.0}{10.4}\selectfont
\setlength{\tabcolsep}{0.7pt}
\renewcommand{\arraystretch}{1.13}
\setlength{\LTcapwidth}{\textwidth}
\begin{longtable}{@{}L{.075\textwidth}L{.185\textwidth}C{.065\textwidth}*{9}{C{.070\textwidth}}@{}}
\caption{Entity Continuity at each observed entity count.}\label{tab:extended-entity_exact-0}\\
\toprule
\multicolumn{3}{@{}>{\columncolor{tablegroup}}c}{}&\multicolumn{3}{>{\columncolor{observedshade}}c}{\textbf{Observed}}&\multicolumn{3}{>{\columncolor{futureshade}}c}{\textbf{Future}}&\multicolumn{3}{>{\columncolor{combinedshade}}c@{}}{\textbf{Combined}}\\
\cellcolor{tablegroup}\textbf{Entities}&\cellcolor{tablegroup}\textbf{Method}&\cellcolor{tablegroup}\textbf{$N$}&\cellcolor{observedshade}\textbf{P-F1}&\cellcolor{observedshade}\textbf{P-R}&\cellcolor{observedshade}\textbf{mIoU}&\cellcolor{futureshade}\textbf{P-F1}&\cellcolor{futureshade}\textbf{P-R}&\cellcolor{futureshade}\textbf{mIoU}&\cellcolor{combinedshade}\textbf{P-F1}&\cellcolor{combinedshade}\textbf{P-R}&\cellcolor{combinedshade}\textbf{mIoU}\\
\midrule
\endfirsthead
\multicolumn{12}{@{}c@{}}{\tablename~\thetable{} (continued)}\\
\toprule
\multicolumn{3}{@{}>{\columncolor{tablegroup}}c}{}&\multicolumn{3}{>{\columncolor{observedshade}}c}{\textbf{Observed}}&\multicolumn{3}{>{\columncolor{futureshade}}c}{\textbf{Future}}&\multicolumn{3}{>{\columncolor{combinedshade}}c@{}}{\textbf{Combined}}\\
\cellcolor{tablegroup}\textbf{Entities}&\cellcolor{tablegroup}\textbf{Method}&\cellcolor{tablegroup}\textbf{$N$}&\cellcolor{observedshade}\textbf{P-F1}&\cellcolor{observedshade}\textbf{P-R}&\cellcolor{observedshade}\textbf{mIoU}&\cellcolor{futureshade}\textbf{P-F1}&\cellcolor{futureshade}\textbf{P-R}&\cellcolor{futureshade}\textbf{mIoU}&\cellcolor{combinedshade}\textbf{P-F1}&\cellcolor{combinedshade}\textbf{P-R}&\cellcolor{combinedshade}\textbf{mIoU}\\
\midrule
\endhead
\midrule
\multicolumn{12}{r@{}}{Continued on next page}\\
\endfoot
\bottomrule
\endlastfoot
1&Qwen3.5-9B&947&20.68&26.63&33.28&12.79&16.35&24.77&16.88&21.68&29.18\\*
\rowcolor{oursrow}1&\textbf{EvoSee (Ours)}&947&\textbf{64.73}&\textbf{70.92}&\textbf{56.06}&\textbf{53.57}&\textbf{57.74}&\textbf{47.54}&\textbf{59.29}&\textbf{64.47}&\textbf{51.89}\\
\noalign{\vskip 2pt}
2&Qwen3.5-9B&3,458&30.82&31.39&34.22&23.44&23.68&27.90&27.28&27.63&31.16\\*
\rowcolor{oursrow}2&\textbf{EvoSee (Ours)}&3,458&\textbf{46.30}&\textbf{46.96}&\textbf{39.88}&\textbf{39.41}&\textbf{40.37}&\textbf{35.48}&\textbf{42.93}&\textbf{43.69}&\textbf{37.74}\\
\noalign{\vskip 2pt}
3&Qwen3.5-9B&1,995&21.50&21.65&24.33&16.85&17.21&21.08&19.38&19.60&22.81\\*
\rowcolor{oursrow}3&\textbf{EvoSee (Ours)}&1,995&\textbf{29.97}&\textbf{34.60}&\textbf{37.66}&\textbf{29.19}&\textbf{34.72}&\textbf{35.86}&\textbf{29.65}&\textbf{34.67}&\textbf{36.82}\\
\noalign{\vskip 2pt}
4&Qwen3.5-9B&1,351&20.16&17.96&20.67&15.28&13.68&17.38&17.90&15.95&19.11\\*
\rowcolor{oursrow}4&\textbf{EvoSee (Ours)}&1,351&\textbf{36.34}&\textbf{35.07}&\textbf{33.25}&\textbf{34.64}&\textbf{34.13}&\textbf{31.41}&\textbf{35.54}&\textbf{34.63}&\textbf{32.36}\\
\noalign{\vskip 2pt}
5&Qwen3.5-9B&492&\textbf{32.68}&\textbf{27.00}&\textbf{25.08}&\textbf{22.69}&18.33&18.32&\textbf{28.03}&\textbf{22.95}&21.93\\*
\rowcolor{oursrow}5&\textbf{EvoSee (Ours)}&492&22.56&22.91&24.08&18.80&\textbf{19.45}&\textbf{20.17}&20.81&21.29&\textbf{22.25}\\
\noalign{\vskip 2pt}
6&Qwen3.5-9B&492&\textbf{31.00}&\textbf{26.02}&\textbf{24.08}&\textbf{22.43}&\textbf{18.83}&\textbf{18.49}&\textbf{26.87}&\textbf{22.53}&\textbf{21.40}\\*
\rowcolor{oursrow}6&\textbf{EvoSee (Ours)}&492&18.54&17.92&20.58&15.67&16.04&17.74&17.12&16.94&19.18\\
\noalign{\vskip 2pt}
7&Qwen3.5-9B&578&\textbf{30.34}&\textbf{24.19}&\textbf{22.88}&\textbf{24.83}&19.86&18.84&\textbf{27.72}&\textbf{22.12}&\textbf{20.95}\\*
\rowcolor{oursrow}7&\textbf{EvoSee (Ours)}&578&24.29&21.46&20.99&22.28&\textbf{19.95}&\textbf{18.92}&23.33&20.75&20.03\\
\noalign{\vskip 2pt}
8&Qwen3.5-9B&395&\textbf{27.93}&\textbf{23.66}&22.47&\textbf{22.23}&\textbf{18.96}&18.73&\textbf{25.28}&\textbf{21.45}&20.71\\*
\rowcolor{oursrow}8&\textbf{EvoSee (Ours)}&395&16.92&16.80&\textbf{22.48}&14.45&14.65&\textbf{19.61}&15.82&15.85&\textbf{21.21}\\
\noalign{\vskip 2pt}
9&Qwen3.5-9B&378&\textbf{28.06}&\textbf{23.69}&\textbf{21.96}&\textbf{21.55}&\textbf{18.09}&17.65&\textbf{24.95}&\textbf{21.00}&19.89\\*
\rowcolor{oursrow}9&\textbf{EvoSee (Ours)}&378&16.98&16.97&21.92&13.56&13.69&\textbf{19.09}&15.36&15.42&\textbf{20.59}\\
\noalign{\vskip 2pt}
10&Qwen3.5-9B&196&\textbf{15.01}&12.43&13.65&\textbf{10.36}&8.51&9.60&\textbf{12.86}&10.57&11.69\\*
\rowcolor{oursrow}10&\textbf{EvoSee (Ours)}&196&12.78&\textbf{12.80}&\textbf{17.34}&8.80&\textbf{8.93}&\textbf{12.90}&10.89&\textbf{10.95}&\textbf{15.20}\\
\noalign{\vskip 2pt}
11&Qwen3.5-9B&100&\textbf{18.46}&\textbf{13.19}&13.53&\textbf{16.41}&\textbf{11.63}&11.34&\textbf{17.47}&\textbf{12.43}&12.46\\*
\rowcolor{oursrow}11&\textbf{EvoSee (Ours)}&100&12.72&12.04&\textbf{16.74}&9.52&9.01&\textbf{12.95}&11.18&10.58&\textbf{14.92}\\
\noalign{\vskip 2pt}
12&Qwen3.5-9B&81&6.48&4.11&5.34&5.01&3.06&3.71&5.86&3.65&4.50\\*
\rowcolor{oursrow}12&\textbf{EvoSee (Ours)}&81&\textbf{10.57}&\textbf{9.92}&\textbf{14.75}&\textbf{8.61}&\textbf{8.48}&\textbf{12.80}&\textbf{9.69}&\textbf{9.29}&\textbf{13.72}\\
\noalign{\vskip 2pt}
13&Qwen3.5-9B&88&\textbf{10.46}&\textbf{6.93}&6.95&\textbf{9.04}&5.99&5.70&\textbf{9.77}&6.47&6.33\\*
\rowcolor{oursrow}13&\textbf{EvoSee (Ours)}&88&6.71&6.32&\textbf{10.37}&6.63&\textbf{6.69}&\textbf{9.03}&6.66&\textbf{6.48}&\textbf{9.68}\\
\noalign{\vskip 2pt}
14&Qwen3.5-9B&121&5.25&3.39&4.37&4.29&2.74&3.23&4.79&3.08&3.82\\*
\rowcolor{oursrow}14&\textbf{EvoSee (Ours)}&121&\textbf{7.52}&\textbf{5.91}&\textbf{9.90}&\textbf{4.88}&\textbf{3.89}&\textbf{7.25}&\textbf{6.24}&\textbf{4.93}&\textbf{8.61}\\
\noalign{\vskip 2pt}
15&Qwen3.5-9B&121&\textbf{5.81}&3.57&4.17&\textbf{4.91}&3.03&3.21&\textbf{5.38}&3.31&3.71\\*
\rowcolor{oursrow}15&\textbf{EvoSee (Ours)}&121&5.71&\textbf{5.10}&\textbf{10.10}&3.77&\textbf{3.67}&\textbf{6.50}&4.80&\textbf{4.43}&\textbf{8.41}\\
\noalign{\vskip 2pt}
16&Qwen3.5-9B&102&4.19&2.58&3.31&3.35&2.06&2.29&3.81&2.34&2.84\\*
\rowcolor{oursrow}16&\textbf{EvoSee (Ours)}&102&\textbf{6.62}&\textbf{5.35}&\textbf{10.41}&\textbf{3.90}&\textbf{3.09}&\textbf{6.11}&\textbf{5.33}&\textbf{4.28}&\textbf{8.41}\\
\noalign{\vskip 2pt}
17&Qwen3.5-9B&62&1.74&1.08&1.90&1.24&0.73&0.69&1.52&0.93&1.35\\*
\rowcolor{oursrow}17&\textbf{EvoSee (Ours)}&62&\textbf{5.49}&\textbf{4.84}&\textbf{9.25}&\textbf{2.02}&\textbf{1.80}&\textbf{4.95}&\textbf{3.87}&\textbf{3.41}&\textbf{7.23}\\
\noalign{\vskip 2pt}
18&Qwen3.5-9B&65&0.22&0.14&1.03&0.03&0.02&0.32&0.12&0.08&0.68\\*
\rowcolor{oursrow}18&\textbf{EvoSee (Ours)}&65&\textbf{5.74}&\textbf{4.82}&\textbf{9.05}&\textbf{2.69}&\textbf{2.22}&\textbf{5.04}&\textbf{4.28}&\textbf{3.57}&\textbf{7.12}\\
\noalign{\vskip 2pt}
19&Qwen3.5-9B&58&1.72&1.16&2.82&1.20&0.78&1.94&1.46&0.97&2.38\\*
\rowcolor{oursrow}19&\textbf{EvoSee (Ours)}&58&\textbf{7.07}&\textbf{6.42}&\textbf{13.12}&\textbf{5.29}&\textbf{4.85}&\textbf{8.65}&\textbf{6.21}&\textbf{5.66}&\textbf{10.97}\\
\noalign{\vskip 2pt}
20&Qwen3.5-9B&72&3.96&2.91&5.59&3.15&2.53&5.13&3.62&2.74&5.38\\*
\rowcolor{oursrow}20&\textbf{EvoSee (Ours)}&72&\textbf{14.68}&\textbf{13.13}&\textbf{20.73}&\textbf{12.38}&\textbf{11.38}&\textbf{18.15}&\textbf{13.57}&\textbf{12.27}&\textbf{19.51}\\
\noalign{\vskip 2pt}
21&Qwen3.5-9B&67&3.42&2.46&5.73&3.16&2.29&5.12&3.29&2.38&5.44\\*
\rowcolor{oursrow}21&\textbf{EvoSee (Ours)}&67&\textbf{15.22}&\textbf{13.08}&\textbf{17.95}&\textbf{11.27}&\textbf{9.92}&\textbf{15.71}&\textbf{13.33}&\textbf{11.58}&\textbf{16.93}\\
\noalign{\vskip 2pt}
22&Qwen3.5-9B&116&5.84&4.77&7.94&5.25&4.37&7.66&5.56&4.58&7.81\\*
\rowcolor{oursrow}22&\textbf{EvoSee (Ours)}&116&\textbf{19.33}&\textbf{16.49}&\textbf{23.28}&\textbf{15.49}&\textbf{13.51}&\textbf{22.13}&\textbf{17.47}&\textbf{15.05}&\textbf{22.73}\\
\noalign{\vskip 2pt}
23&Qwen3.5-9B&153&3.99&2.75&6.68&3.16&2.22&6.26&3.60&2.49&6.47\\*
\rowcolor{oursrow}23&\textbf{EvoSee (Ours)}&153&\textbf{22.42}&\textbf{19.27}&\textbf{24.49}&\textbf{17.22}&\textbf{15.04}&\textbf{22.09}&\textbf{19.88}&\textbf{17.21}&\textbf{23.33}\\
\noalign{\vskip 2pt}
24&Qwen3.5-9B&130&2.55&1.68&5.15&2.15&1.43&4.68&2.35&1.56&4.92\\*
\rowcolor{oursrow}24&\textbf{EvoSee (Ours)}&130&\textbf{25.85}&\textbf{22.07}&\textbf{24.45}&\textbf{22.02}&\textbf{19.10}&\textbf{22.41}&\textbf{23.98}&\textbf{20.62}&\textbf{23.45}\\
\noalign{\vskip 2pt}
25&Qwen3.5-9B&166&3.26&2.01&5.55&2.77&1.74&5.21&3.03&1.88&5.38\\*
\rowcolor{oursrow}25&\textbf{EvoSee (Ours)}&166&\textbf{26.55}&\textbf{22.83}&\textbf{25.50}&\textbf{23.49}&\textbf{20.22}&\textbf{22.92}&\textbf{25.06}&\textbf{21.55}&\textbf{24.23}\\
\noalign{\vskip 2pt}
26&Qwen3.5-9B&51&3.65&2.43&6.62&2.53&1.89&6.04&3.11&2.16&6.33\\*
\rowcolor{oursrow}26&\textbf{EvoSee (Ours)}&51&\textbf{22.85}&\textbf{19.43}&\textbf{21.98}&\textbf{16.55}&\textbf{14.54}&\textbf{18.39}&\textbf{19.89}&\textbf{17.13}&\textbf{20.28}\\
\noalign{\vskip 2pt}
27&Qwen3.5-9B&76&0.84&0.54&1.06&0.78&0.51&0.85&0.81&0.53&0.96\\*
\rowcolor{oursrow}27&\textbf{EvoSee (Ours)}&76&\textbf{6.25}&\textbf{4.77}&\textbf{7.34}&\textbf{5.03}&\textbf{3.95}&\textbf{6.36}&\textbf{5.67}&\textbf{4.38}&\textbf{6.87}\\
\noalign{\vskip 2pt}
28&Qwen3.5-9B&68&1.04&0.58&1.14&0.76&0.43&0.70&0.90&0.50&0.91\\*
\rowcolor{oursrow}28&\textbf{EvoSee (Ours)}&68&\textbf{11.74}&\textbf{9.02}&\textbf{12.69}&\textbf{9.32}&\textbf{7.32}&\textbf{10.92}&\textbf{10.57}&\textbf{8.20}&\textbf{11.81}\\
\noalign{\vskip 2pt}
29&Qwen3.5-9B&130&0.83&0.48&0.99&0.70&0.39&0.81&0.77&0.44&0.91\\*
\rowcolor{oursrow}29&\textbf{EvoSee (Ours)}&130&\textbf{12.79}&\textbf{9.71}&\textbf{16.64}&\textbf{11.14}&\textbf{8.64}&\textbf{15.07}&\textbf{12.03}&\textbf{9.22}&\textbf{15.89}\\
\noalign{\vskip 2pt}
30&Qwen3.5-9B&130&1.28&0.71&1.13&0.74&0.41&0.85&1.03&0.57&0.99\\*
\rowcolor{oursrow}30&\textbf{EvoSee (Ours)}&130&\textbf{11.25}&\textbf{8.66}&\textbf{14.63}&\textbf{7.49}&\textbf{5.87}&\textbf{11.44}&\textbf{9.46}&\textbf{7.33}&\textbf{13.11}\\
\noalign{\vskip 2pt}
$>30$&Qwen3.5-9B&1,387&1.22&0.67&0.93&0.74&0.41&0.66&0.99&0.54&0.80\\*
\rowcolor{oursrow}$>30$&\textbf{EvoSee (Ours)}&1,387&\textbf{6.31}&\textbf{4.74}&\textbf{8.97}&\textbf{4.11}&\textbf{3.11}&\textbf{6.35}&\textbf{5.27}&\textbf{3.97}&\textbf{7.71}\\
\end{longtable}
\endgroup

Table~\ref{tab:extended-entity_exact-9} shows that EvoSee improves Visual Evidence Global F1, Scene-F1, and GPR at every reported count. Its Entity and Event results have local reversals, including Entity GPR and Event F1. For $>30$ entities, it leads all nine metrics, with a 39.15 point Visual Evidence Global F1 gain. The advantage in evidence category recognition therefore covers more individual counts than the advantages in Entity and Event recognition.

\begingroup
\fontsize{9.0}{10.4}\selectfont
\setlength{\tabcolsep}{0.7pt}
\renewcommand{\arraystretch}{1.13}
\setlength{\LTcapwidth}{\textwidth}
\begin{longtable}{@{}L{.075\textwidth}L{.185\textwidth}C{.065\textwidth}*{9}{C{.070\textwidth}}@{}}
\caption{Scene Evolution Reasoning at each observed entity count.}\label{tab:extended-entity_exact-9}\\
\toprule
\multicolumn{3}{@{}>{\columncolor{tablegroup}}c}{}&\multicolumn{3}{>{\columncolor{entityshade}}c}{\textbf{Entity}}&\multicolumn{3}{>{\columncolor{eventshade}}c}{\textbf{Event}}&\multicolumn{3}{>{\columncolor{evidenceshade}}c@{}}{\textbf{Visual Evidence}}\\
\cellcolor{tablegroup}\textbf{Entities}&\cellcolor{tablegroup}\textbf{Method}&\cellcolor{tablegroup}\textbf{$N$}&\cellcolor{entityshade}\textbf{GF1}&\cellcolor{entityshade}\textbf{SF1}&\cellcolor{entityshade}\textbf{GPR}&\cellcolor{eventshade}\textbf{GF1}&\cellcolor{eventshade}\textbf{SF1}&\cellcolor{eventshade}\textbf{GPR}&\cellcolor{evidenceshade}\textbf{GF1}&\cellcolor{evidenceshade}\textbf{SF1}&\cellcolor{evidenceshade}\textbf{GPR}\\
\midrule
\endfirsthead
\multicolumn{12}{@{}c@{}}{\tablename~\thetable{} (continued)}\\
\toprule
\multicolumn{3}{@{}>{\columncolor{tablegroup}}c}{}&\multicolumn{3}{>{\columncolor{entityshade}}c}{\textbf{Entity}}&\multicolumn{3}{>{\columncolor{eventshade}}c}{\textbf{Event}}&\multicolumn{3}{>{\columncolor{evidenceshade}}c@{}}{\textbf{Visual Evidence}}\\
\cellcolor{tablegroup}\textbf{Entities}&\cellcolor{tablegroup}\textbf{Method}&\cellcolor{tablegroup}\textbf{$N$}&\cellcolor{entityshade}\textbf{GF1}&\cellcolor{entityshade}\textbf{SF1}&\cellcolor{entityshade}\textbf{GPR}&\cellcolor{eventshade}\textbf{GF1}&\cellcolor{eventshade}\textbf{SF1}&\cellcolor{eventshade}\textbf{GPR}&\cellcolor{evidenceshade}\textbf{GF1}&\cellcolor{evidenceshade}\textbf{SF1}&\cellcolor{evidenceshade}\textbf{GPR}\\
\midrule
\endhead
\midrule
\multicolumn{12}{r@{}}{Continued on next page}\\
\endfoot
\bottomrule
\endlastfoot
1&Qwen3.5-9B&947&58.62&62.84&61.08&36.05&31.49&37.57&71.50&79.13&74.51\\*
\rowcolor{oursrow}1&\textbf{EvoSee (Ours)}&947&\textbf{68.62}&\textbf{79.86}&\textbf{71.05}&\textbf{81.73}&\textbf{81.73}&\textbf{81.73}&\textbf{84.69}&\textbf{84.69}&\textbf{84.69}\\
\noalign{\vskip 2pt}
2&Qwen3.5-9B&3,458&58.86&51.46&59.12&29.56&26.63&31.04&51.57&54.75&54.15\\*
\rowcolor{oursrow}2&\textbf{EvoSee (Ours)}&3,458&\textbf{69.55}&\textbf{70.71}&\textbf{69.61}&\textbf{49.83}&\textbf{50.38}&\textbf{49.87}&\textbf{64.44}&\textbf{65.28}&\textbf{64.49}\\
\noalign{\vskip 2pt}
3&Qwen3.5-9B&1,995&61.09&55.94&61.37&25.00&23.89&25.25&36.58&38.07&36.94\\*
\rowcolor{oursrow}3&\textbf{EvoSee (Ours)}&1,995&\textbf{72.67}&\textbf{72.67}&\textbf{72.86}&\textbf{45.38}&\textbf{46.23}&\textbf{46.66}&\textbf{56.62}&\textbf{59.79}&\textbf{58.21}\\
\noalign{\vskip 2pt}
4&Qwen3.5-9B&1,351&50.44&46.91&54.21&32.17&34.21&32.54&22.92&23.81&23.18\\*
\rowcolor{oursrow}4&\textbf{EvoSee (Ours)}&1,351&\textbf{76.63}&\textbf{76.56}&\textbf{76.88}&\textbf{52.35}&\textbf{53.89}&\textbf{55.72}&\textbf{56.43}&\textbf{59.15}&\textbf{60.07}\\
\noalign{\vskip 2pt}
5&Qwen3.5-9B&492&57.95&56.37&59.76&23.48&21.08&23.90&29.61&30.49&30.14\\*
\rowcolor{oursrow}5&\textbf{EvoSee (Ours)}&492&\textbf{61.50}&\textbf{60.98}&\textbf{61.97}&\textbf{39.75}&\textbf{38.28}&\textbf{44.22}&\textbf{47.52}&\textbf{48.04}&\textbf{52.86}\\
\noalign{\vskip 2pt}
6&Qwen3.5-9B&492&56.06&53.46&58.19&22.95&21.10&23.23&23.49&24.31&23.77\\*
\rowcolor{oursrow}6&\textbf{EvoSee (Ours)}&492&\textbf{59.63}&\textbf{59.49}&\textbf{60.44}&\textbf{40.36}&\textbf{39.50}&\textbf{45.19}&\textbf{45.39}&\textbf{45.49}&\textbf{50.82}\\
\noalign{\vskip 2pt}
7&Qwen3.5-9B&578&41.02&36.65&44.75&18.22&16.41&18.93&17.23&16.46&17.91\\*
\rowcolor{oursrow}7&\textbf{EvoSee (Ours)}&578&\textbf{48.88}&\textbf{48.97}&\textbf{49.99}&\textbf{42.57}&\textbf{42.21}&\textbf{48.75}&\textbf{46.21}&\textbf{46.31}&\textbf{52.92}\\
\noalign{\vskip 2pt}
8&Qwen3.5-9B&395&35.38&31.53&40.28&23.64&21.76&24.25&21.17&21.00&21.71\\*
\rowcolor{oursrow}8&\textbf{EvoSee (Ours)}&395&\textbf{49.23}&\textbf{49.37}&\textbf{50.08}&\textbf{43.08}&\textbf{42.70}&\textbf{49.24}&\textbf{49.22}&\textbf{49.28}&\textbf{56.26}\\
\noalign{\vskip 2pt}
9&Qwen3.5-9B&378&33.69&31.27&36.86&23.80&21.60&24.16&21.52&21.16&21.84\\*
\rowcolor{oursrow}9&\textbf{EvoSee (Ours)}&378&\textbf{48.52}&\textbf{48.39}&\textbf{49.48}&\textbf{46.14}&\textbf{46.08}&\textbf{53.19}&\textbf{48.54}&\textbf{48.46}&\textbf{55.95}\\
\noalign{\vskip 2pt}
10&Qwen3.5-9B&196&28.99&27.12&33.76&35.06&32.87&35.81&19.50&17.87&19.92\\*
\rowcolor{oursrow}10&\textbf{EvoSee (Ours)}&196&\textbf{37.96}&\textbf{38.13}&\textbf{39.11}&\textbf{41.75}&\textbf{41.92}&\textbf{48.05}&\textbf{46.91}&\textbf{47.02}&\textbf{53.98}\\
\noalign{\vskip 2pt}
11&Qwen3.5-9B&100&38.32&36.53&\textbf{43.43}&29.30&28.03&29.74&21.48&19.90&21.81\\*
\rowcolor{oursrow}11&\textbf{EvoSee (Ours)}&100&\textbf{38.91}&\textbf{40.00}&40.17&\textbf{43.00}&\textbf{43.00}&\textbf{49.65}&\textbf{46.00}&\textbf{46.00}&\textbf{53.12}\\
\noalign{\vskip 2pt}
12&Qwen3.5-9B&81&29.01&28.31&\textbf{35.92}&30.69&27.78&31.92&14.29&13.66&14.86\\*
\rowcolor{oursrow}12&\textbf{EvoSee (Ours)}&81&\textbf{32.18}&\textbf{32.43}&33.05&\textbf{42.86}&\textbf{42.59}&\textbf{49.39}&\textbf{45.34}&\textbf{45.06}&\textbf{52.25}\\
\noalign{\vskip 2pt}
13&Qwen3.5-9B&88&\textbf{27.88}&25.92&\textbf{31.23}&29.46&26.78&29.95&20.09&17.95&20.42\\*
\rowcolor{oursrow}13&\textbf{EvoSee (Ours)}&88&26.43&\textbf{26.56}&27.04&\textbf{45.45}&\textbf{45.45}&\textbf{52.49}&\textbf{48.86}&\textbf{48.86}&\textbf{56.42}\\
\noalign{\vskip 2pt}
14&Qwen3.5-9B&121&17.19&14.53&23.05&\textbf{41.82}&40.66&42.88&11.13&9.72&11.41\\*
\rowcolor{oursrow}14&\textbf{EvoSee (Ours)}&121&\textbf{29.41}&\textbf{29.48}&\textbf{30.56}&40.99&\textbf{41.05}&\textbf{47.30}&\textbf{48.45}&\textbf{48.48}&\textbf{55.90}\\
\noalign{\vskip 2pt}
15&Qwen3.5-9B&121&21.71&19.88&29.85&36.77&34.05&37.85&6.19&5.90&6.37\\*
\rowcolor{oursrow}15&\textbf{EvoSee (Ours)}&121&\textbf{30.39}&\textbf{30.60}&\textbf{31.54}&\textbf{44.07}&\textbf{43.94}&\textbf{50.79}&\textbf{46.15}&\textbf{46.01}&\textbf{53.18}\\
\noalign{\vskip 2pt}
16&Qwen3.5-9B&102&12.58&11.47&18.63&\textbf{44.84}&\textbf{43.07}&45.87&7.94&7.25&8.12\\*
\rowcolor{oursrow}16&\textbf{EvoSee (Ours)}&102&\textbf{31.97}&\textbf{32.15}&\textbf{33.71}&39.80&39.71&\textbf{45.92}&\textbf{43.73}&\textbf{43.63}&\textbf{50.46}\\
\noalign{\vskip 2pt}
17&Qwen3.5-9B&62&7.69&6.86&12.68&\textbf{44.30}&\textbf{43.39}&45.33&8.47&7.80&8.67\\*
\rowcolor{oursrow}17&\textbf{EvoSee (Ours)}&62&\textbf{29.54}&\textbf{29.37}&\textbf{31.27}&41.13&41.13&\textbf{47.49}&\textbf{45.97}&\textbf{45.97}&\textbf{53.08}\\
\noalign{\vskip 2pt}
18&Qwen3.5-9B&65&4.73&5.58&8.37&\textbf{38.41}&\textbf{37.03}&39.12&4.27&4.36&4.35\\*
\rowcolor{oursrow}18&\textbf{EvoSee (Ours)}&65&\textbf{33.33}&\textbf{34.14}&\textbf{35.15}&35.38&35.38&\textbf{40.86}&\textbf{45.38}&\textbf{45.38}&\textbf{52.41}\\
\noalign{\vskip 2pt}
19&Qwen3.5-9B&58&6.95&6.13&12.46&38.64&37.64&40.75&6.06&6.03&6.39\\*
\rowcolor{oursrow}19&\textbf{EvoSee (Ours)}&58&\textbf{24.65}&\textbf{23.28}&\textbf{26.24}&\textbf{46.55}&\textbf{46.55}&\textbf{53.75}&\textbf{47.41}&\textbf{47.41}&\textbf{54.75}\\
\noalign{\vskip 2pt}
20&Qwen3.5-9B&72&8.45&7.32&12.63&36.00&32.78&37.03&16.00&13.80&16.46\\*
\rowcolor{oursrow}20&\textbf{EvoSee (Ours)}&72&\textbf{25.54}&\textbf{25.35}&\textbf{26.49}&\textbf{48.61}&\textbf{48.61}&\textbf{56.13}&\textbf{48.61}&\textbf{48.61}&\textbf{56.13}\\
\noalign{\vskip 2pt}
21&Qwen3.5-9B&67&12.21&11.29&17.25&30.00&27.26&31.90&9.33&8.01&9.92\\*
\rowcolor{oursrow}21&\textbf{EvoSee (Ours)}&67&\textbf{30.32}&\textbf{30.25}&\textbf{31.97}&\textbf{47.01}&\textbf{47.01}&\textbf{54.29}&\textbf{50.00}&\textbf{50.00}&\textbf{57.74}\\
\noalign{\vskip 2pt}
22&Qwen3.5-9B&116&7.49&6.92&10.81&36.14&31.87&37.27&17.89&15.78&18.45\\*
\rowcolor{oursrow}22&\textbf{EvoSee (Ours)}&116&\textbf{23.73}&\textbf{23.73}&\textbf{24.45}&\textbf{49.14}&\textbf{49.14}&\textbf{56.74}&\textbf{49.57}&\textbf{49.57}&\textbf{57.24}\\
\noalign{\vskip 2pt}
23&Qwen3.5-9B&153&5.47&4.72&8.29&34.90&30.70&35.88&20.67&17.63&21.25\\*
\rowcolor{oursrow}23&\textbf{EvoSee (Ours)}&153&\textbf{19.63}&\textbf{19.53}&\textbf{20.53}&\textbf{49.02}&\textbf{49.02}&\textbf{56.60}&\textbf{49.35}&\textbf{49.35}&\textbf{56.98}\\
\noalign{\vskip 2pt}
24&Qwen3.5-9B&130&4.57&3.79&7.05&18.78&16.31&19.59&18.45&16.26&19.25\\*
\rowcolor{oursrow}24&\textbf{EvoSee (Ours)}&130&\textbf{19.59}&\textbf{19.62}&\textbf{20.65}&\textbf{47.31}&\textbf{47.31}&\textbf{54.63}&\textbf{48.46}&\textbf{48.46}&\textbf{55.96}\\
\noalign{\vskip 2pt}
25&Qwen3.5-9B&166&3.55&3.12&5.42&30.67&27.63&31.46&18.40&16.12&18.88\\*
\rowcolor{oursrow}25&\textbf{EvoSee (Ours)}&166&\textbf{18.86}&\textbf{18.88}&\textbf{20.14}&\textbf{45.78}&\textbf{45.78}&\textbf{52.87}&\textbf{49.40}&\textbf{49.40}&\textbf{57.04}\\
\noalign{\vskip 2pt}
26&Qwen3.5-9B&51&8.22&7.44&10.84&30.04&28.04&30.72&17.39&14.38&17.79\\*
\rowcolor{oursrow}26&\textbf{EvoSee (Ours)}&51&\textbf{20.13}&\textbf{20.05}&\textbf{21.47}&\textbf{38.24}&\textbf{38.24}&\textbf{44.15}&\textbf{45.10}&\textbf{45.10}&\textbf{52.07}\\
\noalign{\vskip 2pt}
27&Qwen3.5-9B&76&4.11&3.86&5.51&33.33&34.12&34.81&16.95&14.65&17.70\\*
\rowcolor{oursrow}27&\textbf{EvoSee (Ours)}&76&\textbf{28.07}&\textbf{28.40}&\textbf{29.77}&\textbf{42.11}&\textbf{42.11}&\textbf{48.62}&\textbf{46.71}&\textbf{46.71}&\textbf{53.94}\\
\noalign{\vskip 2pt}
28&Qwen3.5-9B&68&10.59&11.68&13.44&30.43&31.47&31.58&11.18&9.02&11.60\\*
\rowcolor{oursrow}28&\textbf{EvoSee (Ours)}&68&\textbf{25.06}&\textbf{25.40}&\textbf{26.50}&\textbf{38.97}&\textbf{38.97}&\textbf{45.00}&\textbf{44.85}&\textbf{44.85}&\textbf{51.79}\\
\noalign{\vskip 2pt}
29&Qwen3.5-9B&130&8.18&7.98&9.71&31.24&32.10&31.55&18.69&16.56&18.87\\*
\rowcolor{oursrow}29&\textbf{EvoSee (Ours)}&130&\textbf{26.64}&\textbf{26.77}&\textbf{27.58}&\textbf{42.69}&\textbf{42.69}&\textbf{49.30}&\textbf{48.08}&\textbf{48.08}&\textbf{55.51}\\
\noalign{\vskip 2pt}
30&Qwen3.5-9B&130&7.39&7.06&8.99&30.34&30.51&31.02&15.17&13.10&15.51\\*
\rowcolor{oursrow}30&\textbf{EvoSee (Ours)}&130&\textbf{21.68}&\textbf{21.90}&\textbf{22.69}&\textbf{43.08}&\textbf{43.08}&\textbf{49.74}&\textbf{48.46}&\textbf{48.46}&\textbf{55.96}\\
\noalign{\vskip 2pt}
$>30$&Qwen3.5-9B&1,387&11.30&11.42&13.27&38.01&38.39&39.07&10.71&8.77&11.01\\*
\rowcolor{oursrow}$>30$&\textbf{EvoSee (Ours)}&1,387&\textbf{20.01}&\textbf{20.49}&\textbf{21.23}&\textbf{42.14}&\textbf{42.14}&\textbf{48.66}&\textbf{49.86}&\textbf{49.86}&\textbf{57.57}\\
\end{longtable}
\endgroup

\subsection{Temporal Extent}

Equal observation and future durations define each temporal group. Table~\ref{tab:extended-horizon-0} shows that EvoSee leads all trajectory metrics in every group. Its observed accuracy stays close across groups, while future accuracy decreases with longer intervals. Its Future Point-F1 advantage over Qwen3.5-9B reaches 15.14 points in the longest group, accompanied by recall and overlap gains.

\begingroup
\fontsize{9.0}{10.4}\selectfont
\setlength{\tabcolsep}{0.7pt}
\renewcommand{\arraystretch}{1.13}
\setlength{\LTcapwidth}{\textwidth}
\begin{longtable}{@{}L{.075\textwidth}L{.185\textwidth}C{.065\textwidth}*{9}{C{.070\textwidth}}@{}}
\caption{Entity Continuity across temporal extents.}\label{tab:extended-horizon-0}\\
\toprule
\multicolumn{3}{@{}>{\columncolor{tablegroup}}c}{}&\multicolumn{3}{>{\columncolor{observedshade}}c}{\textbf{Observed}}&\multicolumn{3}{>{\columncolor{futureshade}}c}{\textbf{Future}}&\multicolumn{3}{>{\columncolor{combinedshade}}c@{}}{\textbf{Combined}}\\
\cellcolor{tablegroup}\textbf{Extent}&\cellcolor{tablegroup}\textbf{Method}&\cellcolor{tablegroup}\textbf{$N$}&\cellcolor{observedshade}\textbf{P-F1}&\cellcolor{observedshade}\textbf{P-R}&\cellcolor{observedshade}\textbf{mIoU}&\cellcolor{futureshade}\textbf{P-F1}&\cellcolor{futureshade}\textbf{P-R}&\cellcolor{futureshade}\textbf{mIoU}&\cellcolor{combinedshade}\textbf{P-F1}&\cellcolor{combinedshade}\textbf{P-R}&\cellcolor{combinedshade}\textbf{mIoU}\\
\midrule
\endfirsthead
\multicolumn{12}{@{}c@{}}{\tablename~\thetable{} (continued)}\\
\toprule
\multicolumn{3}{@{}>{\columncolor{tablegroup}}c}{}&\multicolumn{3}{>{\columncolor{observedshade}}c}{\textbf{Observed}}&\multicolumn{3}{>{\columncolor{futureshade}}c}{\textbf{Future}}&\multicolumn{3}{>{\columncolor{combinedshade}}c@{}}{\textbf{Combined}}\\
\cellcolor{tablegroup}\textbf{Extent}&\cellcolor{tablegroup}\textbf{Method}&\cellcolor{tablegroup}\textbf{$N$}&\cellcolor{observedshade}\textbf{P-F1}&\cellcolor{observedshade}\textbf{P-R}&\cellcolor{observedshade}\textbf{mIoU}&\cellcolor{futureshade}\textbf{P-F1}&\cellcolor{futureshade}\textbf{P-R}&\cellcolor{futureshade}\textbf{mIoU}&\cellcolor{combinedshade}\textbf{P-F1}&\cellcolor{combinedshade}\textbf{P-R}&\cellcolor{combinedshade}\textbf{mIoU}\\
\midrule
\endhead
\midrule
\multicolumn{12}{r@{}}{Continued on next page}\\
\endfoot
\bottomrule
\endlastfoot
1--3 s&Qwen3.5-9B&11,001&22.28&21.30&22.32&17.11&16.31&18.45&19.86&18.93&20.48\\*
\rowcolor{oursrow}1--3 s&\textbf{EvoSee (Ours)}&11,001&\textbf{30.25}&\textbf{31.02}&\textbf{30.12}&\textbf{26.99}&\textbf{28.21}&\textbf{27.22}&\textbf{28.70}&\textbf{29.67}&\textbf{28.74}\\
\noalign{\vskip 2pt}
4--6 s&Qwen3.5-9B&1,616&14.08&13.95&19.32&10.21&9.90&15.15&12.21&11.96&17.29\\*
\rowcolor{oursrow}4--6 s&\textbf{EvoSee (Ours)}&1,616&\textbf{31.90}&\textbf{31.69}&\textbf{30.28}&\textbf{26.58}&\textbf{26.51}&\textbf{26.03}&\textbf{29.28}&\textbf{29.10}&\textbf{28.18}\\
\noalign{\vskip 2pt}
$\geq7$ s&Qwen3.5-9B&1,009&11.97&11.80&17.15&6.34&5.92&11.59&9.14&8.81&14.40\\*
\rowcolor{oursrow}$\geq7$ s&\textbf{EvoSee (Ours)}&1,009&\textbf{31.47}&\textbf{30.99}&\textbf{29.96}&\textbf{21.48}&\textbf{20.90}&\textbf{23.08}&\textbf{26.49}&\textbf{25.92}&\textbf{26.54}\\
\end{longtable}
\endgroup

Table~\ref{tab:extended-horizon-9} shows that EvoSee also improves all semantic metrics in every temporal group. Its Event Global F1 margins narrow with longer intervals, while Visual Evidence Global F1 margins remain above 24 points. The accompanying Scene-F1 gains show that the advantage also holds when scenes receive equal weight.

\begingroup
\fontsize{9.0}{10.4}\selectfont
\setlength{\tabcolsep}{0.7pt}
\renewcommand{\arraystretch}{1.13}
\setlength{\LTcapwidth}{\textwidth}
\begin{longtable}{@{}L{.075\textwidth}L{.185\textwidth}C{.065\textwidth}*{9}{C{.070\textwidth}}@{}}
\caption{Scene Evolution Reasoning across temporal extents.}\label{tab:extended-horizon-9}\\
\toprule
\multicolumn{3}{@{}>{\columncolor{tablegroup}}c}{}&\multicolumn{3}{>{\columncolor{entityshade}}c}{\textbf{Entity}}&\multicolumn{3}{>{\columncolor{eventshade}}c}{\textbf{Event}}&\multicolumn{3}{>{\columncolor{evidenceshade}}c@{}}{\textbf{Visual Evidence}}\\
\cellcolor{tablegroup}\textbf{Extent}&\cellcolor{tablegroup}\textbf{Method}&\cellcolor{tablegroup}\textbf{$N$}&\cellcolor{entityshade}\textbf{GF1}&\cellcolor{entityshade}\textbf{SF1}&\cellcolor{entityshade}\textbf{GPR}&\cellcolor{eventshade}\textbf{GF1}&\cellcolor{eventshade}\textbf{SF1}&\cellcolor{eventshade}\textbf{GPR}&\cellcolor{evidenceshade}\textbf{GF1}&\cellcolor{evidenceshade}\textbf{SF1}&\cellcolor{evidenceshade}\textbf{GPR}\\
\midrule
\endfirsthead
\multicolumn{12}{@{}c@{}}{\tablename~\thetable{} (continued)}\\
\toprule
\multicolumn{3}{@{}>{\columncolor{tablegroup}}c}{}&\multicolumn{3}{>{\columncolor{entityshade}}c}{\textbf{Entity}}&\multicolumn{3}{>{\columncolor{eventshade}}c}{\textbf{Event}}&\multicolumn{3}{>{\columncolor{evidenceshade}}c@{}}{\textbf{Visual Evidence}}\\
\cellcolor{tablegroup}\textbf{Extent}&\cellcolor{tablegroup}\textbf{Method}&\cellcolor{tablegroup}\textbf{$N$}&\cellcolor{entityshade}\textbf{GF1}&\cellcolor{entityshade}\textbf{SF1}&\cellcolor{entityshade}\textbf{GPR}&\cellcolor{eventshade}\textbf{GF1}&\cellcolor{eventshade}\textbf{SF1}&\cellcolor{eventshade}\textbf{GPR}&\cellcolor{evidenceshade}\textbf{GF1}&\cellcolor{evidenceshade}\textbf{SF1}&\cellcolor{evidenceshade}\textbf{GPR}\\
\midrule
\endhead
\midrule
\multicolumn{12}{r@{}}{Continued on next page}\\
\endfoot
\bottomrule
\endlastfoot
1--3 s&Qwen3.5-9B&11,001&39.90&41.97&42.17&27.68&26.21&27.71&30.49&34.66&30.51\\*
\rowcolor{oursrow}1--3 s&\textbf{EvoSee (Ours)}&11,001&\textbf{51.00}&\textbf{57.98}&\textbf{51.38}&\textbf{47.38}&\textbf{49.27}&\textbf{50.40}&\textbf{54.93}&\textbf{58.41}&\textbf{58.44}\\
\noalign{\vskip 2pt}
4--6 s&Qwen3.5-9B&1,616&28.80&35.79&32.90&37.82&35.98&37.82&27.71&32.80&27.71\\*
\rowcolor{oursrow}4--6 s&\textbf{EvoSee (Ours)}&1,616&\textbf{42.40}&\textbf{54.44}&\textbf{43.46}&\textbf{46.58}&\textbf{48.26}&\textbf{49.51}&\textbf{52.46}&\textbf{55.13}&\textbf{55.77}\\
\noalign{\vskip 2pt}
$\geq7$ s&Qwen3.5-9B&1,009&26.79&35.20&31.47&39.47&37.35&39.47&26.33&31.78&26.33\\*
\rowcolor{oursrow}$\geq7$ s&\textbf{EvoSee (Ours)}&1,009&\textbf{41.00}&\textbf{54.91}&\textbf{42.60}&\textbf{44.85}&\textbf{46.02}&\textbf{47.87}&\textbf{52.38}&\textbf{55.07}&\textbf{55.92}\\
\end{longtable}
\endgroup

\paragraph{Individual temporal extents.}

The individual temporal extents reveal variation within these groups. Table~\ref{tab:extended-horizon_exact-0} shows that EvoSee leads all trajectory metrics at all ten extents. Its observed scores vary less than its future scores, which decrease over longer intervals. Identity matching remains fixed from observation, so these gains concern continuation of the same entities at each temporal extent.

\begingroup
\fontsize{9.0}{10.4}\selectfont
\setlength{\tabcolsep}{0.7pt}
\renewcommand{\arraystretch}{1.13}
\setlength{\LTcapwidth}{\textwidth}
\begin{longtable}{@{}L{.075\textwidth}L{.185\textwidth}C{.065\textwidth}*{9}{C{.070\textwidth}}@{}}
\caption{Entity Continuity at each temporal extent.}\label{tab:extended-horizon_exact-0}\\
\toprule
\multicolumn{3}{@{}>{\columncolor{tablegroup}}c}{}&\multicolumn{3}{>{\columncolor{observedshade}}c}{\textbf{Observed}}&\multicolumn{3}{>{\columncolor{futureshade}}c}{\textbf{Future}}&\multicolumn{3}{>{\columncolor{combinedshade}}c@{}}{\textbf{Combined}}\\
\cellcolor{tablegroup}\textbf{Extent}&\cellcolor{tablegroup}\textbf{Method}&\cellcolor{tablegroup}\textbf{$N$}&\cellcolor{observedshade}\textbf{P-F1}&\cellcolor{observedshade}\textbf{P-R}&\cellcolor{observedshade}\textbf{mIoU}&\cellcolor{futureshade}\textbf{P-F1}&\cellcolor{futureshade}\textbf{P-R}&\cellcolor{futureshade}\textbf{mIoU}&\cellcolor{combinedshade}\textbf{P-F1}&\cellcolor{combinedshade}\textbf{P-R}&\cellcolor{combinedshade}\textbf{mIoU}\\
\midrule
\endfirsthead
\multicolumn{12}{@{}c@{}}{\tablename~\thetable{} (continued)}\\
\toprule
\multicolumn{3}{@{}>{\columncolor{tablegroup}}c}{}&\multicolumn{3}{>{\columncolor{observedshade}}c}{\textbf{Observed}}&\multicolumn{3}{>{\columncolor{futureshade}}c}{\textbf{Future}}&\multicolumn{3}{>{\columncolor{combinedshade}}c@{}}{\textbf{Combined}}\\
\cellcolor{tablegroup}\textbf{Extent}&\cellcolor{tablegroup}\textbf{Method}&\cellcolor{tablegroup}\textbf{$N$}&\cellcolor{observedshade}\textbf{P-F1}&\cellcolor{observedshade}\textbf{P-R}&\cellcolor{observedshade}\textbf{mIoU}&\cellcolor{futureshade}\textbf{P-F1}&\cellcolor{futureshade}\textbf{P-R}&\cellcolor{futureshade}\textbf{mIoU}&\cellcolor{combinedshade}\textbf{P-F1}&\cellcolor{combinedshade}\textbf{P-R}&\cellcolor{combinedshade}\textbf{mIoU}\\
\midrule
\endhead
\midrule
\multicolumn{12}{r@{}}{Continued on next page}\\
\endfoot
\bottomrule
\endlastfoot
1 s&Qwen3.5-9B&7,294&27.12&25.91&26.85&20.99&20.00&22.35&24.24&23.10&24.70\\*
\rowcolor{oursrow}1 s&\textbf{EvoSee (Ours)}&7,294&\textbf{30.59}&\textbf{31.50}&\textbf{30.49}&\textbf{26.83}&\textbf{28.48}&\textbf{27.60}&\textbf{28.79}&\textbf{30.04}&\textbf{29.11}\\
\noalign{\vskip 2pt}
2 s&Qwen3.5-9B&2,498&18.01&17.25&18.93&13.43&12.81&15.23&15.88&15.17&17.19\\*
\rowcolor{oursrow}2 s&\textbf{EvoSee (Ours)}&2,498&\textbf{29.53}&\textbf{30.16}&\textbf{29.43}&\textbf{27.41}&\textbf{27.79}&\textbf{26.60}&\textbf{28.56}&\textbf{29.03}&\textbf{28.12}\\
\noalign{\vskip 2pt}
3 s&Qwen3.5-9B&1,209&1.92&1.84&2.02&1.27&1.25&1.54&1.61&1.55&1.79\\*
\rowcolor{oursrow}3 s&\textbf{EvoSee (Ours)}&1,209&\textbf{29.71}&\textbf{29.91}&\textbf{29.31}&\textbf{27.08}&\textbf{27.45}&\textbf{26.15}&\textbf{28.46}&\textbf{28.71}&\textbf{27.81}\\
\noalign{\vskip 2pt}
4 s&Qwen3.5-9B&690&14.84&14.81&20.11&10.90&10.73&16.17&12.94&12.81&18.20\\*
\rowcolor{oursrow}4 s&\textbf{EvoSee (Ours)}&690&\textbf{31.75}&\textbf{31.64}&\textbf{30.51}&\textbf{28.11}&\textbf{28.15}&\textbf{26.99}&\textbf{29.98}&\textbf{29.92}&\textbf{28.79}\\
\noalign{\vskip 2pt}
5 s&Qwen3.5-9B&520&14.25&13.99&19.11&10.60&10.20&15.00&12.47&12.12&17.09\\*
\rowcolor{oursrow}5 s&\textbf{EvoSee (Ours)}&520&\textbf{31.73}&\textbf{31.38}&\textbf{29.90}&\textbf{26.31}&\textbf{26.10}&\textbf{25.61}&\textbf{29.04}&\textbf{28.72}&\textbf{27.77}\\
\noalign{\vskip 2pt}
6 s&Qwen3.5-9B&406&12.58&12.45&18.27&8.55&8.10&13.62&10.65&10.30&15.99\\*
\rowcolor{oursrow}6 s&\textbf{EvoSee (Ours)}&406&\textbf{32.36}&\textbf{32.16}&\textbf{30.39}&\textbf{24.33}&\textbf{24.25}&\textbf{24.94}&\textbf{28.38}&\textbf{28.21}&\textbf{27.69}\\
\noalign{\vskip 2pt}
7 s&Qwen3.5-9B&329&13.33&13.07&18.67&7.63&7.03&12.99&10.50&10.02&15.86\\*
\rowcolor{oursrow}7 s&\textbf{EvoSee (Ours)}&329&\textbf{32.89}&\textbf{32.49}&\textbf{30.27}&\textbf{22.91}&\textbf{22.45}&\textbf{23.90}&\textbf{27.90}&\textbf{27.44}&\textbf{27.11}\\
\noalign{\vskip 2pt}
8 s&Qwen3.5-9B&268&11.80&11.65&16.90&6.54&6.06&11.60&9.13&8.78&14.28\\*
\rowcolor{oursrow}8 s&\textbf{EvoSee (Ours)}&268&\textbf{30.84}&\textbf{30.25}&\textbf{29.66}&\textbf{21.72}&\textbf{20.93}&\textbf{23.11}&\textbf{26.33}&\textbf{25.58}&\textbf{26.41}\\
\noalign{\vskip 2pt}
9 s&Qwen3.5-9B&222&10.34&10.26&15.63&4.62&4.45&10.39&7.42&7.27&13.02\\*
\rowcolor{oursrow}9 s&\textbf{EvoSee (Ours)}&222&\textbf{31.37}&\textbf{30.77}&\textbf{29.85}&\textbf{21.36}&\textbf{20.73}&\textbf{22.61}&\textbf{26.33}&\textbf{25.65}&\textbf{26.22}\\
\noalign{\vskip 2pt}
10 s&Qwen3.5-9B&190&11.75&11.61&16.63&5.81&5.54&10.53&8.80&8.57&13.64\\*
\rowcolor{oursrow}10 s&\textbf{EvoSee (Ours)}&190&\textbf{30.03}&\textbf{29.72}&\textbf{29.98}&\textbf{18.81}&\textbf{18.40}&\textbf{22.15}&\textbf{24.47}&\textbf{24.06}&\textbf{26.13}\\
\end{longtable}
\endgroup

Table~\ref{tab:extended-horizon_exact-9} shows that EvoSee also leads every semantic metric at every temporal extent. Its Event and Visual Evidence Global F1 scores vary less than Entity Global F1, and Scene-F1 gains accompany the pooled improvements. Both continuity and semantic evaluations therefore show gains across the individual temporal settings.

\begingroup
\fontsize{9.0}{10.4}\selectfont
\setlength{\tabcolsep}{0.7pt}
\renewcommand{\arraystretch}{1.13}
\setlength{\LTcapwidth}{\textwidth}
\begin{longtable}{@{}L{.075\textwidth}L{.185\textwidth}C{.065\textwidth}*{9}{C{.070\textwidth}}@{}}
\caption{Scene Evolution Reasoning at each temporal extent.}\label{tab:extended-horizon_exact-9}\\
\toprule
\multicolumn{3}{@{}>{\columncolor{tablegroup}}c}{}&\multicolumn{3}{>{\columncolor{entityshade}}c}{\textbf{Entity}}&\multicolumn{3}{>{\columncolor{eventshade}}c}{\textbf{Event}}&\multicolumn{3}{>{\columncolor{evidenceshade}}c@{}}{\textbf{Visual Evidence}}\\
\cellcolor{tablegroup}\textbf{Extent}&\cellcolor{tablegroup}\textbf{Method}&\cellcolor{tablegroup}\textbf{$N$}&\cellcolor{entityshade}\textbf{GF1}&\cellcolor{entityshade}\textbf{SF1}&\cellcolor{entityshade}\textbf{GPR}&\cellcolor{eventshade}\textbf{GF1}&\cellcolor{eventshade}\textbf{SF1}&\cellcolor{eventshade}\textbf{GPR}&\cellcolor{evidenceshade}\textbf{GF1}&\cellcolor{evidenceshade}\textbf{SF1}&\cellcolor{evidenceshade}\textbf{GPR}\\
\midrule
\endfirsthead
\multicolumn{12}{@{}c@{}}{\tablename~\thetable{} (continued)}\\
\toprule
\multicolumn{3}{@{}>{\columncolor{tablegroup}}c}{}&\multicolumn{3}{>{\columncolor{entityshade}}c}{\textbf{Entity}}&\multicolumn{3}{>{\columncolor{eventshade}}c}{\textbf{Event}}&\multicolumn{3}{>{\columncolor{evidenceshade}}c@{}}{\textbf{Visual Evidence}}\\
\cellcolor{tablegroup}\textbf{Extent}&\cellcolor{tablegroup}\textbf{Method}&\cellcolor{tablegroup}\textbf{$N$}&\cellcolor{entityshade}\textbf{GF1}&\cellcolor{entityshade}\textbf{SF1}&\cellcolor{entityshade}\textbf{GPR}&\cellcolor{eventshade}\textbf{GF1}&\cellcolor{eventshade}\textbf{SF1}&\cellcolor{eventshade}\textbf{GPR}&\cellcolor{evidenceshade}\textbf{GF1}&\cellcolor{evidenceshade}\textbf{SF1}&\cellcolor{evidenceshade}\textbf{GPR}\\
\midrule
\endhead
\midrule
\multicolumn{12}{r@{}}{Continued on next page}\\
\endfoot
\bottomrule
\endlastfoot
1 s&Qwen3.5-9B&7,294&47.43&49.49&49.76&25.40&24.11&25.44&30.49&34.45&30.52\\*
\rowcolor{oursrow}1 s&\textbf{EvoSee (Ours)}&7,294&\textbf{52.32}&\textbf{58.62}&\textbf{52.62}&\textbf{47.56}&\textbf{49.52}&\textbf{50.64}&\textbf{54.70}&\textbf{58.15}&\textbf{58.25}\\
\noalign{\vskip 2pt}
2 s&Qwen3.5-9B&2,498&35.39&37.71&37.75&30.28&28.50&30.30&30.26&34.85&30.28\\*
\rowcolor{oursrow}2 s&\textbf{EvoSee (Ours)}&2,498&\textbf{49.53}&\textbf{57.10}&\textbf{49.99}&\textbf{47.25}&\textbf{49.10}&\textbf{50.22}&\textbf{55.62}&\textbf{59.26}&\textbf{59.11}\\
\noalign{\vskip 2pt}
3 s&Qwen3.5-9B&1,209&5.37&5.38&5.87&35.93&34.20&35.93&30.95&35.54&30.95\\*
\rowcolor{oursrow}3 s&\textbf{EvoSee (Ours)}&1,209&\textbf{46.44}&\textbf{55.98}&\textbf{47.16}&\textbf{46.55}&\textbf{48.13}&\textbf{49.36}&\textbf{54.89}&\textbf{58.16}&\textbf{58.20}\\
\noalign{\vskip 2pt}
4 s&Qwen3.5-9B&690&28.25&34.56&32.27&36.87&34.69&36.87&28.93&33.49&28.93\\*
\rowcolor{oursrow}4 s&\textbf{EvoSee (Ours)}&690&\textbf{42.72}&\textbf{54.30}&\textbf{43.64}&\textbf{46.69}&\textbf{48.43}&\textbf{49.59}&\textbf{52.45}&\textbf{54.90}&\textbf{55.71}\\
\noalign{\vskip 2pt}
5 s&Qwen3.5-9B&520&29.57&37.14&33.57&38.55&36.93&38.55&26.92&32.48&26.93\\*
\rowcolor{oursrow}5 s&\textbf{EvoSee (Ours)}&520&\textbf{42.25}&\textbf{54.36}&\textbf{43.35}&\textbf{46.51}&\textbf{48.24}&\textbf{49.44}&\textbf{52.99}&\textbf{56.09}&\textbf{56.33}\\
\noalign{\vskip 2pt}
6 s&Qwen3.5-9B&406&28.70&36.14&33.10&38.51&36.95&38.51&26.63&32.03&26.64\\*
\rowcolor{oursrow}6 s&\textbf{EvoSee (Ours)}&406&\textbf{42.08}&\textbf{54.77}&\textbf{43.33}&\textbf{46.48}&\textbf{47.99}&\textbf{49.47}&\textbf{51.82}&\textbf{54.27}&\textbf{55.16}\\
\noalign{\vskip 2pt}
7 s&Qwen3.5-9B&329&28.90&37.55&33.40&39.52&37.65&39.53&25.95&31.67&25.95\\*
\rowcolor{oursrow}7 s&\textbf{EvoSee (Ours)}&329&\textbf{41.00}&\textbf{54.51}&\textbf{42.40}&\textbf{44.00}&\textbf{45.24}&\textbf{46.94}&\textbf{51.54}&\textbf{53.85}&\textbf{54.97}\\
\noalign{\vskip 2pt}
8 s&Qwen3.5-9B&268&25.96&34.81&30.54&36.33&33.94&36.33&26.62&32.82&26.62\\*
\rowcolor{oursrow}8 s&\textbf{EvoSee (Ours)}&268&\textbf{41.10}&\textbf{54.79}&\textbf{42.67}&\textbf{45.87}&\textbf{47.01}&\textbf{48.96}&\textbf{52.67}&\textbf{55.53}&\textbf{56.22}\\
\noalign{\vskip 2pt}
9 s&Qwen3.5-9B&222&26.11&34.35&30.72&41.55&39.50&41.58&27.70&32.03&27.72\\*
\rowcolor{oursrow}9 s&\textbf{EvoSee (Ours)}&222&\textbf{40.61}&\textbf{54.79}&\textbf{42.34}&\textbf{44.80}&\textbf{46.17}&\textbf{47.83}&\textbf{52.42}&\textbf{55.26}&\textbf{55.95}\\
\noalign{\vskip 2pt}
10 s&Qwen3.5-9B&190&25.24&32.68&30.38&41.26&39.12&41.26&24.91&30.19&24.91\\*
\rowcolor{oursrow}10 s&\textbf{EvoSee (Ours)}&190&\textbf{41.28}&\textbf{55.90}&\textbf{43.16}&\textbf{44.90}&\textbf{45.79}&\textbf{48.00}&\textbf{53.40}&\textbf{56.32}&\textbf{57.09}\\
\end{longtable}
\endgroup

\subsection{Video Source}
\label{app:source-results}

The source analysis covers 32 APIDIS, 11,048 OVT-B, and 2,546 TAO test windows. Table~\ref{tab:extended-source-0} separates localization in observation from continuation into the future. OVT-B has the highest future and combined trajectory scores. APIDIS has the highest observed MeanIoU alongside lower future accuracy, and TAO also shows a decline from observation to future. Spatial continuation therefore varies across sources even when observed localization is strong.

\begingroup
\fontsize{9.0}{10.4}\selectfont
\setlength{\tabcolsep}{0.7pt}
\renewcommand{\arraystretch}{1.13}
\setlength{\LTcapwidth}{\textwidth}
\begin{longtable}{@{}L{.20\textwidth}C{.065\textwidth}*{9}{C{.077\textwidth}}@{}}
\caption{Entity Continuity across video sources for EvoSee.}\label{tab:extended-source-0}\\
\toprule
\multicolumn{2}{@{}>{\columncolor{tablegroup}}c}{}&\multicolumn{3}{>{\columncolor{observedshade}}c}{\textbf{Observed}}&\multicolumn{3}{>{\columncolor{futureshade}}c}{\textbf{Future}}&\multicolumn{3}{>{\columncolor{combinedshade}}c@{}}{\textbf{Combined}}\\
\cellcolor{tablegroup}\textbf{Source}&\cellcolor{tablegroup}\textbf{$N$}&\cellcolor{observedshade}\textbf{P-F1}&\cellcolor{observedshade}\textbf{P-R}&\cellcolor{observedshade}\textbf{mIoU}&\cellcolor{futureshade}\textbf{P-F1}&\cellcolor{futureshade}\textbf{P-R}&\cellcolor{futureshade}\textbf{mIoU}&\cellcolor{combinedshade}\textbf{P-F1}&\cellcolor{combinedshade}\textbf{P-R}&\cellcolor{combinedshade}\textbf{mIoU}\\
\midrule
\endfirsthead
\multicolumn{11}{@{}c@{}}{\tablename~\thetable{} (continued)}\\
\toprule
\multicolumn{2}{@{}>{\columncolor{tablegroup}}c}{}&\multicolumn{3}{>{\columncolor{observedshade}}c}{\textbf{Observed}}&\multicolumn{3}{>{\columncolor{futureshade}}c}{\textbf{Future}}&\multicolumn{3}{>{\columncolor{combinedshade}}c@{}}{\textbf{Combined}}\\
\cellcolor{tablegroup}\textbf{Source}&\cellcolor{tablegroup}\textbf{$N$}&\cellcolor{observedshade}\textbf{P-F1}&\cellcolor{observedshade}\textbf{P-R}&\cellcolor{observedshade}\textbf{mIoU}&\cellcolor{futureshade}\textbf{P-F1}&\cellcolor{futureshade}\textbf{P-R}&\cellcolor{futureshade}\textbf{mIoU}&\cellcolor{combinedshade}\textbf{P-F1}&\cellcolor{combinedshade}\textbf{P-R}&\cellcolor{combinedshade}\textbf{mIoU}\\
\midrule
\endhead
\midrule
\multicolumn{11}{r@{}}{Continued on next page}\\
\endfoot
\bottomrule
\endlastfoot
\rowcolor{oursrow}APIDIS&32&13.38&25.79&\textbf{34.04}&2.09&3.44&10.15&7.59&14.31&21.70\\
\rowcolor{oursrow}OVT-B&11,048&\textbf{32.97}&\textbf{32.74}&31.09&\textbf{29.55}&\textbf{30.09}&\textbf{28.70}&\textbf{31.29}&\textbf{31.42}&\textbf{29.92}\\
\rowcolor{oursrow}TAO&2,546&20.19&24.04&25.88&13.77&16.39&18.60&17.24&20.42&22.49\\
\end{longtable}
\endgroup

Table~\ref{tab:extended-source-9} reveals a different source profile for semantic prediction. TAO has the highest Entity scores, while APIDIS leads all Event and Visual Evidence metrics. On APIDIS, strong category recognition coexists with lower future geometric accuracy. Evaluating both predictions on the same source exposes this difference between recognition of change and spatial continuation.

\begingroup
\fontsize{9.0}{10.4}\selectfont
\setlength{\tabcolsep}{0.7pt}
\renewcommand{\arraystretch}{1.13}
\setlength{\LTcapwidth}{\textwidth}
\begin{longtable}{@{}L{.20\textwidth}C{.065\textwidth}*{9}{C{.077\textwidth}}@{}}
\caption{Scene Evolution Reasoning across video sources for EvoSee.}\label{tab:extended-source-9}\\
\toprule
\multicolumn{2}{@{}>{\columncolor{tablegroup}}c}{}&\multicolumn{3}{>{\columncolor{entityshade}}c}{\textbf{Entity}}&\multicolumn{3}{>{\columncolor{eventshade}}c}{\textbf{Event}}&\multicolumn{3}{>{\columncolor{evidenceshade}}c@{}}{\textbf{Visual Evidence}}\\
\cellcolor{tablegroup}\textbf{Source}&\cellcolor{tablegroup}\textbf{$N$}&\cellcolor{entityshade}\textbf{GF1}&\cellcolor{entityshade}\textbf{SF1}&\cellcolor{entityshade}\textbf{GPR}&\cellcolor{eventshade}\textbf{GF1}&\cellcolor{eventshade}\textbf{SF1}&\cellcolor{eventshade}\textbf{GPR}&\cellcolor{evidenceshade}\textbf{GF1}&\cellcolor{evidenceshade}\textbf{SF1}&\cellcolor{evidenceshade}\textbf{GPR}\\
\midrule
\endfirsthead
\multicolumn{11}{@{}c@{}}{\tablename~\thetable{} (continued)}\\
\toprule
\multicolumn{2}{@{}>{\columncolor{tablegroup}}c}{}&\multicolumn{3}{>{\columncolor{entityshade}}c}{\textbf{Entity}}&\multicolumn{3}{>{\columncolor{eventshade}}c}{\textbf{Event}}&\multicolumn{3}{>{\columncolor{evidenceshade}}c@{}}{\textbf{Visual Evidence}}\\
\cellcolor{tablegroup}\textbf{Source}&\cellcolor{tablegroup}\textbf{$N$}&\cellcolor{entityshade}\textbf{GF1}&\cellcolor{entityshade}\textbf{SF1}&\cellcolor{entityshade}\textbf{GPR}&\cellcolor{eventshade}\textbf{GF1}&\cellcolor{eventshade}\textbf{SF1}&\cellcolor{eventshade}\textbf{GPR}&\cellcolor{evidenceshade}\textbf{GF1}&\cellcolor{evidenceshade}\textbf{SF1}&\cellcolor{evidenceshade}\textbf{GPR}\\
\midrule
\endhead
\midrule
\multicolumn{11}{r@{}}{Continued on next page}\\
\endfoot
\bottomrule
\endlastfoot
\rowcolor{oursrow}APIDIS&32&48.75&49.90&52.88&\textbf{65.63}&\textbf{65.63}&\textbf{65.63}&\textbf{81.25}&\textbf{81.25}&\textbf{81.25}\\
\rowcolor{oursrow}OVT-B&11,048&46.99&56.55&47.68&48.47&50.52&51.79&54.82&58.15&58.58\\
\rowcolor{oursrow}TAO&2,546&\textbf{59.45}&\textbf{60.81}&\textbf{59.49}&40.42&41.72&42.29&52.43&55.81&54.85\\
\end{longtable}
\endgroup

\subsection{Scene Description Results}
\label{app:language-results}

Table~\ref{tab:description-binding} reports BERTScore F1 for complete scene descriptions. Table~\ref{tab:language} separates the present, future, and evidence content of those descriptions. EvoSee improves all three parts and gains 0.28 points overall over Qwen3.5-9B, with the largest gain in the present description.

\begin{table}[htbp]
\centering
\tablefont
\renewcommand{\arraystretch}{1.08}
\setlength{\tabcolsep}{2.2pt}
\captionsetup{width=\linewidth,position=top}
\caption{BERTScore F1 for the complete scene description and its present, future, and evidence parts.}
\label{tab:language}
\begin{tabularx}{\columnwidth}{@{}Z*{4}{Y}@{}}
\toprule
\multicolumn{1}{@{}c}{\textbf{Method}}&\textbf{Overall}&\textbf{Present}&\textbf{Future}&\textbf{Evidence}\\
\midrule
Qwen3.5-9B&95.98&95.59&96.65&96.31\\
\rowcolor{oursrow}\makecell[l]{\textbf{\method{} (Ours)}}&\textbf{96.26}&\textbf{96.06}&\textbf{96.81}&\textbf{96.52}\\
\bottomrule
\end{tabularx}
\end{table}

\FloatBarrier

\Needspace{9\baselineskip}
\section{Qualitative Comparisons}
\label{app:qualitative}
\makeatletter
\setlength{\@fptop}{0pt plus 1fil}
\setlength{\@fpsep}{16pt}
\setlength{\@fpbot}{0pt plus 1fil}
\makeatother
\setlength{\floatsep}{16pt}

The examples compare ground truth (GT), Qwen3.5-9B, and EvoSee under the same observation and future durations. The common question is ``Given the observed $H$-second video window, its fixed keyframes, and the camera view, what scene evolution follows within the next $H$ seconds?'' The accompanying instruction requests observed trajectories, future trajectories, and a description relating the forecast to observed evidence. Reference close-ups from observation and future appear in the right panel.

Each row uses the same four timestamps. For each entity, the displayed centers accumulate from one to four points, with solid connections in observation and dashed connections into the future. Colors follow the fixed observed identity correspondence, and labels retain each method's output identities. The right panel places reference close-ups above the two descriptions. Gains displayed in each figure are EvoSee minus Qwen3.5-9B on that sample, measured in percentage points.

\begin{figure*}[htbp]
  \centering
  \includegraphics[page=1,width=\textwidth]{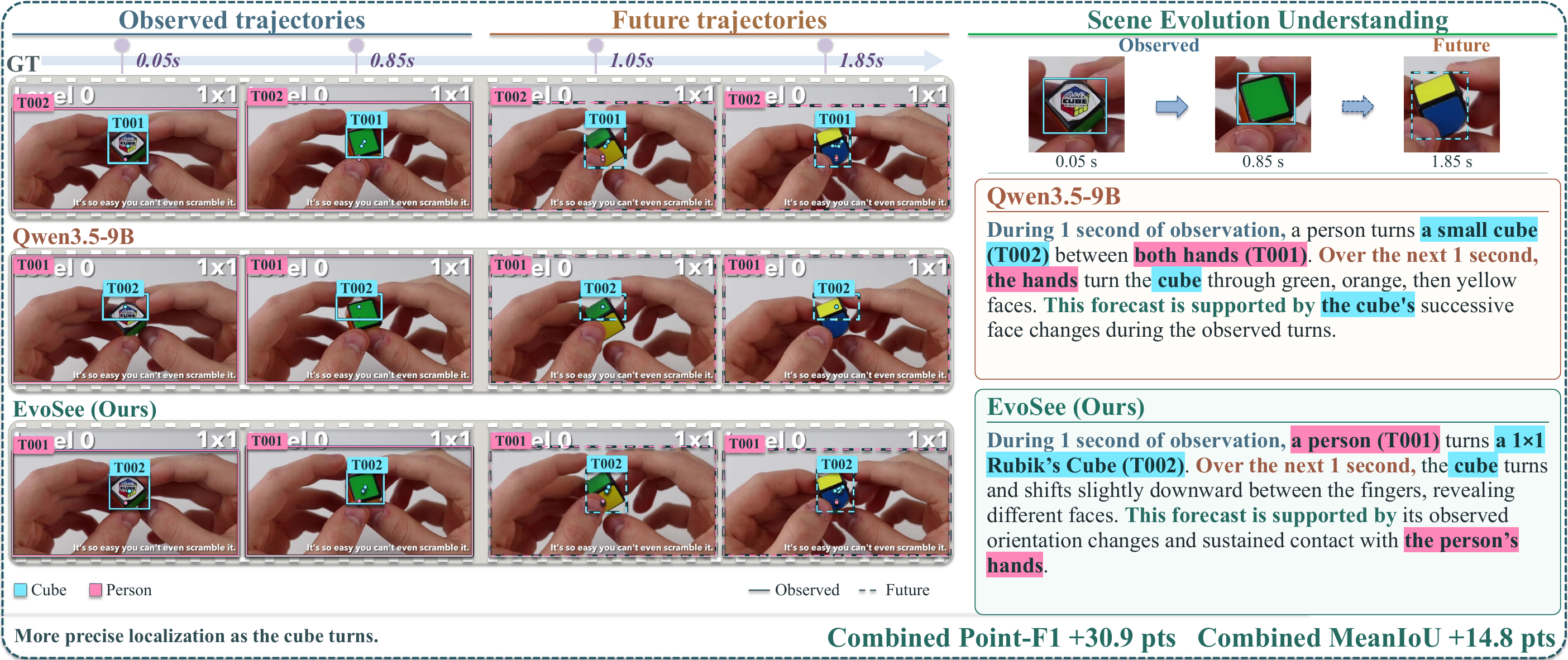}
  \caption{Cube rotation. EvoSee follows the changing spatial extent of the handheld cube, improving Combined Point-F1 by 30.9 points and MeanIoU by 14.8 points.}
  \label{fig:qual-cube}
\end{figure*}

\begin{figure*}[htbp]
  \centering
  \includegraphics[page=2,width=\textwidth]{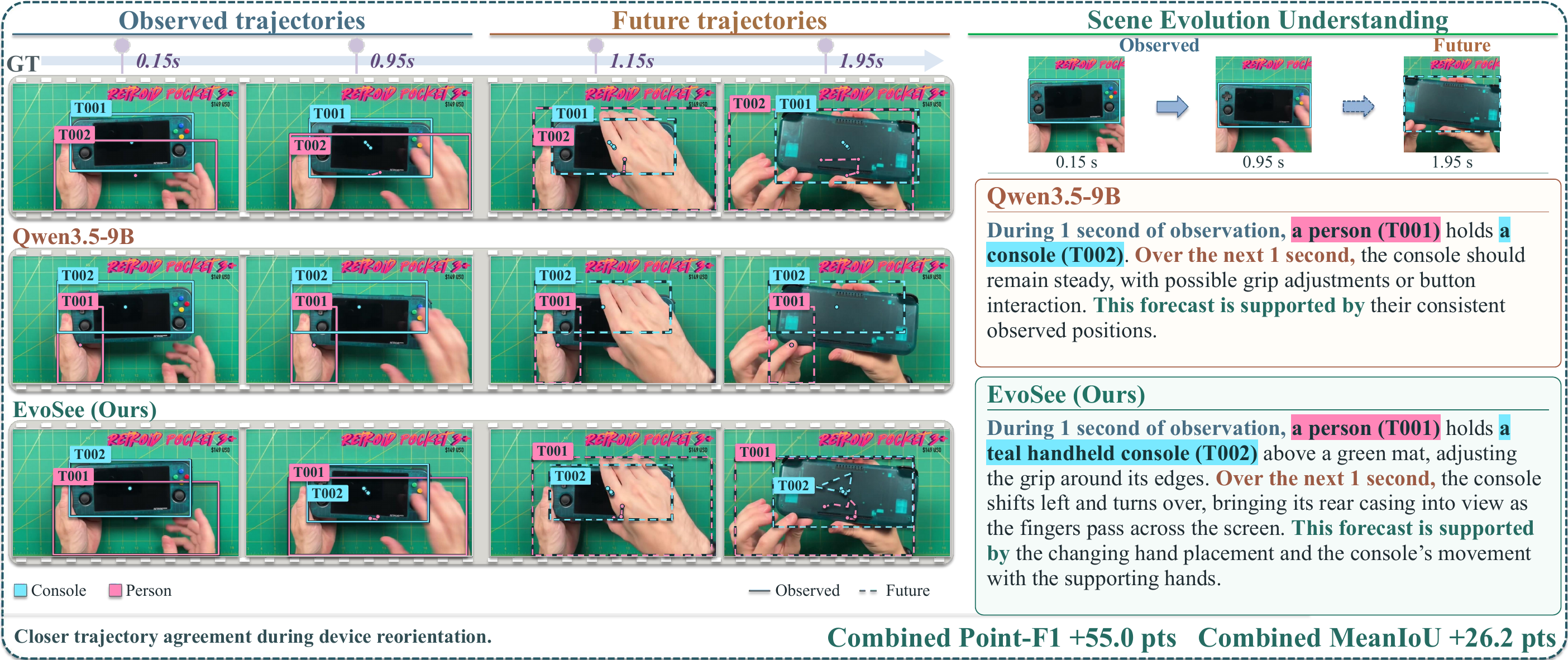}
  \caption{Device manipulation. Observed and future tracks follow a person and handheld console during reorientation. Combined Point-F1 and MeanIoU gains are 55.0 and 26.2 points.}
  \label{fig:qual-device}
\end{figure*}

\begin{figure*}[htbp]
  \centering
  \includegraphics[page=3,width=\textwidth]{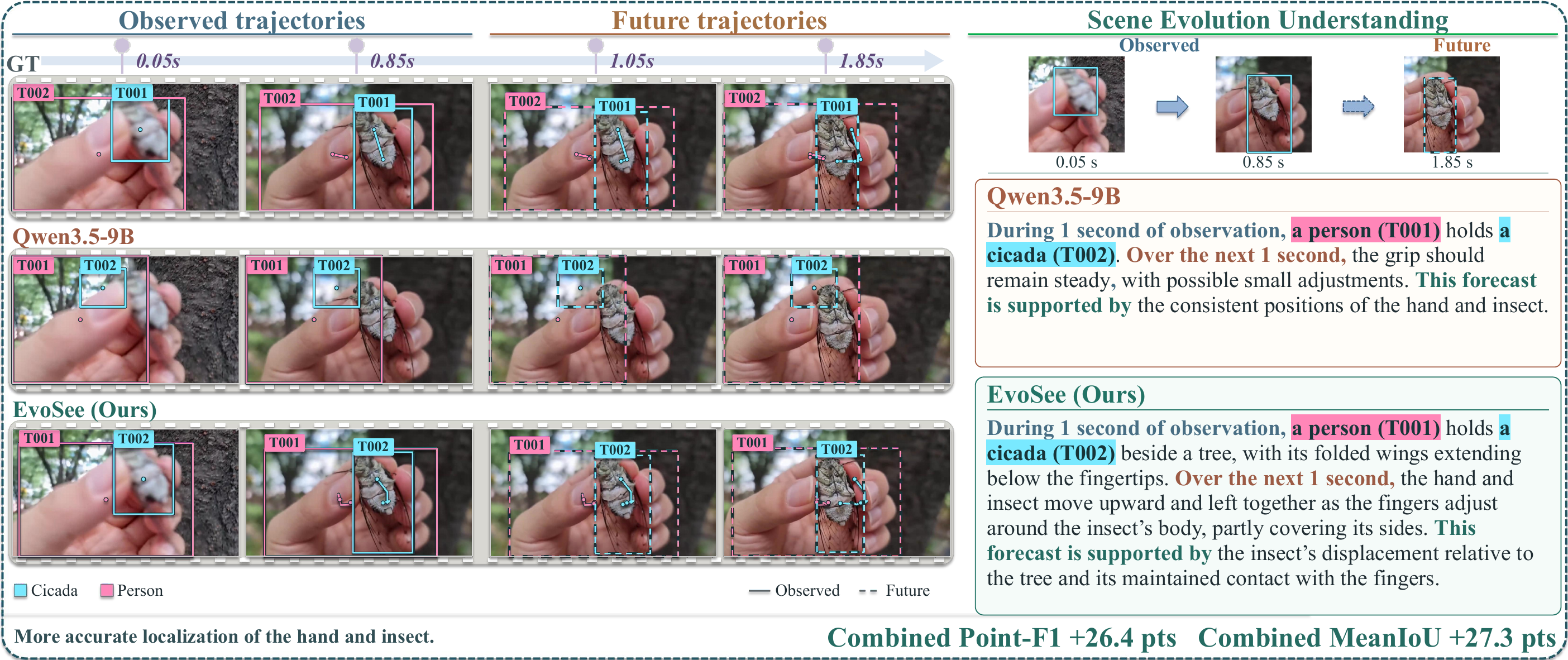}
  \caption{Hand and insect motion. The comparison shows entity localization at different spatial scales. Combined Point-F1 and MeanIoU improve by 26.4 and 27.3 points.}
  \label{fig:qual-insect}
\end{figure*}

\begin{figure*}[htbp]
  \centering
  \includegraphics[page=5,width=\textwidth]{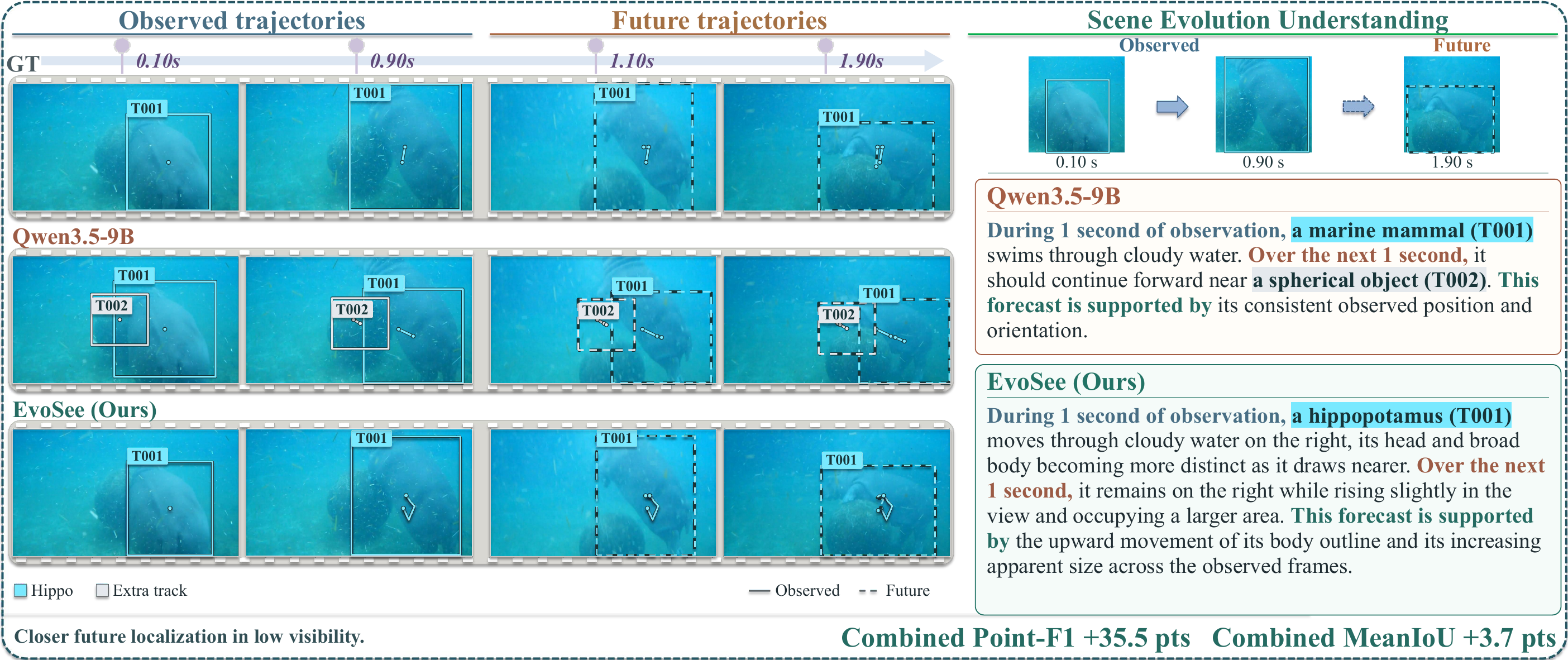}
  \caption{Underwater motion. EvoSee more closely localizes the animal through low visibility, with Combined Point-F1 and MeanIoU gains of 35.5 and 3.7 points.}
  \label{fig:qual-underwater}
\end{figure*}

\begin{figure*}[htbp]
  \centering
  \includegraphics[page=6,width=\textwidth]{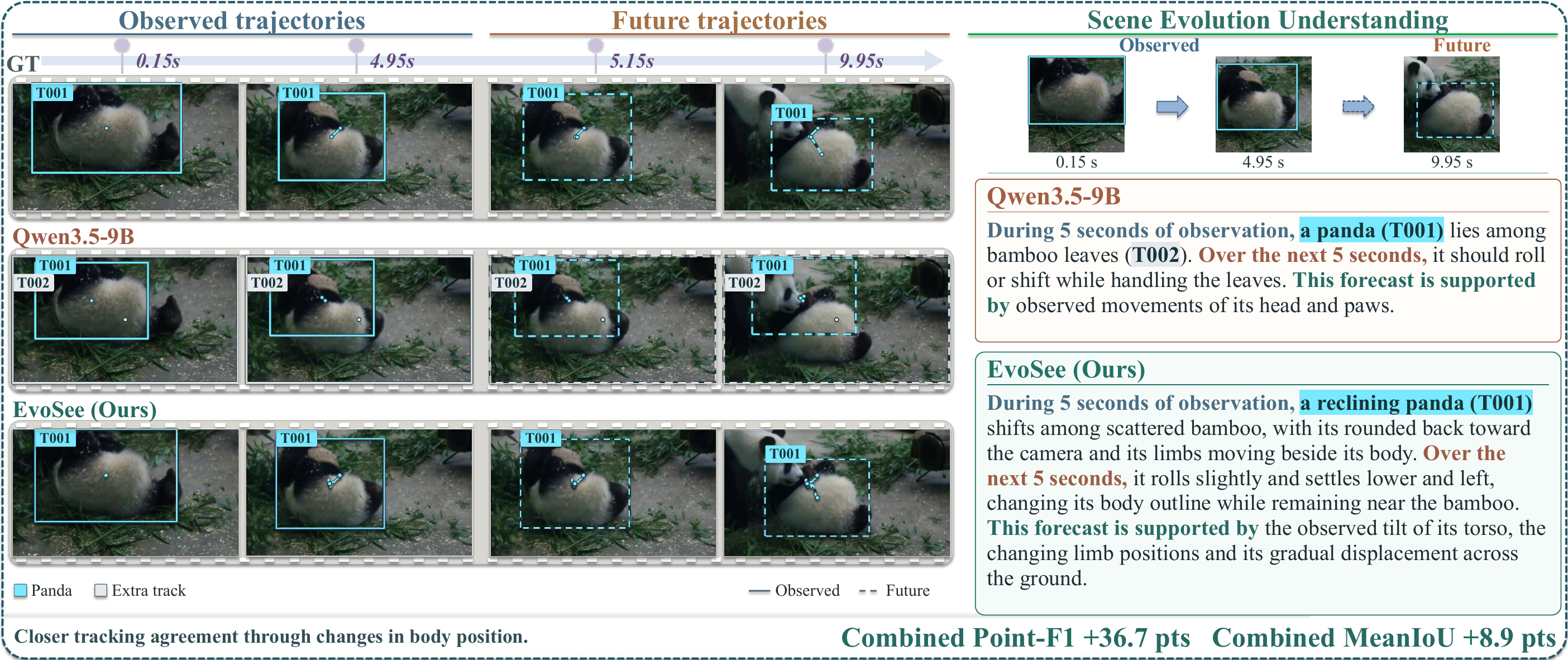}
  \caption{Extended body motion. Five seconds of observation precede a forecast of five seconds. Combined Point-F1 and MeanIoU gains are 36.7 and 8.9 points.}
  \label{fig:qual-body}
\end{figure*}

\begin{figure*}[htbp]
  \centering
  \includegraphics[page=7,width=\textwidth]{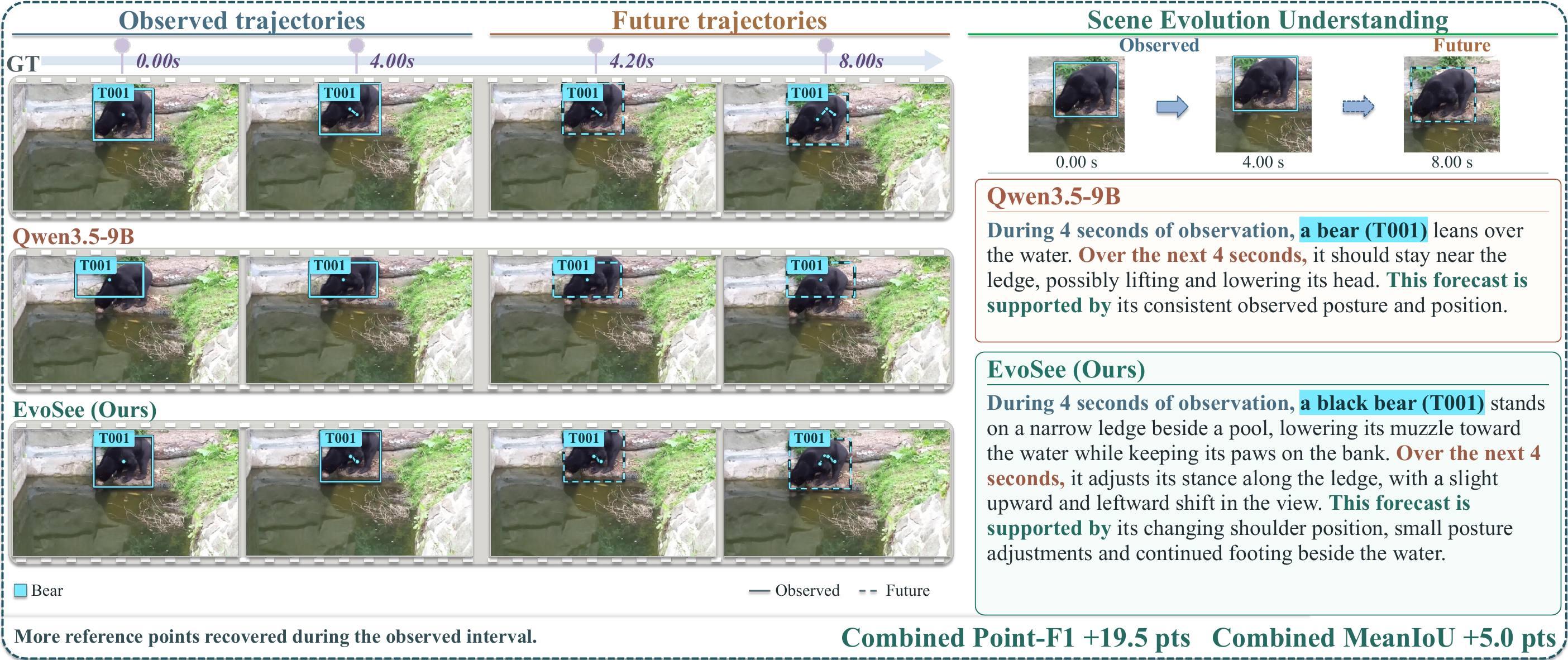}
  \caption{Motion near the water edge. Intervals of four seconds show localization around a stationary scene structure. Combined Point-F1 and MeanIoU improve by 19.5 and 5.0 points.}
  \label{fig:qual-water-edge}
\end{figure*}
\FloatBarrier

\end{document}